\documentclass[journal]{IEEEtran}

\usepackage[T1]{fontenc}
\usepackage{amsmath,amssymb,amsfonts}
\usepackage{newtxtext,newtxmath}
\usepackage{algorithmic}
\usepackage[ruled,vlined,linesnumbered]{algorithm2e}
\usepackage{graphicx}
\usepackage{url}
\usepackage{cite}
\usepackage{hyperref}

\usepackage{multirow}
\usepackage{longtable}
\usepackage{booktabs}
\usepackage{tabularx}

\usepackage{xcolor}
\usepackage{pgfplots}
\pgfplotsset{compat=1.18}
\usepackage{pgf-pie}
\usepackage{circuitikz}
\usepackage{tcolorbox}
\usepackage{subfig}

\usepackage[inline]{enumitem}

\graphicspath{{./}}
\DeclareGraphicsExtensions{.pdf,.png,.jpg}

\begin{document}

\title{Distributed LLMs and Multimodal Large Language Models: A Survey on Advances, Challenges, and Future Directions}

\author{Hadi Amini$^{1,2,*}$,~\IEEEmembership{Senior Member,~IEEE}, Md Jueal Mia$^{1,2}$, Yasaman Saadati$^{1,2}$, Ahmed Imteaj$^{3}$,~\IEEEmembership{Member,~IEEE}, \\ Seyedsina Nabavirazavi$^{1}$, Urmish Thakker$^{4}$, Md Zarif Hossain$^{3}$, \\  Awal Ahmed Fime$^{3}$, and S.S. Iyengar$^{1}$,~\IEEEmembership{Life Fellow,~IEEE}\\ \vspace{0.1in}
\small {1: Knight Foundation School of Computing and Information Sciences, Florida International University, Miami, FL, USA \\2: Security, Optimization, and Learning for InterDependent networks laboratory (solid lab), Miami, FL, USA \\ 3: School of Computing, Southern Illinois University, Carbondale, IL, USA \\ 4:
SambaNova Systems, Palo Alto, CA, USA}
    \thanks{* Corresponding Author: Hadi Amini, Florida International University, Miami, FL 33199, amini@cs.fiu.edu and hadi.amini@ieee.org.}
    \thanks{Hadi Amini, Md Jueal Mia, and Yasaman Saadati  are with the Knight Foundation School of Computing and Information Sciences, Florida International University, Miami, FL 33199, USA. They are also with the Security, Optimization, and Learning for InterDependent Networks Laboratory (solid lab) at FIU. Seyedsina Nabavirazavi and S.S. Iyengar are with FIU, Miami, FL 33199, USA. Ahmed Imteaj, Md Zarif Hossain, and Awal Ahmed Fime are with the School of Computing, Southern Illinois University, Carbondale, IL 62901, USA.}
 \thanks{ Seyedsina Nabavirazavi and S.S. Iyengar Knight Foundation School of Computing and Information Sciences, Florida International University, Miami, FL 33199, USA. }
    \thanks{Urmish Thakker is with Deep Learning Research, SambaNova Systems, CA,  USA.}
   
}

\markboth{}
{Amini \MakeLowercase{\textit{et al.}}}

\maketitle

\begin{abstract}
Language models (LMs) are machine learning models designed to predict linguistic patterns by estimating the probability of word sequences based on large-scale datasets, such as text.
LMs have a wide range of applications in natural language processing (NLP) tasks, including autocomplete and machine translation. Although larger datasets typically enhance LM performance, scalability remains a challenge due to constraints in computational power and resources. Distributed computing strategies offer essential solutions for improving scalability and managing the growing computational demand. Further, the use of sensitive datasets in training and deployment raises significant privacy concerns. Recent research has focused on developing decentralized techniques to enable distributed training and inference  while utilizing diverse computational resources and enabling edge AI.  
This paper presents a  survey on distributed solutions for various LMs, including large language models (LLMs), vision language models (VLMs),  multimodal LLMs (MLLMs), and small language models (SLMs).  While LLMs  focus on processing and generating text, MLLMs are designed to handle multiple modalities of data (e.g., text, images, and audio) and to integrate them for broader  applications. To this end, this paper reviews key advancements across the MLLM pipeline, including distributed training, inference, fine-tuning, and deployment, while also identifying the contributions,  limitations, and future areas of improvement. Further, it categorizes the literature based on six primary focus areas of decentralization. 
 Our analysis describes gaps in current methodologies for enabling distributed solutions for LMs and outline future research directions, emphasizing the need for novel solutions to enhance the robustness and applicability of distributed LMs.  By analyzing insights from existing studies and outlining future research paths, this survey aims to serve as a resource for researchers and practitioners seeking to advance the state-of-the-art in distributed MLLMs. 

\end{abstract}


\IEEEpeerreviewmaketitle

\vspace{0.1in}

~~~~~~~~~~~~~~~~~~~~~{\textbf{Paper Organization}}

 \ref{introo} Introduction (\hyperref[introo]{\color{blue}\textbf{Link}})

\ref{llmoverview} Overview of LLMs and MLLMs (\hyperref[llmoverview]{\color{blue}\textbf{Link}})

\ref{Decentral_llm_View} Decentralizing the LLM Pipeline (\hyperref[Decentral_llm_View]{\color{blue}\textbf{Link}})

\ref{summaryView} Summary, Contributions, Challenges, and Future Directions of  Related  Studies on LLMs (\hyperref[summaryView]{\color{blue}\textbf{Link}})

\ref{vlms} Summary, Contributions, Challenges, and Future Directions of  Related  Studies on VLMs (\hyperref[vlms]{\color{blue}\textbf{Link}})

\ref{slm} A Brief Overview of SLMs (\hyperref[slm]{\color{blue}\textbf{Link}})

\ref{frd} Highlights of Future Research Directions (\hyperref[frd]{\color{blue}\textbf{Link}})

\ref{conclusion} Conclusion (\hyperref[conclusion]{\color{blue}\textbf{Link}})

\noindent {\color{blue}Given the rapid advancements in this field, we have created a GitHub page to update the list of papers relevant to this survey. It is available at: \href{https://github.com/solidlabnetwork/awesome-distributed-LLM}{\textbf{Link}}.
}

\section{Introduction} \label{introo}
\IEEEPARstart{I}{n} this section, we provide an overview of the motivation for this survey on multimodal large language models (MLLMs), followed by recently published prior works, and differentiate how our proposed survey is necessary for the MLLM domain. We then highlight key aspects of this survey and the necessity of conducting this research. Finally, at the end of this section, we briefly describe the organization of this paper\footnote{Disclaimer: Considering the rapidly-evolving research on distributed/federated MLLMs, we will continue to update this article to include the most recent studies. We have created an evolving GitHub page where we list relevant papers for the research community: \href{https://github.com/solidlabnetwork/awesome-distributed-LLM}{https://github.com/solidlabnetwork/awesome-distributed-LLM}. We encourage the research community to provide feedback and suggestions to further improve future versions of this survey. 
In this version, we try to highlight the most relevant studies  including but not limited to
\cite{1zeng2023distributedLLM,2wu2023fastLLM,3li2024distflashattnLLM,4nabli2024accoLLM,5brakel2024modelLLM,6he2024distributedLLM,7borzunov2024distributedLLM,8qin2023federatedLLM,9qu2024mobileLLM,10khoshsirat2024decentralizedLLM,11ren2024taskLLM,12yao2024scalellmLLM,13duan2024efficientLLM,14li2024largeHardware,15kuang2024federatedscopeLLM,16xu2023fwdllmLLM,17ling2024convergenceLLM,18yao2024federatedLLM,19zhang2024fedrdmaLLM,20shu2024ferretLLM,21ye2024openfedllmLLM,22xu2024surveyLLM,23ye2024safetyLLM,24ye2024fedllmLLM,25wang2024federatedLLM,26wu2024fedbiotLLM,27zhang2024fedpitLLM,28huang2024frameworkLLM,29pan2024cloudLLM,30chua2023fedpeatLLM,31sani2024futureLLM,32peng2024fedpftLLM,33liu2024timeFFMLLM,34li2024synergizingLLM,35wang2024cycleblackLLM,36qu2024swarmLLM,37kang2023groundingLLM,38chen2023federated,39woisetschlager2024survey,40yu2023federated,41zhuang2023foundation, 42zheng2024safely, 43qin2024empirical, 44liu2024resource, 45li2024collm, 46pentyala2024paft, 47li2024unity, 48markov2023quantized, 49koo2024towards, 50xu2024device, 51liu2024asynchronous, 52yang2024perllm, 53hagemann2023efficient, 54huang2024distmm, 55wang2024efficient, 56li2024tpi, 57yang2024meta, 58rayLLMscaling, 59hu2023llm, 60gao2024dlora, 61gao2024fedpt, 62ghiasvand2024communication, 63qin2024federated, 64zhao2024frag, 65elbakary2024mira, 66qi2024fdlora, 67yao2024sharingLLM, 68yang2024research, 69ouyang2024pluto, 70tang2024fusionllm, 71sheng2024hybridflow, 72shen2024edgeqat, 73wang2023privatelora, 74du2024distributed, 75chen2023confidant, 76huang2024edgellm, 77xu2024hethub, 78shuai2024mitigating, 79guo2024survey, 80fang2024automated, 81fan2023fate, 82woisetschlager2024federated, 83friha2024llm, 84fu2024serverlessllm, 85xinimmediate, 86shabani2024harnessing, 87yan2024lightweight, 88zeng2024fair, 89raje2024communication, 90sadeepa2024disllm, 91popov2018distributed, 92lin2024splitlora, 93wu2024cg, 94sun2024improving, 95bai2024federated, 96gao2024efficient, 97wang2024flora, 98li2024mllm, 99zhang2024distributed, 100nguyen2024flora}. We also include some other studies that focus on LLM/MLLM/VLM/SLM and distributed/federated architectures to provide additional information to the readers of this survey}.

\subsection{Motivation}

LLMs have gained significant attention due to their ability to perform a wide range of tasks, from natural language understanding to complex problem-solving. Their development has transformed artificial intelligence, making them indispensable tools across various domains.  One of the key challenges of training and deploying LLMs is the need for extensive \textit{computational} resources. While larger training data can improve performance of LMs, there is a need to leverage decentralized techniques to enhance scalability of these models to deal with the computational limits. Further, during various stages of training and deploying LMs, there might be privacy concerns regarding the used datasets. Recently, several studies have focused on developing distributed/decentralized techniques to enhance \textit{robustness}, \textit{scalability}, \textit{efficiency}, and \textit{privacy} of LMs. The main goal is to enable distributed training and inference across multiple datasets while leveraging diverse computational resources.

\subsection{Brief History of LLMs}
Language modeling has an extensive history with fundamental studies such as  Shannon’s  seminal paper in 1951, ``Prediction and entropy of printed English'' on  application of information theory to human language \cite{shannon1951prediction}. Shannon developed a novel method to estimate  the \textit{entropy }and \textit{redundancy} of a language, leveraging the known statistics of that language \cite{shannon1951prediction}.  Later in the 1980s, statistical language models, such as n-grams, were developed to predict word sequences using probabilistic methods \cite{HLLM2jelinek1998statistical}. These models estimate the likelihood of each word based on preceding words through Maximum Likelihood Estimation, but they face challenges with storage and accuracy as sequence length increases \cite{HLLM3chu2024history}. While statistical language models marked a significant improvement over earlier rule-based systems, their struggle with data sparsity and the inability to effectively capture long-range dependencies  and contextual nuances\cite{HLLM4manning1999foundations} paved the way for more advanced approaches.

A major breakthrough occurred in the early 2010s with the introduction of Neural Language Models (NLMs), which used deep learning to significantly enhance language processing [5]. One key innovation was Word2Vec, developed by Mikolov et al. \cite{HLLM6mikolov2013distributed}, which created continuous word vectors, enabling models to understand the relationships between words by positioning similar words closer together in a multi-dimensional space. This further enhanced the accuracy of context-based predictions. In 2017, the introduction of the Transformer architecture in the paper ``Attention is All You Need'' by Vaswani et al. \cite{HLLM7vaswani2017attention} marked a significant advancement. This architecture allowed models to process text sequences in parallel and capture long-range dependencies more accurately.
This led to the rise of Pre-trained Language Models (PLMs), which are initially trained on large volumes of unlabeled text to learn basic language structures and then fine-tuned on smaller, task-specific datasets. This ``pre-training and fine-tuning'' approach has proven highly effective for tasks such as  translation and summarization, with models such as GPT-2 \cite{LLMH8radford2019language}  built on this framework. This innovation laid the foundation for large-scale models such as OpenAI's GPT series, which demonstrated exceptional generative capabilities. GPT-3, released in 2020 with 175 billion parameters, showcased unprecedented abilities in handling diverse tasks  without the need for task-specific fine-tuning \cite{LLM10mann2020language}. By 2024, models such as GPT-4 \cite{LLMH11achiam2023gpt}, LLaMA \cite{LLMH12touvron2023llama}, and NVLM \cite{LLMH13dai2024nvlm} further advanced in both scale and functionality. The expansion in model size and data volume has led to unlocking potential of LLMs as essential tools for more advanced tasks, such as  advanced reasoning.

\begin{table}
\centering
\caption{List of abbreviations used in this paper.} 
\label{table1}
\begin{tabular}{|l|l|}
\hline \textbf{Abbreviation} & \textbf{Description} \\
\hline \textbf{LM} & Language Model \\
\hline \textbf{LLM} & Large Language Model \\
\hline \textbf{NLP} & Natural Language Processing \\
\hline \textbf{NLM} & Neural Language Model \\
\hline \textbf{VLM} & Vision Language Model \\
\hline \textbf{MLLM} & Multimodal Large Language Model \\
\hline \textbf{RLHF} &  Reinforcement Learning from Human Feedback  \\
\hline \textbf{GPT} & Generative Pre-trained Transformer \\
\hline \textbf{LLaMA} &  Large Language Model Meta AI \\
\hline \textbf{PaLM} & Pathways Language Model \\
\hline \textbf{M-ICL} & Multimodal In-Context Learning \\
\hline \textbf{LAVR} & LLM-Aided Visual Reasoning \\
\hline \textbf{FL} & Federated Learning \\
\hline \textbf{FedIT} & Federated Instruction Tuning \\
\hline \textbf{FedVA} & Federated Value Alignment \\
\hline \textbf{MEI} & Mobile Edge Intelligence \\
\hline
 \textbf{DPO} & Direct Preference Optimization \\
\hline
 \textbf{PPO} & Proximal Policy Optimization \\
\hline
\textbf{SLM} & Small Language Model \\
\hline
\textbf{PEFT} & Parameter-Efficient Fine-Tuning \\
\hline
\textbf{FM} & Foundation Model\\
\hline
\textbf{NPU} & Neural Processing Unit\\
\hline 
\textbf{RDMA} & Remote Direct Memory Access \\
\hline 
\textbf{LCM} & Large Concept Model \\
\hline
\textbf{IID} & Independent and Identically Distributed \\
\hline
\end{tabular}
\end{table}

\vspace*{-0.25cm}
\subsection{Related works and Contributions}
\underline{Large Language Models}: LLMs have received growing attention since  the launch of ChatGPT in November 2022 due to their superior performance in multiple natural language tasks \cite{minaee2024largeLLM}. They can achieve general-purpose language understanding and generation by training billions of parameters on extensive text datasets, following principles predicted by scaling laws. The field of LLMs is evolving rapidly, with ongoing research producing new models and techniques at a rapid pace \cite{minaee2024largeLLM}. Advances in LLMs research have affected the entire AI community and has the potential to revolutionize not only the way researchers develop and use AI \cite{LLMSurvey1zhao2023survey}, but also the way several real-world systems such as robotics and critical infrastructures operate. 
Minaee et al \cite{minaee2024largeLLM} provides a comprehensive survey of notable LLM families, such as GPT, LLaMA, and PaLM that highlights their characteristics, contributions, and limitations. They also  evaluated techniques for building and augmenting LLMs, along with the datasets used for training, fine-tuning, and evaluation; followed by an  assessment of popular evaluation metrics and comparison of the performance of several LLMs against representative benchmarks \cite{minaee2024largeLLM}. While LLMs offer significant advances, there is a broader integration of 
MLLMs. MLLMs have emerged as a significant research focus, leveraging the capabilities of powerful LLMs to tackle multimodal tasks \cite{yin2023surveyMMLM}. According to Yin et al \cite{yin2023surveyMMLM}, these models have better capabilities, including generating narratives from images and performing mathematical reasoning without optical character recognition (OCR). A detailed overview of the literature on MLLMs,  foundational concepts, architectural frameworks, training methodologies, and evaluation strategies of MLLMs, as well as challenges such as multimodal hallucination and advanced techniques such as Multimodal In-Context Learning (M-ICL) and LLM-Aided Visual Reasoning (LAVR) are provided in \cite{yin2023surveyMMLM}. There are several comprehensive surveys that cover the fundamentals of LLMs, VLMs, and MLLMs, including \cite{LLMSurvey1zhao2023survey, yin2023surveyMMLM, LLMSurvey3hadi2023survey,LLMSurvey4chang2024survey,minaee2024largeLLM};  applications of  LLMs for various domains and use-cases, including information retrieval \cite{11ren2024taskLLM}, recommendation \cite{LLMSurvey8wu2024survey},  graphs \cite{LLMSurvey9ren2024survey}, education \cite{LLMSurvey10wang2024large}, healthcare \cite{LLMSurvey11he2023survey}, and autonomous driving \cite{LLMSurvey12cui2024survey}; and security and privacy challenges of LLMs \cite{SurveyLLM14yao2024survey,SurveyLLM15das2024security}. Given the main focus of our survey article, i.e., distributed MLLMs, we refer the audience to the existing surveys for fundamentals of LLMs.

{\color{black}
\underline{Low-Cost Edge Devices for Local LLM Inference:} Ongoing research efforts are focused on enabling the deployment of LLMs on edge devices to bring the power of advanced AI closer to users (at the edge) while addressing challenges related to resource constraints, latency, and privacy. As a part of this, edge computing devices such as  \textit{NVIDIA Jetson Orin Nano} marks a significant step in enabling local inference of LLMs on edge devices to revolutionize AI deployment.  
These models traditionally rely on cloud-based infrastructure. With emerging low-price computing devices can now be processed locally. For instance,  Jetson’s advanced GPU architecture and energy-efficient design can ensure low latency, enhanced privacy, and  reliable performance. Low-cost devices such as the Jetson Orin Nano are designed for developers, small businesses, and hobbyists, and can pave the way democratize access to advanced AI tools \cite{verge2024nvidia,wsj2024nvidia}. According to the manufacturer,  Jetson Orin Nano can handle LLMs with up to 7 billion parameters, making it suitable for a range of applications that demand real-time processing and secure data handling. For larger-scale needs, the Jetson Orin NX 16GB supports models with 13 billion parameters, while the Jetson AGX Orin 64GB manages massive 70 billion parameter models, such as Llama-2-70B, at interactive rates, demonstrating its versatility and capability for robust edge AI solutions \cite{nvidia2024jetson}. }

\begin{figure}[h!]
  \centering
  \includegraphics[width=0.95\linewidth]{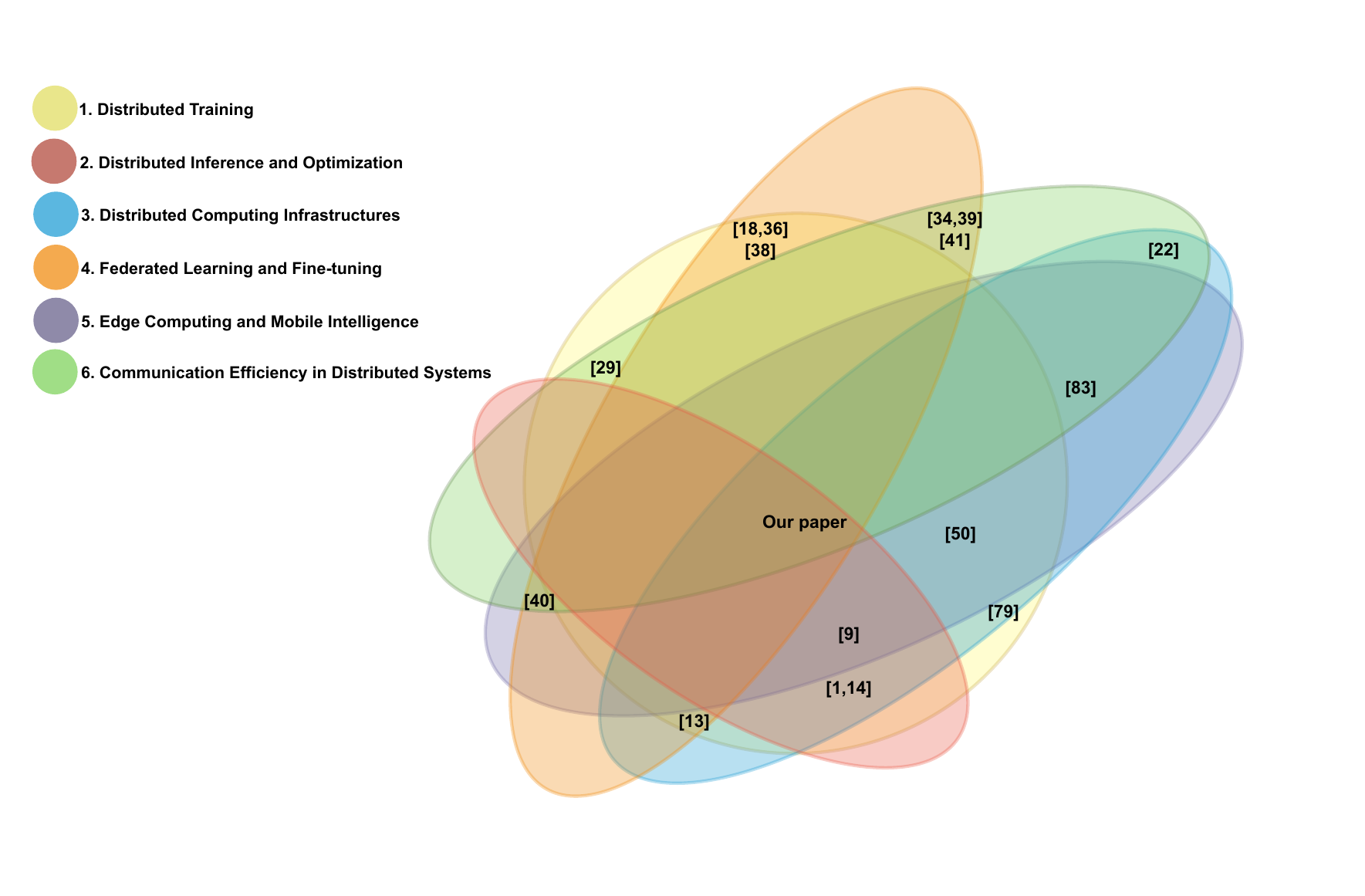} 
  \caption{Overview of selected existing Survey papers in six categories and their overlapping areas.}
  \label{surveysummaries}  
\end{figure}

\underline{Existing Surveys on Distributed/Federated LLMs}: A number of domain-specific studies explored existing works on decentralizing LLMs from multiple perspectives, including but not limited to \cite{9qu2024mobileLLM, 5brakel2024modelLLM, 13duan2024efficientLLM, 18yao2024federatedLLM}. Table \ref{tablesurveys} presents an overview of the primary and secondary focus areas covered in selected existing survey papers. Note that  some existing surveys may discuss  topics beyond those listed implicitly. Further, Figure \ref{surveysummaries} shows an overview of these survey articles and their overlapping areas in the six categories.

\begin{table}[h]
\centering
\caption{Comparing Our Work with the some of the Existing Surveys on Federated/Distributed LLMs (The numbers in the \textbf{Focus} column correspond to the following topics: \textbf{1.} Distributed Training; \textbf{2.} Distributed Inference and Optimization; \textbf{3.} Distributed Computing Infrastructures;  \textbf{4.} FL and Fine-tuning; \textbf{5.} Edge Computing and Mobile Intelligence; \textbf{6. } Communication Efficiency in Distributed Systems)}

\label{tablesurveys}
\begin{tabular}{|c|c|c|}
\hline
\textbf{Survey} & \textbf{ Main Focus } & \textbf{Secondary Focus }  \\  \hline
Zeng et al. \cite{1zeng2023distributedLLM} & \textbf{1, 2, 3} & \textbf{6} \\ \hline
 Qu et al. \cite{9qu2024mobileLLM} &  \textbf{1, 2, 3, 5} & \textbf{6}  \\ \hline
Duan et al. \cite{13duan2024efficientLLM} &  \textbf{1, 3, 4} & \textbf{6}  \\ \hline
Li et al. \cite{14li2024largeHardware
} &  \textbf{1, 2, 3} & \textbf{6}  \\ \hline
Yao et al. \cite{18yao2024federatedLLM} & \textbf{1, 4} & \textbf{6}  \\ \hline
Xu et al. \cite{22xu2024surveyLLM} & \textbf{3, 6} & \textbf{1, 2, 4} \\ \hline
Pan et al. \cite{29pan2024cloudLLM} & \textbf{1, 6} & \textbf{4} \\ \hline
Li et al. \cite{34li2024synergizingLLM} &  \textbf{4, 6} & \textbf{1, 2}  \\ \hline
Qu et al. \cite{36qu2024swarmLLM
} &  \textbf{1, 4} & \textbf{2, 3, 6}  \\ \hline
Chen et al. \cite{38chen2023federated} & \textbf{1, 4 } & \textbf{6}  \\ \hline
Woisetschl{\"a}ger et al. \cite{39woisetschlager2024survey} & \textbf{4, 6} & \textbf{1}  \\ \hline
Yu et al. \cite{40yu2023federated} &  \textbf{4, 5, 6} & \textbf{1}  \\ \hline
Zhuang et al. \cite{41zhuang2023foundation} &  \textbf{4, 6} & \textbf{1, 5}  \\ \hline
Xu et al. \cite{50xu2024device} & \textbf{1, 3, 5} & \textbf{4, 6} \\ \hline
Guo et al. \cite{79guo2024survey} & \textbf{1, 3} & \textbf{4, 6} \\ \hline
Friha et al. \cite{83friha2024llm} & \textbf{3, 5, 6} & \textbf{1, 4} \\ \hline
Our Work & \textbf{1, 2, 3, 4, 5, 6 }& -  \\ \hline
\end{tabular}
\end{table}

\begin{figure*}[h!]
  \centering
  \includegraphics[width=0.95\linewidth]{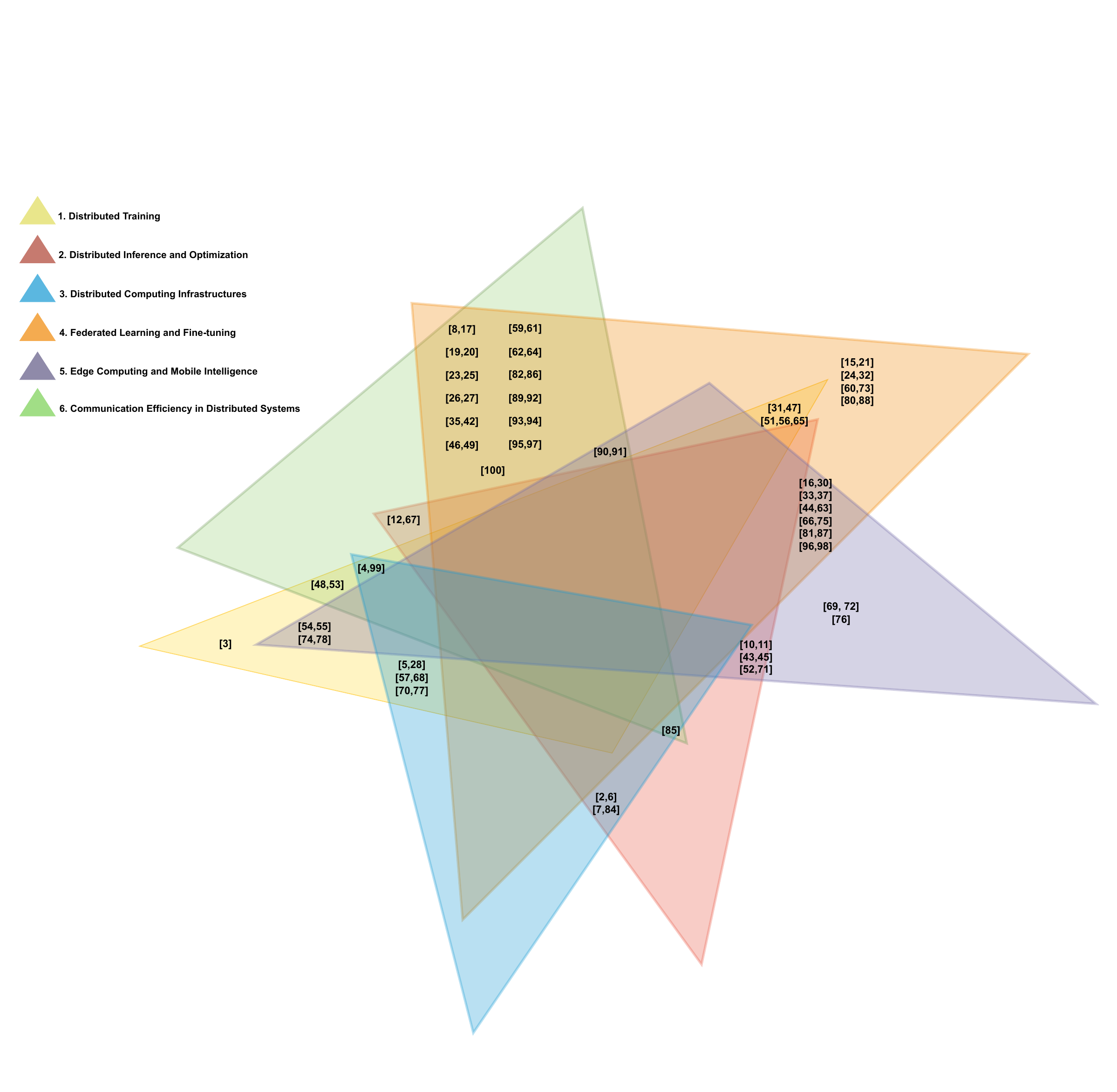} 
  \caption{Overview of selected related works in six categories and their overlapping areas.}
  \label{intersection}  
\end{figure*}

Yao et al \cite{18yao2024federatedLLM} explored FL for LLMs (FedLLM) with a focus on recent findings and future directions on \textit{ﬁne-tuning} and \textit{prompt learning} in a federated setting. Qu et al \cite{9qu2024mobileLLM} provided a survey on how mobile edge intelligence (MEI) can serve a promising solution to bridge the gap between on-device and cloud-based AI, with a focus on leveraging MEIs for LLMs, namely MEI4LLM. Brakel et al \cite{5brakel2024modelLLM} conducted a thorough literature review to answer three research questions ``What types of model parallelism exist?, What are the challenges of model parallelism?, and What is a modern use-case
of model parallelism?''. Duan et al \cite{13duan2024efficientLLM} conducted a survey on  recent advances in training systems for LLMs that includes but not limited to parallelism strategies,  optimizations for computation, communication, and memory in distributed LLM training. 
While these surveys provide valuable insights on various aspects of the decentralization/parallelization in LLMs, there is a need for a survey that covers various aspects of existing studies, and also outlines the future directions not only for LLMs but also for VLMs and in general for MLLMs. To this end, we conduct a survey that covers multiple layers of decentralization, including: 1) Distributed Training, 2) Distributed  Inference and Optimization, 3) Distributed Computing  Infrastructures, 4) FL and Fine-tuning, 5) Edge Computing and Mobile Intelligence, and 6) Communication Efficiency in Distributed Systems. Tables \ref{table1allpapers} and \ref{table1allpapers2} provide a summary of these studies. Further, Fig.  \ref{intersection} shows the papers in these six categories while secondary focus of highlighting studies that cover more than one category.

\begin{table*}
\centering
\caption{Summary of Papers on Distributed Computing for LLMs (The numbers in the \textbf{Focus} column correspond to the following topics: \textbf{1.} Distributed Training; \textbf{2.} Distributed Inference and Optimization; \textbf{3.} Distributed Computing Infrastructures;  \textbf{4.} FL and Fine-tuning; \textbf{5.} Edge Computing and Mobile Intelligence; \textbf{6. } Communication Efficiency in Distributed Systems)}
\label{table1allpapers}
\begin{tabular}{|c|c|c|c|}
\hline
\textbf{Citation} & \textbf{Primary Focus} & \textbf{Novelty} & \textbf{Challenges} \\ \hline
\cite{2wu2023fastLLM} & 2, 3 & Inference Serving & Latency, Resource Allocation \\ \hline
\cite{3li2024distflashattnLLM} & 1 & Memory-Efficient Attention & Scalability, Communication Overhead  \\ \hline 
\cite{4nabli2024accoLLM} & 1, 3, 6 & Communication Hiding & Synchronization, Model Convergence \\ \hline
\cite{5brakel2024modelLLM} & 1, 3 & Model Parallelism & Hardware Limitations, Load Balancing \\ \hline
\cite{6he2024distributedLLM} & 2, 3 & CPU Inference & Performance Bottlenecks, Memory Management  \\ \hline
\cite{7borzunov2024distributedLLM} & 2, 3 & Decentralized Inference & Device Reliability, Data Privacy \\ \hline
\cite{8qin2023federatedLLM} & 4, 6 & Communication Efficiency & Model Convergence, Scalability \\ \hline
\cite{10khoshsirat2024decentralizedLLM} & 2, 5 & Decentralized Inference & Energy Harvesting, Network Dynamics \\ \hline
\cite{11ren2024taskLLM} & 2, 5 & Task Scheduling & Heterogeneity, Network Constraints \\ \hline
\cite{12yao2024scalellmLLM} & 2, 6 & Resource-Frugal Inference & High Latency, Concurrency Bottlenecks \\ \hline
\cite{15kuang2024federatedscopeLLM} & 4 & Federated Fine-Tuning & Data Heterogeneity, Resource Constraints \\ \hline
\cite{16xu2023fwdllmLLM} & 4, 5 & BP-Free Training & Scalability, Perturbation Efficiency \\ \hline
\cite{17ling2024convergenceLLM} & 4, 6 & Zeroth-Order Optimization & Convergence, Personalization \\ \hline
\cite{19zhang2024fedrdmaLLM} & 4, 6 & Cross-Silo RDMA Efficiency & WAN Stability, Compatibility \\ \hline
\cite{20shu2024ferretLLM} & 4, 6 & Full-Parameter Federated Tuning & Communication Overhead, Privacy \\ \hline
\cite{21ye2024openfedllmLLM} & 4 & Federated Instruction Tuning (FedIT) & Data Scarcity, Privacy Concerns \\ \hline
\cite{23ye2024safetyLLM} & 4, 6 & Safety Attack and Defense & Malicious Clients, Model Vulnerabilities \\ \hline
\cite{24ye2024fedllmLLM} & 4 & Realistic Federated Benchmarks & Data Heterogeneity, Performance Evaluation \\ \hline
\cite{25wang2024federatedLLM} & 4, 6 & Domain Coverage Augmentation & Privacy, Scalability \\ \hline
\cite{26wu2024fedbiotLLM} & 4, 6 & Bi-Level Optimization & Data Heterogeneity, Alignment Issues \\ \hline
\cite{27zhang2024fedpitLLM} & 4, 6 & Few-Shot Federated Tuning & Privacy, Scalability \\ \hline
\cite{28huang2024frameworkLLM} & 1, 3 & Secure Distributed Training & Scalability, Communication Overhead \\ \hline
\cite{30chua2023fedpeatLLM} & 4, 5 & Emulator-Assisted Tuning & Privacy, Resource Constraints \\ \hline
\cite{31sani2024futureLLM} & 1, 4 & Federated LLM Pre-Training & Privacy, Statistical Heterogeneity \\ \hline
\cite{32peng2024fedpftLLM} & 4 & Proxy Fine-Tuning & Gradient Errors, Scalability \\ \hline
\cite{33liu2024timeFFMLLM} & 4, 5 & Time Series Forecasting using LMs & Cross-Domain Heterogeneity, Privacy \\ \hline
\cite{35wang2024cycleblackLLM} & 4, 6 & Full Parameter Tuning with Block Updates & Computational Bottlenecks, Scalability \\ \hline
\cite{37kang2023groundingLLM} & 4, 5 & Federated Transfer Learning Framework & Resource Constraints, Data Privacy \\ \hline
\cite{42zheng2024safely} & 4, 6 & Secure Federated Training & Privacy, Scalability \\ \hline
\cite{43qin2024empirical} & 2, 5 & Edge Deployment Guidelines & Customization, Resource Constraints \\ \hline
\cite{44liu2024resource} & 4, 5 & Mobile Edge Resource Allocation & Latency, Model Stability \\ \hline
\cite{45li2024collm} & 2, 5 & Collaborative Inference Framework & Communication, Load Balancing \\ \hline
\cite{46pentyala2024paft} & 4, 6 & Parallel Fine-Tuning Framework & Alignment Tax, Sparsity \\ \hline
\cite{47li2024unity} & 1, 4 & Semi-Asynchronous Training & Stragglers, Resource Efficiency \\ \hline
\cite{48markov2023quantized} & 1, 6 & Quantized Distributed Training & Communication Bottlenecks \\ \hline
\cite{49koo2024towards} & 4, 6 & Federated Low-Rank Adaptation & Aggregation Discordance, Heterogeneity \\ \hline
\cite{51liu2024asynchronous} & 1, 4 & Asynchronous Local-SGD & Gradient Staleness \\ \hline
\cite{52yang2024perllm} & 2, 5 & Personalized Inference Scheduling & Dynamic Resources \\ \hline
\cite{53hagemann2023efficient} & 1, 6 & Efficient Parallelization Layouts & Memory Constraints, Scaling \\ \hline
\cite{54huang2024distmm} & 1, 5 & Distributed Multimodal Training & Communication, Submodule Heterogeneity \\ \hline
\cite{55wang2024efficient} & 1, 5 & Multi-Task Training Optimization & Workload Heterogeneity \\ \hline
\cite{56li2024tpi} & 1, 4 & Data Heterogeneity Awareness & Scalability, Efficiency in Multi-task Scenarios \\ \hline
\cite{57yang2024meta} & 1, 3 & Meta-Learning for Inference & Resource Sharing, Decentralized Environment \\ \hline
\cite{59hu2023llm} & 4, 6 & Parameter-Efficient Tuning & Latency, Privacy Preservation \\ \hline
\cite{60gao2024dlora} & 4 & Distributed LoRA Fine-Tuning & Communication Costs, Convergence \\ \hline
\cite{61gao2024fedpt} & 4, 6 & Proxy-Tuning for FL & Edge Device Limitations, Model Generalization \\ \hline
\cite{62ghiasvand2024communication} & 4, 6 & Tensorized Communication & Memory Overhead, Latency \\ \hline
\cite{63qin2024federated} & 4, 5 & FedIT & Data Privacy, Scalability \\ \hline
\cite{64zhao2024frag} & 4, 6 & Federated Vector DB Management & Communication Bottlenecks, Security \\ \hline
\cite{65elbakary2024mira} & 1, 4 & Multi-Task FL Framework & Task Allocation, Resource Sharing \\ \hline
\cite{66qi2024fdlora} & 4, 5 & Dual LoRA Tuning & Resource Constraints, Scalability \\ \hline
\cite{67yao2024sharingLLM} & 2, 6 & Subspace Analysis in LoRA & Bottleneck Mitigation, Personalization \\ \hline
\cite{68yang2024research} & 1, 3 & Cross-Cloud Federated Training & Resource Allocation, Synchronization \\ \hline
\cite{69ouyang2024pluto} & 5 & Collaborative Edge AI Framework & Latency, Data Privacy \\ \hline
\cite{70tang2024fusionllm} & 1, 3 & Adaptive Compression & GPU Utilization, Memory Sharing  \\ \hline
\end{tabular}
\end{table*}

\begin{table*}
\centering
\caption{\textbf{[Continued from Table \ref{table1allpapers}]}Summary of Papers on Distributed Computing for LLMs  (The numbers in the \textbf{Focus} column correspond to the following topics: \textbf{1.} Distributed Training; \textbf{2.} Distributed Inference and Optimization; \textbf{3.} Distributed Computing Infrastructures;  \textbf{4.} FL and Fine-tuning; \textbf{5.} Edge Computing and Mobile Intelligence; \textbf{6. } Communication Efficiency in Distributed Systems)}
\label{table1allpapers2}
\begin{tabular}{|c|c|c|c|}
\hline
\textbf{Citation} & \textbf{Primary Focus} & \textbf{Novelty} & \textbf{Challenges} \\ \hline
\cite{71sheng2024hybridflow} & 2, 5 & Flexible RLHF Framework & Scalability, Training Convergence \\ \hline
\cite{72shen2024edgeqat} & 5 & Quantization Aware Training & Accuracy Loss, Latency \\ \hline
\cite{73wang2023privatelora} & 4 & Privacy-Preserving LoRA & Model Generalization, Privacy Tradeoffs \\ \hline
\cite{74du2024distributed} & 1, 5 & Multi-modal Distributed Training & Modality Integration, Resource Efficiency \\ \hline
\cite{75chen2023confidant} & 4, 5 & Collaborative Edge Training & Data Heterogeneity, Edge Constraints \\ \hline
\cite{76huang2024edgellm} & 5 & CPU-FPGA Heterogeneous Accelerator & Hardware Integration, Efficiency \\ \hline
\cite{77xu2024hethub} & 1, 3 & Heterogeneous Cluster Management & Load Balancing, Communication Overhead \\ \hline
\cite{78shuai2024mitigating} & 1, 5 & Cross-Modal Alignment & Modality Privacy, Data Labeling \\ \hline
\cite{80fang2024automated} & 4 & Automated Federated Pipeline & Resource Management, Scalability \\ \hline
\cite{81fan2023fate} & 4, 5 & Industrial-Grade FL Framework & Real-time Processing, Interoperability \\ \hline
\cite{82woisetschlager2024federated} & 4, 6 & Fine-Tuning on Edge Devices & Latency, Model Adaptation \\ \hline
\cite{84fu2024serverlessllm} & 2, 3 & Low-Latency Serverless Inference & Scalability, Dynamic Workloads \\ \hline
\cite{85xinimmediate} & 2, 3, 6 & Immediate Communication & Latency Hiding, GPU Utilization \\ \hline \cite{86shabani2024harnessing} & 4, 6 & FL with LoRA for LLM Fine-tuning & Convergence on Non-IID Data, Communication Overhead, Privacy \\ \hline
\cite{87yan2024lightweight} & 4, 5 & Unsupervised FL with Pretrained VLMs & Pseudo-Labeling, Data Heterogeneity, Resource Constraints \\ \hline
\cite{88zeng2024fair} & 4 & Fairness-Aware FL with Biased VLMs & Debiasing VLMs, Data Heterogeneity, Computational Cost \\ \hline
\cite{89raje2024communication} & 4, 6 & Sparsified LoRA for Efficient FL & Comm.-Performance Trade-off, Optimal Sparsity, Scalability \\ \hline
\cite{90sadeepa2024disllm} & 4, 5, 6 & DisLLM: SFL + LoRA for LLMs & Model Splitting, Privacy-Performance Balance, Resource Constraints \\ \hline
\cite{91popov2018distributed} & 4, 5, 6 & Federated Fine-Tuning with Privacy Guarantees & Catastrophic Forgetting, Privacy Preservation, Communication Efficiency \\ \hline
\cite{92lin2024splitlora} & 4, 6 & SplitLoRA: Split Learning with LoRA & Model Splitting, Data Heterogeneity, Communication Overhead \\ \hline
\cite{93wu2024cg} & 4, 6 & Gradient Compression in FL & Gradient Reconstruction, Privacy, Compression-Performance Balance \\ \hline
\cite{94sun2024improving} & 4, 6 & FFA-LoRA for Privacy in FL & Noisy Gradients, Hyperparameter Tuning, Data Heterogeneity \\ \hline
\cite{95bai2024federated} & 4, 6 & FlexLoRA: SVD-based Aggregation & Resource Heterogeneity, Computational Overhead, Scalability \\ \hline
\cite{96gao2024efficient} & 4, 5 & PMG-FL: Personalized FL & Non-IID Data, Resource Constraints, Knowledge Fusion \\ \hline
\cite{97wang2024flora} & 4, 6 & FLoRA: Stacking-Based Aggregation & Aggregation Noise, Heterogeneous LoRA, Communication Overhead \\ \hline
\cite{98li2024mllm} & 4, 5 & MLLM-FL: Multimodal LLMs in FL & Data Heterogeneity, Scalability, Long-Tailed Distributions \\ \hline
\cite{99zhang2024distributed} & 1, 3, 6 & Distributed FM Training in 6G & Data Heterogeneity, Communication Instability, Device Disparities \\ \hline
\cite{100nguyen2024flora} & 4, 6 & LoRA-Enhanced VLMs in FL & Data Heterogeneity, Communication Costs, Privacy, Efficiency \\ \hline
\end{tabular}
\end{table*}

\begin{figure}[h!]
  \centering  \includegraphics[width=0.95\linewidth]{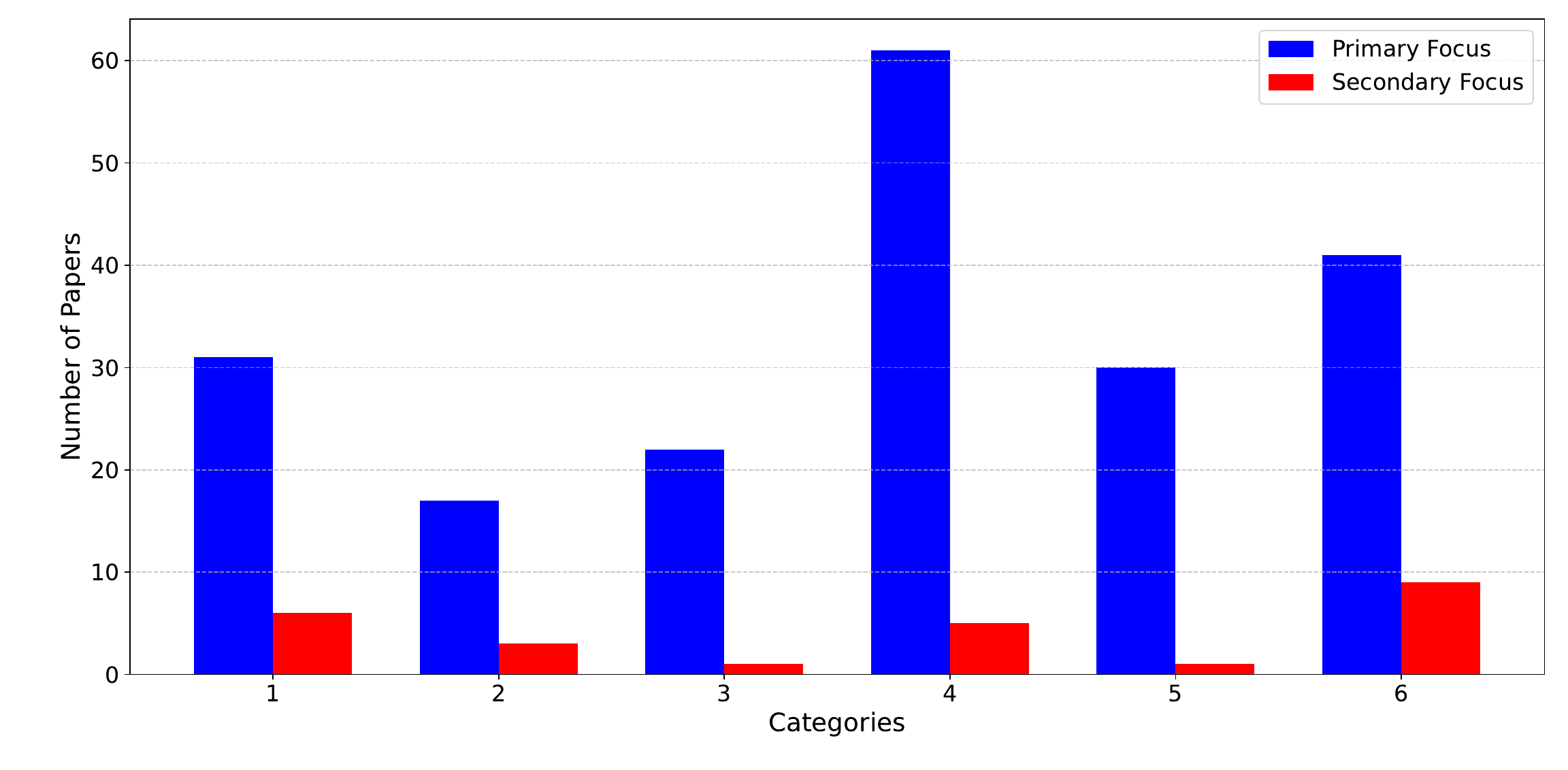} 
  \caption{Distribution of primary and secondary focus of   papers [1]-[100] considering following categories:  \textbf{1.} Distributed Training; \textbf{2.} Distributed Inference and Optimization; \textbf{3.} Distributed Computing Infrastructures;  \textbf{4.} FL and Fine-tuning; \textbf{5.} Edge Computing and Mobile Intelligence; \textbf{6. } Communication Efficiency in Distributed Systems).}
  \label{fig:1}  
\end{figure}

 \begin{center}
\begin{tcolorbox}
\vspace{-0.05in}
\noindent \textbf{Contributions:}
 We provide a comprehensive survey of recent advancements on distributed MLLMs/LLMs/VLMs/SLMs, discuss the challenges of the state of the art solutions, and provide a roadmap for future research. We also categorize the reviewed papers in terms of primary and secondary focus areas from six critical aspects of decentralization during training, inference, and fine-tuning stages and the underlying computing infrastructures.

\end{tcolorbox}
\end{center}

\vspace*{-0.25cm}
\subsection{Organization}

The rest of this paper is organized as follows. In Section \textbf{II}, we present an overview of the LLM and MLLM pipelines, and explain  their differences. Section \textbf{III} describes how each step of the described LLM pipeline can be  decentralized. This section also explains overview of FL. Section \textbf{IV} provides a summary of selected studies on LLMs with a main  focus on non-survey articles. This section, highlights the key contributions of each study, its  challenges, and identifies some of the  potential future research directions for that specific study.  Section \textbf{V} summarizes slected related works on VLMs and decentralized VLMs by  outlining their contributions, challenges, and suggested future directions. Section \textbf{VI} presents a brief overview of Small Language Models (SLMs) and their applications.  Section \textbf{VII} provides a more aggregated look into future research directions. Finally, section \textbf{VIII} concludes the paper.

\section{Overview of  LLMs and MLLMs} \label{llmoverview}

This section covers the definition of LLMs, MLLM, a 
 detailed description of the MLLM/LLM/VLM taxonomy, and  a brief highlight of the LLM pipeline.

\subsection{Definition of Multimodal Large Language Models}

While LLMs have led to a paradigm shift in natural language processing (NLP) due to their reasoning, instruction-following, and in-context learning capabilities \cite{MLLM1lmsys_vicuna_2023}, their focus has been limited to a single data modality, i.e.,  \textit{text}. However, they can possibly benefit from other modalities of data, such as vision \cite{yin2023surveyMMLM}. In order to leverage multiple modalities beyond text, MLLMs have been introduced and studies in the prior works. MLLMs are capable perceiving,  reasoning, and generating outputs across multiple modalities \cite{yin2023surveyMMLM}. A MLLM refers to an LLM-based model that can receive, process, and produce information in various modalities, including but not limited to text, images, videos, and audio \cite{yin2023surveyMMLM, Brown2020language, Wei2021zero_shot}. 

\begin{center}
\begin{tcolorbox}
\vspace{-0.05in}
\noindent 
 ``\textit{\textbf{Multimodal LLMs} offer the possibility of expanding the impact of language-only systems with novel interfaces and capabilities, enabling them to solve new tasks and provide novel experiences for their users }''\cite{openai2023gpt4vsystemcard}

\end{tcolorbox}
\end{center}

Generally, MLLMs are different than traditional multimodal approaches in two main aspects \cite{yin2023surveyMMLM}. They leverage the extensive scale and integrated knowledge of billion-parameter LLMs, a feature that is not available in previous multimodal models. Further, MLLMs adopt novel training paradigms, such as multimodal instruction tuning, to maximize their capabilities \cite{Brown2020language}. Simultaneous use of advanced LLM architectures with innovative training methodologies enables MLLMs to achieve enhanced performance. Some of the major advantages of MLLMs include but not limited to enhanced capabilities, improved reasoning abilities, and increased interactivity, e.g.,  MLLMs can complete tasks such as  generating website code from images and solving mathematical problems \cite{Wei2021zero_shot, Zhu2023minigpt4, Driess2023palme}. Yuan et al. \cite{yuan2025survey} presented a survey on multimodal machine learning, which integrates various modalities, e.g., vision, audio, and text, to enhance the ability of AI systems  to process and understand complex data. The study categorizes approaches into audio-visual, text-visual, touch-visual, and depth-visual learning, reviewing recent advancements, key challenges such as data alignment and fusion strategies, and applications in computer vision, robotics, and human-computer interaction \cite{yuan2025survey}. 

Although there have been several advancements in MLLMs,  there are still some major challenges that needs to be addressed. One of the major issues is the limited ability of MLLMs to handle extended sequences of multimodal data. This  slows down  the  development  of models that are capable of processing long videos or complex datasets containing multimodal media \cite{Driess2023palme}. Further, the instruction-following capabilities of MLLMs requires further research compared to those of closed-source models such as GPT-4V \cite{yin2023surveyMMLM}. 
 GPT-4 with vision (GPT-4V) expands GPT-4’s capabilities by allowing it to process and analyze image inputs from users, marking a significant step in multimodal AI \cite{openai2023gpt4vsystemcard}.

\subsection{A Brief Overview of the LLM Pipeline}
\begin{figure}
  \centering
  \includegraphics[width=0.98\linewidth]{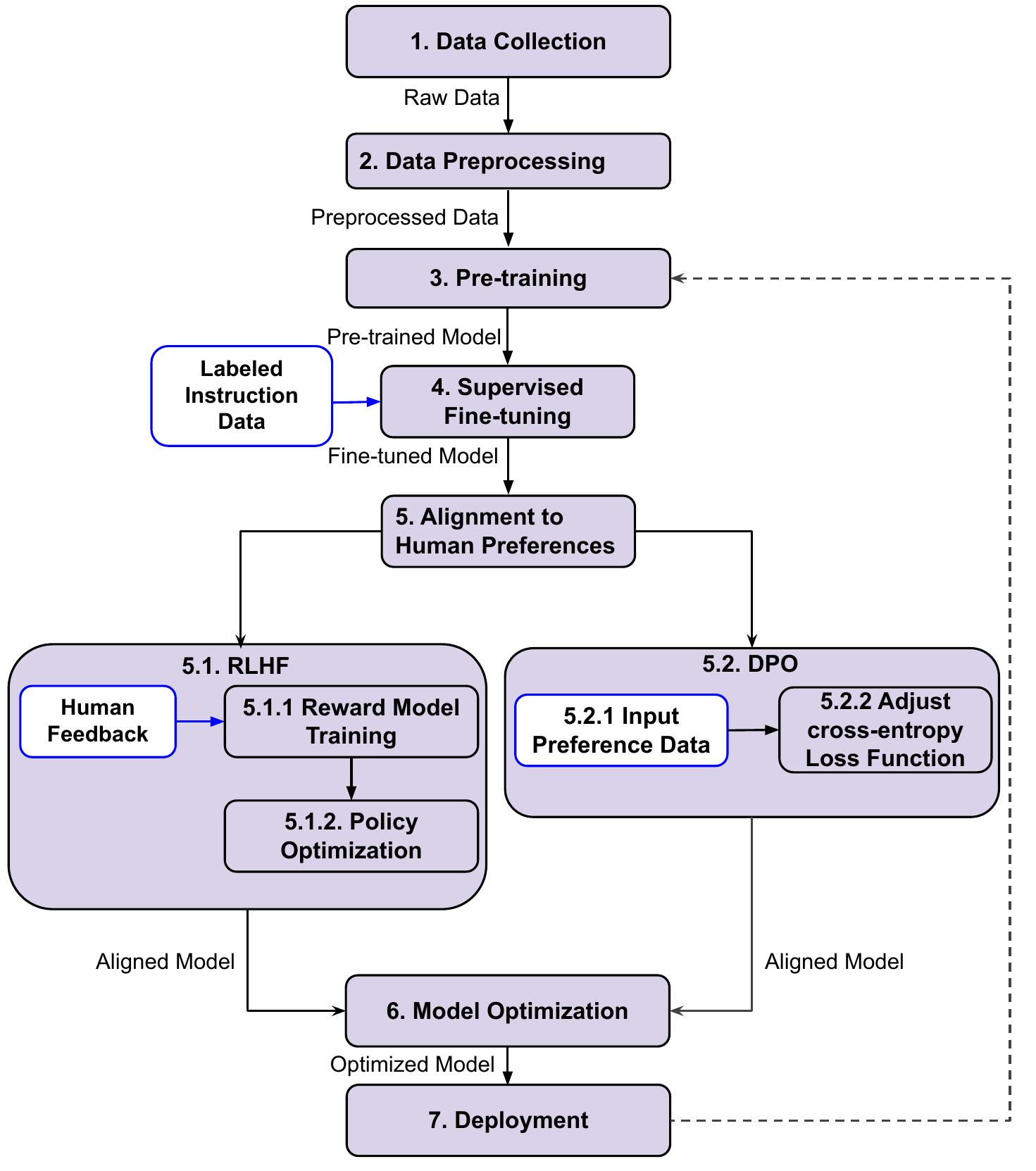} 
  \caption{Overview of the LLM Pipeline.}
  \label{llmpip}  
\end{figure}

We first provide a high-level overview of the LLM pipeline. This will serve as a basis for future discussions on MLLMs as well as potential decentralization strategies based on these stages. Figure \ref{llmpip} represents a flowchart summarizing these stages that are describing in the following.

\underline{1. Data Collection:} The first step  of training LLMs is the collection of extensive datasets, including text from a variety of sources \cite{yin2023surveyMMLM, Brown2020language, LLMSurvey1zhao2023survey}. High quality and diverse datasets allow the model to capture a wide range of language styles, topics, and contexts, which are necessary for robust performance across different tasks. Curating, cleaning, and preparing these large datasets require significant resources and numerous considerations \cite{Brown2020language, LLMSurvey1zhao2023survey}.  Further, LLMs themselves can help generate synthetic datasets through techniques such as prompt engineering and multi-step generation, which have been shown to enhance the training process \cite{tang2023does, long2024llms}. These methods provide a practical way to increase the utility and applicability of training datasets, thereby improving the overall robustness of language models \cite{li2023synthetic, guo2024generative}. A distributed device approach for synthetic data generation remains an open question, particularly in the development of techniques to incrementally enhance data diversity and complexity. This data will be then preprocessed in the next step of the LLM pipeline.

\underline{2. Data Preprocessing:} This step preprocesses the collected data from Step 1 to prepare the raw data for efficient pre-training. This step involves multiple sub-tasks including but not limited to:  
 \textit{Data Deduplication} to remove redundant data, minimize volume, and prevent model overfitting; 
\textit{Data Cleaning} to filter out irrelevant or low-quality data and enhance the quality of dataset;  \textit{Normalization} to standardizes text formats, such as lowercase conversion and removing extra whitespace; \textit{Data Augmentation}  to expand data by creating synthetic samples for a wide range of input structures; 
M\textit{asking Sensitive Information} to anonymize personal data and ensure privacy; \textit{Tokenization} to divide text into tokens for efficient model processing; and
\textit{Shuffling and Balancing} to ensures data diversity and prevents model bias.

\underline{3. Pre-training:} LLMs learn language representations using self-supervised learning objectives \cite{Brown2020language, yin2023surveyMMLM, LLMSurvey1zhao2023survey}. This stage needs extensive computational power and can benefit from distributed training methods, such as model parallelism and data parallelism, to handle the large data size and computational complexity effectively \cite{Brown2020language, HLLM7vaswani2017attention}. An evaluation step (\underline{Evaluation 1}) assesses the pre-trained model's generalization capability on unseen data using  metrics such as  perplexity and F1 score to identify strengths and enhance further training tasks \cite{LLMSurvey4chang2024survey}.

 \underline{4. Supervised Fine-tuning:} The model is fine-tuned on smaller task-specific datasets to tailor it for specific use-cases/applications \cite{yin2023surveyMMLM, Brown2020language, Wei2021zero_shot, Zhu2023minigpt4}. This step improves task-specific performance and can utilize Parameter-Efficient Fine-Tuning (PEFT) methods such as  low-rank adaptation (LoRA) to reduce computational burden while enhancing accuracy \cite{Zhu2023minigpt4, LLMSurvey7zhu2023large}. An evaluation step (\underline{Evaluation 2}) validates the performance of the fine-tuned model on the specific  tasks based on the   metrics that are appropriate for the  underlying  applications \cite{LLMSurvey4chang2024survey}.

\underline{5. Alignment to Human Preferences}:   Two  methods are commonly deployed for achieving this alignment are Reinforcement Learning from Human Feedback (RLHF) and Direct Preference Optimization (DPO). \textbf{i) RLHF} aligns the model with human preferences by leveraging the human feedback into the training process \cite{ouyang2022training, christiano2017deep}. RLHF involves two major key sub-tasks:  \textit{5.1 Reward Model Training:} A reward model is trained to predict human preference scores based on collected human feedback \cite{ouyang2022training}. Human evaluators provide feedback on model outputs, and the reward model learns to predict these human preferences.  \textit{5.2 Policy Optimization:} Using the reward model (which is aligned with the human feedback), the LLM policy is fine-tuned by using reinforcement learning  techniques, such as Proximal Policy Optimization (PPO) \cite{schulman2017proximal}, to generate outputs that are aligned with human preferences \cite{christiano2017deep}. This process adjusts the model to generate more realistic and accurate responses.  \textbf{ii) DPO} is an efficient method for aligning LLMs with human preferences. Offering a simpler alternative to RLHF, DPO eliminates the need for a reward model by directly utilizing paired human-labeled preference data (as shown in Step 5.2.1: Input Preference Data). This approach adjusts the cross-entropy loss function (Step 5.2.2: Adjust Cross-Entropy Loss Function) to optimize model outputs through comparisons, leveraging maximum likelihood estimation to reduce both complexity and computational demands \cite{rafailov2024direct, superannotate_dpo}. Models such as LLaMA 3.1 have successfully employed DPO which paves the way for  its wide deployment for  LLM alignment \cite{dubey2024llama}. By streamlining the fine-tuning process while maintaining efficiency and interpretability, DPO has become a powerful and popular method for improving the capabilities of state-of-the-art LLMs.\\
    An evaluation step (\underline{Evaluation 3}) assesses the performance of the aligned model  in terms of compliance with human preferences and ethical guidelines \cite{LLMSurvey6zhao2024explainability}. This can involve both qualitative and quantitative metrics.
    
 \underline{6. Model Optimization:} Post-training optimization techniques are applied to enhance the model's efficiency and performance. This may include model compression \cite{han2015deep, ganesh2021compressing}, quantization \cite{shen2020q, bai2020binarybert}, and distillation \cite{sanh2019distilbert, jiao2020tinybert} to reduce computational requirements and improve inference speed without notably degrading the performance \cite{LLMSurvey1zhao2023survey, LLMSurvey4chang2024survey}.

\underline{7. Deployment:} The final model is deployed into real-world applications. Key considerations for this step include scalability, latency, and integration with existing systems \cite{LLMSurvey1zhao2023survey}. Frequent monitoring and maintenance are essential to ensure robust performance of the model. Generated data, such as user interactions, feedback, or performance metrics, can be utilized to enhance or refine the pre-training process. This establishes a feedback loop that continuously improves the model's capabilities over time, allowing it to adapt and perform better with new datasets.

\subsection{Differences of LLM and MLLM pipelines}
Here, we elaborate on how MLLMs are different than LLMs at each of the mentioned stages.

{\underline{1. Data Collection:}}
 In MLLMs, data collection includes  multimodal datasets that are no longer limited to text, and can include  images, audio, and/or videos \cite{yin2023surveyMMLM, alayrac2022flamingo}. This allows MLLMs to learn cross-modal representations and capture relations among multiple data modalities. Ultimately, this can lead to new capabilities such as  image captioning and visual question answering that are not achievable using  standard LLMs that only rely on text data.

{\underline{2. Data Preprocessing:}} 
As MLLMs need preprocessing of multiple data modalities \cite{yin2023surveyMMLM}, they need to conduct more tasks such as image normalization, audio feature extraction, and aligning textual data with corresponding visual or auditory content during this step as compared with LLMs. These additional steps  are not required in the text-only preprocessing pipeline of standard LLMs and allow for synchronizing multimodal data  for  MLLMs.

{\underline{3. Pre-training:}} 
MLLMs are pre-trained using objectives that utilize multimodal data to learn cross-modality representations  \cite{Driess2023palme, Zhu2023minigpt4, li2023blip2}. While  standard LLMs trained only on text, MLLMs utilize multimodal contrastive learning and other cross-modal objectives to capture relationships between text and other data types and improve their performance in terms of understanding and generating multimodal content \cite{alayrac2022flamingo}.

{\underline{4. Supervised Fine-tuning:}} 
Fine-tuning MLLMs involves adapting the model to specific multimodal tasks using datasets with multimodal annotations \cite{yin2023surveyMMLM}. Here, the main difference of MLLMs with  standard LLMs is its requirement of more modalities of datasets and training procedures for tasks such as image captioning, visual question answering, or speech recognition.

{\underline{5. Alignment to Human Preferences:}} 
5.1. In MLLMs, RLHF incorporates human feedback on multimodal outputs, such as evaluating the relevance of generated captions to images or the accuracy of responses in visual dialogues \cite{liu2023llava}. Designing reward models that can assess multimodal content adds complexity compared to the text-only feedback in standard LLMs. This required more complex methods  to deal with the cross-modal evaluation.
5.2. DPO also faces some challenges when applied to MLLMs (as compared to LLMs), as it must capture multimodal preference data rather than focusing only on text-based comparisons. In MLLMs, DPO uses paired human-labeled preference data across multiple modalities (e.g., text as well as images or audio) and directly optimizes the cross-entropy loss function to align the model's outputs with human preferences. This requires additional complexity in processing and comparing multimodal data, as the preference signal includes diverse input-output combinations (e.g., evaluating whether a caption matches an image or whether an audio explanation corresponds to a visual input). Recent implementations of DPO in MLLMs, such as Flamingo \cite{ 110vlmalayrac2022flamingo} and LLaVA \cite{ liu2023llava}, have demonstrated its ability to achieve efficient multimodal alignment.

{\underline{6. Model Optimization:}} 
Optimizing MLLMs is  computationally more expensive  compared with standard LLMs due to the need for analyzing and optimizing  additional modalities \cite{alayrac2022flamingo}. In order to deal with these challenges, there is a need for   modality-specific compression, efficient cross-modal attention mechanisms, and shared parameter architectures to reduce the computational overhead \cite{gafni2022make}.

{\underline{7. Specialized Encoders and Integration Phases:}} 
 A key distinction between LLMS and MLLMs lies in their use of specialized encoders and integration phases. While LLMs rely on a unified architecture for processing textual data, MLLMs require separate encoders for different modalities, such as speech and images. These encoders are trained individually to optimize for their specific input types and are later integrated with the core language model during an adaptation phase. This approach allows MLLMs to align diverse feature spaces effectively and process multimodal inputs effectively \cite{dubey2024llama}. The modular nature of MLLM pipelines makes them uniquely suited for tasks involving a variety of input modalities, setting them apart from the streamlined design of LLMs.

{\underline{8. Deployment:}}
Efficient deployment of MLLMs needs extensive infrastructures with the ability to manage the multimodal inputs and outputs \cite{yin2023surveyMMLM}. This includes processing various modalities of data, e.g., images, audio, and video, as well as ensuring efficient integration and scalability in real-world applications. Similar to previous step, these complexities lead to increased computational load and latency while  processing multimodal data.

\section{Decentralizing the LLM Pipeline}
\label{Decentral_llm_View}
\subsection{Overview of FL}

In \cite{imteaj2021survey}, we provided a comprehensive survey of FL algorithms tailored for resource-constrained IoT devices. The concept of FL, initially proposed by McMahan et al. \cite{mcmahan2016communication}, can also be extended to distributed LLMs for addressing challenges in data privacy, computational resource optimization, and scalability.

The FL process includes the following primary steps:

\noindent\emph{\textbf{Step 1 (Initiate training task and global model)}}: In the initial phase, the central server identifies the task requirements and the target LLM application, such as text generation, summarization, or language translation. A global LLM ($W_G^0$) is initialized, which could be a foundational pre-trained model. The server then broadcasts this global model to selected local clients (e.g., edge devices or distributed data centers), referred to as participants.

\noindent\emph{\textbf{Step 2 (Local model update)}}: Each participant fine-tunes the global LLM using their locally available data. Upon receiving the global model $W_G^t$ (where $t$ represents the $t$-th iteration), each client $k$ updates its model parameters $W_k^t$ by optimizing a local objective function $F_k(W_k^t)$. This local fine-tuning process ensures that the model adapts to client-specific data while preserving privacy. The updated local models are then shared with the central server.

\noindent\emph{\textbf{Step 3 (Global aggregation)}}: After collecting the locally fine-tuned models, the server aggregates them to produce an updated global model ($W_G^{t+1}$). The aggregation process  involves methods such as  weighted averaging while considering the size and quality of data at each client. The updated global LLM is then redistributed to the clients for the next round of training.

Steps 2 and 3 are iteratively repeated until the global model converges by minimizing the overall objective function $F(W_G^t)$. The optimization objective for distributed LLMs can be expressed as \cite{li2018federated}:

$$
\min _{w} f(w) = \sum_{k=1}^{N} P_{k} F_{k}(w)
$$

\noindent where \( N \) represents the total number of participating clients, \( P_k (\geq 0) \) indicates the relative contribution of client \( k \) while satisfying \( \sum_{k} P_k = 1 \), and \( F_k(w) \) is the expected loss for client \( k \) on parameter \( w \). If each client \( k \) has \( n_k \) data samples (and \( n = \sum_{k} n_k \)), the relative weight \( P_k \) can be expressed as \( P_k = \frac{n_k}{n} \).

This adaptation enables distributed LLMs to leverage local data while preserving privacy, reducing communication overhead, and ensuring resource-efficient training.

\begin{figure*}[h]
  \centering
  \includegraphics[width=0.8\linewidth]{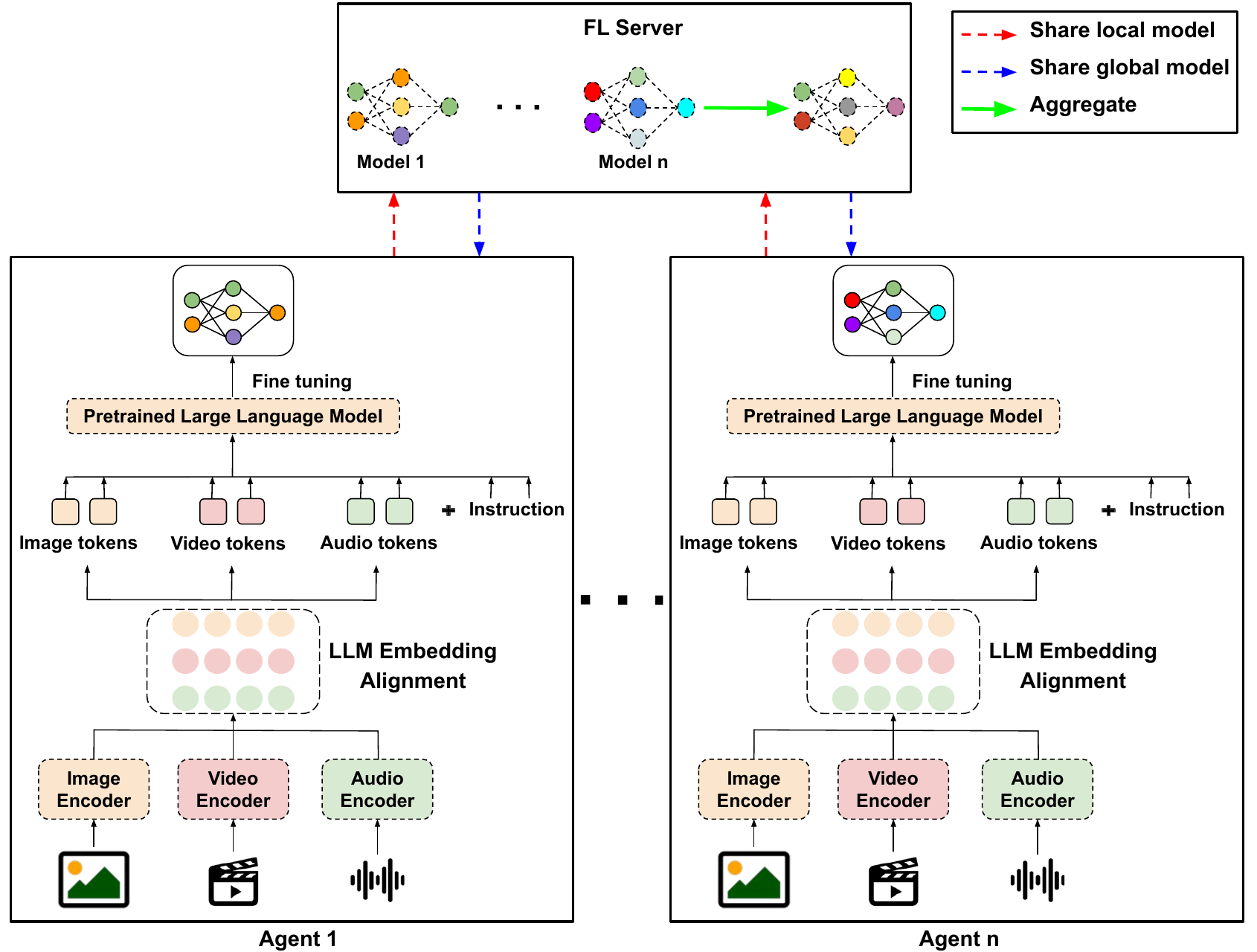} 
  \caption{A Framework for Federated MLLMs (partly inspired by \cite{lyu2023macaw}).}
  \label{fig:1}  
\end{figure*}
\subsection{Decentralization at Different Steps of the MLLM Pipeline} 
Here, we explore how each step of the MLLM pipeline can be decentralized using methods such as FL, distributed optimization, and distributed hyperparameter tuning. We discuss potential decentralization strategies for each step. Table \ref{decben} represents a brief summary of how each MLLM stage can benefit from decentralization.  
{ Figure \ref{fig:1}  represents a federated architecture for multimodal data processing, integrating modality-specific encoders, alignment mechanisms, and a cognitive module to enable collaborative model fine-tuning across agents. Some parts of this Figure is inspired by an ``an overview of MACAW-LLM model architecture'' proposed by Lyu et al. \cite{lyu2023macaw}. Each agent in this framework leverages modality encoders to process diverse input types, such as images, videos, and audio. Specifically, a visual modality encoder, based on CLIP (trained with extensive supervision on textual and visual data)\cite{ 108vlmradford2021learning}, is used to encode both image and video data. For audio data, WHISPER, a multilingual speech recognition model trained on a large-scale audio dataset \cite{radford2023robust}, is utilized to extract meaningful audio features. These modality-specific encoders independently generate feature-rich token representations for their respective modalities, which are then prepared for integration with the global model. 
Since modality encoders are trained separately, their output representations  may lack compatibility. To address this, an alignment module is employed to unify these features into a shared embedding space. This process involves transforming the encoded features using 1D convolutional layers for compression and linear layers for dimensional alignment. The unified features are then processed using multi-head self-attention mechanisms to align them with the textual embedding space of the pretrained LLM, such as LLAMA, which serves as the cognitive module \cite{ LLMH12touvron2023llama}. The aligned multimodal representations are concatenated with textual embeddings to create a multimodal instruction representation. Locally trained models at each agent are sent to the FL server for aggregation, which produces a new global model. This iterative process preserves data privacy, ensures the effective integration of multimodal information, and maintains a robust collaborative training mechanism across all agents in the FL environment.} 

\begin{table*}

\centering
\caption{Decentralization Benefits Across MLLM Pipeline Stages (S1: Data Collection; S2: Data Preprocessing; S3: Pre-training; S4: Supervised Fine-tuning); S5: RLHF/DPO; S6: Model Optimization; S7: Deployment)}
\begin{tabular}{|l|p{15cm}|}
\hline
\textbf{Stage} & \textbf{Benefits of Decentralization} \\
\hline
\textbf{S1} & 
\begin{itemize}
\item Enables distributed sourcing and storage of multimodal data, which can  reduce central storage demands and mitigate single-point failures.
\item Enhances privacy by keeping sensitive data on local devices, which can  reduce the risk of data leakage and facilitate compliance with data protection regulations.
\item Reduces bandwidth usage by eliminating the need to transfer large datasets to a central server.
\item Promotes data diversity by aggregating data from diverse sources, which can  strengthen model generalization across varied data distributions.
\item Ensures data integrity through peer-to-peer networks and distributed data marketplaces.
\end{itemize} \\
\hline
\textbf{S2} & 
\begin{itemize}
\item Distributes computational load by performing preprocessing tasks (e.g., cleaning, normalization, augmentation, tokenization) locally.
\item Minimizes raw data transfer to central servers, which can  preserve privacy and reduce communication overhead.
\item Enhances scalability and efficiency via parallel processing across decentralized computing nodes.
\item Leverages distributed computing frameworks  adapted for multimodal data.
\item Allows for local anonymization before sharing, which can contribute to privacy preservation.
\end{itemize} \\
\hline
\textbf{S3} & 
\begin{itemize}
\item Utilizes distributed resources to handle extensive computational requirements of large-scale multimodal datasets.
\item Balances workload using data and model parallelism to alleviate memory bottlenecks.
\item Optimizes computation and communication, which can  
 reduce latency by processing data closer to its source.
\item Enhances scalability with minimal central resource needs by leveraging edge computing resources.
\item Employs advanced distributed training techniques to improve efficiency.
\end{itemize} \\
\hline
\textbf{S4} & 
\begin{itemize}
\item Enables federated fine-tuning, which allows   models to adapt locally without sharing raw data and preserve privacy.
\item Lowers bandwidth demands by transmitting only model updates or subsets of parameters.
\item Enhances generalization across unique client data by addressing data heterogeneity.
\item Supports personalization by enabling local fine-tuning on edge devices to meet specific user needs.
\item Manages computational burden locally, which reduces reliance on central servers and supporting low-latency adaptation.
\end{itemize} \\
\hline
\textbf{S5} & 
\begin{itemize}
\item Preserves privacy by collecting human feedback locally without accessing sensitive user data.
\item Decentralizes reward model training, reducing communication overhead.
\item Addresses data heterogeneity among clients that can enhance model adaptation to diverse user preferences.
\item Decentralizes policy optimization using distributed reinforcement learning, enabling local policy updates.
\item Enhances scalability and stabilizes training.
\end{itemize} \\
\hline
\textbf{S6} & 
\begin{itemize}
\item Handles resource-constrained devices efficiently via model compression, quantization, and distillation.
\item Enables collaborative optimization without sharing raw data.
\item Saves bandwidth by sharing only optimized models or parameters across nodes.
\item Considers diverse computational capacities, enhancing deployment feasibility across heterogeneous devices.
\item Preserves privacy while optimizing models collaboratively.
\end{itemize} \\
\hline
\textbf{S7} & 
\begin{itemize}
\item Leverages edge and server resources through distributed deployment, which can improve scalability and robustness.
\item Minimizes latency by enabling real-time inference on edge devices close to end-users.
\item Distributes computational load via model partitioning and collaborative inference among devices.
\item Reduces central server dependency through load balancing among devices.
\item Preserves data privacy by performing local inference, that can reduce bandwidth requirements.
\item Supports scalable deployment across networks with varying device capabilities and constraints.
\end{itemize} \\
\hline
\end{tabular}
\label{decben}
\end{table*}

\begin{figure*}
  \centering
  \includegraphics[width=0.95 \textwidth]{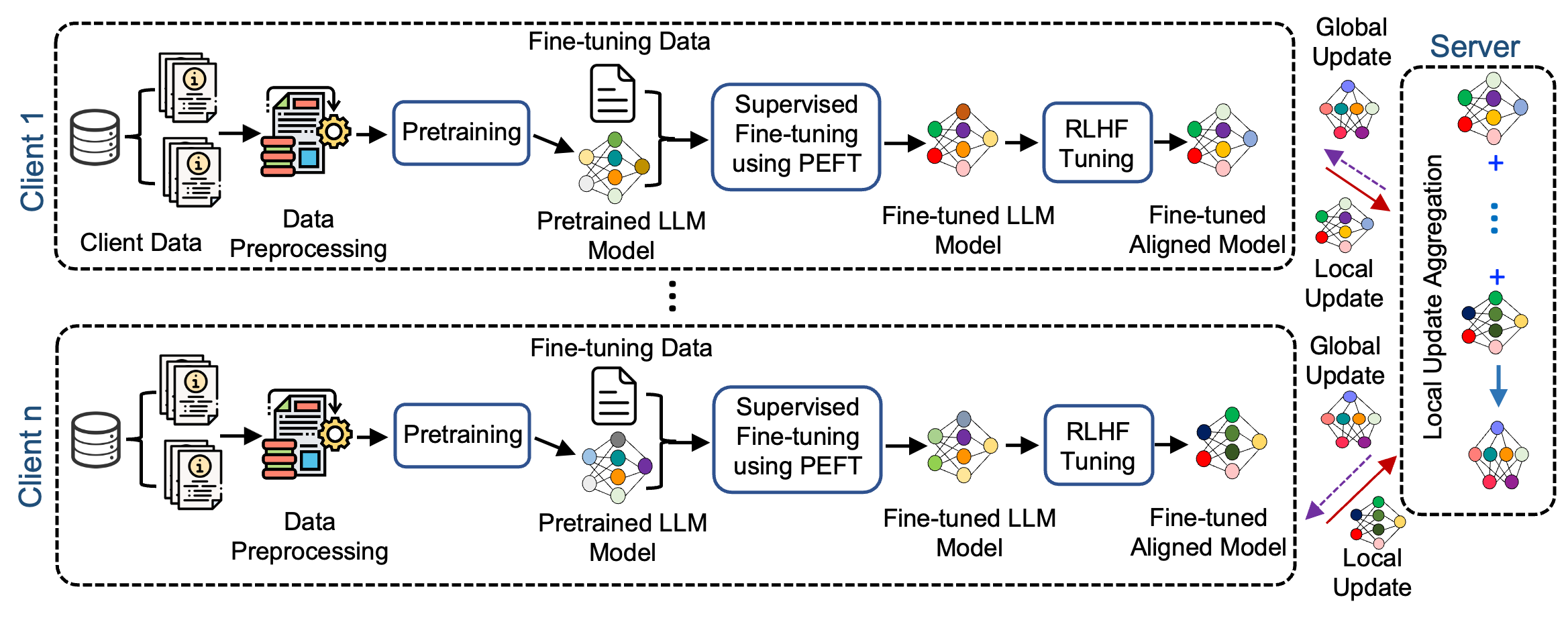} 
  \caption{Overview of Federated LLM Pipeline.}
  \label{dllmpip}  
\end{figure*}

Figure \ref{dllmpip} provides a comprehensive overview of the federated LLM pipeline. This process includes multiple clients. At each edge client, data is initially  preprocessed to ensure compatibility and quality for training purposes. Then, a pretrained LLM is utilized as the foundation for supervised fine-tuning using PEFT. The generated fine-tuned model is then refined using RLHF  to create an aligned model that is optimized for real-world application scenarios. This  process is locally performed by multiple clients to ensure that their specific data characteristics are captured without the need to share their raw data. After the local updates, they are aggregated at a central server. The aggregated model updates are then disseminated again to clients for further iterations. This pipeline can effectively combines the advantages of FL with the adaptability of LLM fine-tuning to address various client-specific and global challenges.

\subsubsection{Data Collection}
Data collection process for MLLMs can be decentralized  through distributed systems and collaborative data sourcing techniques. Distributed devices across different locations, such as edge devices and IoT sensors, can be leveraged to analyze and use diverse multimodal data such as text, images, audio, and video without a need to centrally store all date \cite{21ye2024openfedllmLLM, 9qu2024mobileLLM}. This not only enhances the diversity of the dataset but also partly addresses privacy concerns by keeping potentially sensitive data on local devices.

FL frameworks  pave the way for decentralized data collection by allowing for several clients to contribute to the model training without a need to share raw data \cite{15kuang2024federatedscopeLLM, 18yao2024federatedLLM}. In the context of MLLMs, FL can be extended to support multimodal data by coordinating the collection and preprocessing of different data modalities across clients \cite{98li2024mllm}. This distributed structure can  reduce the risk of data leakage and even in some scenarios help to   comply with data protection regulations. Further, distributed data crawling joint with distributed data marketplaces can enable the aggregation of large-scale multimodal datasets from diverse sources. Peer-to-peer networks can  also be employed to enable distributed data sharing and to ensure data integrity in a decentralized manner. Ye et al \cite{21ye2024openfedllmLLM} proposed a framework for training large language models on decentralized private data via FL, referred to as OpenFedLLM.  This approach benefits from keeping local data on user devices without transferring it to a central server \cite{21ye2024openfedllmLLM}. OpenFedLLM  also uses techniques to manage challenges in communication efficiency, model heterogeneity, and scalability, making FL an effective method for developing LLMs that respect user privacy across distributed data sources \cite{21ye2024openfedllmLLM}. 
Further, Chen et al. \cite{38chen2023federated} explored federated LLMs as a way to leverage distributed data sources efficiently.

\subsubsection{Data Preprocessing}

Decentralizing the data preprocessing step for MLLMs requires distributing the preprocessing tasks over multiple clients. This can significantly reduce computational overhead and enhance scalability. For instance, authors in \cite{jin2024decoagent} developed a decentralized solution that enables  collaboration between LLM-empowered agents; this framework can also benefit from decentralized preprocessing while leveraging smart contracts. The preprocessing steps such as data cleaning, normalization, augmentation, and tokenization can be performed locally where the data is stored. This minimizes the need to transfer large volumes of raw multimodal data to a central server and can ultimately   preserve data privacy and reduce communication overhead. Existing methods such as  federated data preprocessing enables local  preprocessing of the distributed datasets at the client side. Each client then  shares only the necessary metadata or processed outputs with the central server or other clients \cite{21ye2024openfedllmLLM}. This is  also contributing to more privacy-preserving   handling of sensitive information, as personal data can be anonymized  locally before sharing the data.

Further, distributed computing  frameworks such as MapReduce \cite{dean2008mapreduce} and Apache Spark \cite{zaharia2010spark} can be adapted for multimodal data preprocessing tasks and enabling parallel processing of large datasets across decentralized computing nodes. For instance, MapReduce paradigms can be employed to efficiently handle tasks such as feature extraction from images or audio files in a distributed fashion.  They allow for distributed  preprocessing of large datasets across multiple clients, which is crucial for preparing the data for training LLMs/MLLMs.  Although these frameworks can be considered  for more efficient preprocessing, LLM-specific data preprocessing  involves tasks that might be beyond their standard capabilities, such as  tokenization, embeddings, and complex language-specific cleaning.

\subsubsection{Pre-training}

 As extensive computational resources are required for training on large-scale multimodal datasets, decentralizing the pre-training process within the LLM/MLLM pipeline is essential for efficient processing. Distributed training techniques, such as data parallelism and model parallelism, allow the workload to be distributes across multiple computing nodes. Data parallelism enables simultaneous processing of multiple data batches across nodes, while model parallelism divides the model itself—partitioning layers or operations across devices to alleviate memory bottlenecks. As elaborately discussed in \cite{1zeng2023distributedLLM, 5brakel2024modelLLM}, this approach addresses the scalability and latency challenges in training massive multimodal models by optimizing both computation and communication within the infrastructure.

Advanced distributed training frameworks such as Horovod and DeepSpeed can be utilized to optimize communication and computation efficiency during pre-training \cite{sergeev2018horovod, gibiansky2017bringing, rasley2020deepspeed}. Techniques such as gradient compression, asynchronous updates, and communication overlap can further enhance the scalability of distributed pre-training. For instance, the ACCO method \cite{4nabli2024accoLLM} enables this by using gradient accumulation during communication rounds and reducing the frequency of gradient synchronization  to minimize communication overhead.  ACCO’s asynchronous update mechanism allows for  computations to proceed without waiting for global synchronization, making it adaptable to heterogeneous hardware. The use of communication overlap enables simultaneous gradient computation and data transfer. This ultimately handles communication latency and improves overall throughput which is crucial in large-scale multimodal training tasks \cite{4nabli2024accoLLM}. 

Edge computing resources can also be leveraged to distribute the pre-training process closer to data sources, reducing latency and bandwidth usage \cite{9qu2024mobileLLM}. By utilizing available computational capacities at the  edge, it is possible to scale the pre-training of MLLMs without relying solely on centralized data centers. In this context, FL can be used to  train a global MLLM without sharing local data to protect data privacy and ensure that the model leverages diverse data distributions from different clients. It further enhances the generalizability and robustness of the multimodal model \cite{18yao2024federatedLLM}.

{\color{black}Mamba \cite{gu2023mamba} introduces selective state space models (SSMs) as a highly efficient alternative to Transformers for long-sequence modeling. By dynamically parameterizing SSM dynamics based on input data, Mamba selectively propagates or filters information, achieving linear scaling with sequence length. This approach significantly reduces memory requirements while enhancing throughput, making it five times faster than Transformers. Its performance across diverse modalities such as language, audio, and genomics demonstrates its adaptability and efficiency for large-scale pretraining tasks.

LoLCATs \cite{zhang2024lolcats} and BitNet \cite{wang2023bitnet} take unique approaches to address the computational challenges of pretraining large models. LoLCATs replaces quadratic attention mechanisms in Transformers with subquadratic linear attention using a two-step process of attention transfer and LoRA. This method improves scalability and reduces memory and compute overhead, enabling the efficient pretraining of models with up to 405 billion parameters. Meanwhile, BitNet introduces a 1-bit Transformer architecture, utilizing the BitLinear module to train models with 1-bit precision, drastically reducing memory and energy consumption without compromising performance. Similarly, Srinivasan \textit{et al.} \cite{srinivasan2023training} leverage sparsity techniques and dataflow execution to optimize pretraining for GPT-13B, achieving a 4.5x speedup through sparse weights and advanced kernel fusion. SWARM Parallelism \cite{ryabinin2023swarm}, a decentralized training algorithm, facilitates the pretraining of large models across unreliable and heterogeneous devices. By employing randomized pipelines that dynamically rebalance, it reduces communication overhead and ensures efficient large-scale training in cost-effective environments.}

\subsubsection{Supervised Fine-tuning}
Decentralizing the supervised fine-tuning of MLLMs can be enabled by leveraging federated fine-tuning methods. In these methods, clients fine-tune the global model on their local task-specific multimodal datasets and share the model updates rather than the  raw data  \cite{15kuang2024federatedscopeLLM, 20shu2024ferretLLM}. This approach preserves data privacy and leverages the diversity of data across clients to improve model generalization. 
In \cite{20shu2024ferretLLM}, authors introduce \textit{Ferret}, a federated full-parameter tuning framework for LLMs. Further, \cite{15kuang2024federatedscopeLLM} presents \textit{FederatedScope-LLM}, a comprehensive package for fine-tuning LLMs in FL environments.

PEFT techniques such as LoRA \cite{hu2021lora} can be adapted for distributed settings to reduce the communication overhead during model updates \cite{16xu2023fwdllmLLM}. By updating only a small subset of model parameters or using adapter modules, clients can significantly decrease the amount of data that needs to be transmitted. This  makes  federated fine-tuning more practical in scenarios that with limited bandwidth.

Edge devices equipped with sufficient computational resources can perform local fine-tuning of the MLLM to adapt it to specific user needs or contexts \cite{26wu2024fedbiotLLM}. This localized fine-tuning enables personalization while maintaining user privacy, as the data remains on the device. Further, quantization techniques can be employed in a distributed manner to fine-tune smaller models, using the outputs of a larger pre-trained MLLM. They can reduce both model size and computational requirements. This approach is particularly helpful for deploying MLLMs on devices with limited resources.

FL with Heterogeneous Low-Rank Adaptation (FLoRA) is a federated fine-tuning framework for LLMs that supports decentralized, privacy-preserving model adaptation across clients with diverse computational capacities \cite{97wang2024flora}. FLoRA  tackles  aggregation noise and heterogeneous data distribution challenges by using a cumulative stacking of low-rank adaptation (LoRA) \cite{hu2021lora} modules  to ensure efficient and accurate fine-tuning in decentralized settings \cite{97wang2024flora}.

{\color{black}ZeRO-Offload \cite{ren2021zero} enables efficient fine-tuning of large models by offloading gradients, optimizer states, and computations to CPUs while retaining forward and backward passes on GPUs. This approach significantly reduces GPU memory usage, enabling the fine-tuning of models with up to 13 billion parameters on a single GPU and scaling efficiently across multiple GPUs for models with over 70 billion parameters. PETALS \cite{borzunov2022petals} complements this by facilitating collaborative fine-tuning and inference through distributed server-hosted model layers, allowing users to pool resources dynamically. By supporting parameter-efficient techniques such as prompt tuning and adapter training, PETALS enables flexible adaptation of large models such as BLOOM-176B, making fine-tuning accessible even in resource-constrained settings.}

\subsubsection{RLHF}
Decentralizing RLHF for MLLMs requires collecting human feedback from distributed sources and integrating it into the training process without centralizing sensitive user data \cite{23ye2024safetyLLM}. Federated RLHF allows clients to locally train reward models based on their users' feedback on multimodal outputs and share the model updates with a central server for aggregation \cite{27zhang2024fedpitLLM}. This approach preserves user privacy and accommodates diverse human preferences across different regions or user groups while enabling scalability.
 Decentralizing RLHF needs distributing both the reward model training and the policy optimization across multiple  clients. 

5.1. \textit{Decentralizing Reward Model Training}
The reward model predicts human preference scores based on collected feedback. FL can help decentralizing its  training  \cite{18yao2024federatedLLM,20shu2024ferretLLM}.  As ensuring Communication efficiency is also crucial in this decentralized setup,  methods such as  gradient compression and parameter-efficient fine-tuning (PEFT) (e.g., LoRA) can be deployed to reduce the communication overhead between clients and the central server \cite{66qi2024fdlora,61gao2024fedpt}.  Further, addressing data heterogeneity among clients through personalized FL can ensure the reward model generalizes adequately across different data distributions \cite{97wang2024flora}.

5.2. \textit{Decentralizing Policy Optimization}
Policy optimization in RLHF aims to fine-tune the LLM/MLLM's policy to generate outputs that align with human preferences. Decentralizing this step can be enabled  using distributed RL techniques. Clients perform local policy updates using their data and the shared reward model, then contribute to a global policy through model aggregation \cite{31sani2024futureLLM, 65elbakary2024mira}.

Elbakary et al. \cite{65elbakary2024mira} propose MIRA, a method for federated multi-task learning suitable for LLMs. Their study proposed an approach to reduce the variance of policy gradient estimators by leveraging gradient norm constraints, which can ultimately stabilize training and improve policy learning efficiency. MIRA could contribute to decentralizing policy optimization in RLHF by enabling more reliable gradient updates across distributed clients and reducing the dependency on centralized coordination for consistent policy improvement.
 Distributed Proximal Policy Optimization (PPO) algorithms can be used to enable local policy optimization of clients  while maintaining convergence guarantees. Ling et al. \cite{17ling2024convergenceLLM} analyze the convergence of zeroth-order federated tuning methods, which could be used for optimizing policies without sharing raw data. It can be used to  decentralize policy optimization in RLHF.

\subsubsection{Model Optimization}
Model optimization benefits from different techniques such as  model compression, quantization, and distillation to enhance efficiency. Decentralizing this step allows multiple clients to collaboratively optimize the model while considering the resource constraints and preserving privacy.
 Federated distillation enables clients to distill knowledge from a global model into local models without sharing data. Zhang et al. \cite{27zhang2024fedpitLLM} introduce FewFedPIT, a privacy-preserving and few-shot FedIT approach. FewFedPIT can decentralize the model optimization stage of LLMs by enabling FedIT, which allows multiple clients to optimize the model collaboratively without sharing the data. As FewFedPIT can enable generating synthetic data locally and aggregating model updates through parameter isolation, it reduces reliance on centralized data sharing and reduces communication overhead. Hence, it lends itself as a promising solution to  optimize large models across distributed clients.

\subsubsection{Deployment}
Distributed inference serving enables the model to utilize computational resources from multiple servers or devices during the inference phase \cite{2wu2023fastLLM, 7borzunov2024distributedLLM}. 
Wu et al.  introduced FastServe, a distributed inference serving system for LLMs that improves the deployment stage by enabling efficient decentralized processing \cite{2wu2023fastLLM}. By implementing preemptive scheduling with a novel skip-join Multi-Level Feedback Queue and proactive GPU memory management, FastServe reduces latency and head-of-line blocking, facilitating scalable and responsive LLM deployment across distributed infrastructures \cite{2wu2023fastLLM}.

Edge deployment allows LLMs to run on local devices. It further enables  real-time inference without reliance on central servers. Qu et al. conducted a thorough survey on mobile edge intelligence (MEI) for LLMs \cite{9qu2024mobileLLM}. This paper also discussed the deployment of LLMs on edge devices through MEI, which benefits from the computational resources at the  edge to enhance privacy and reduce latency. By using  model compression, quantization, and partitioning, MEI enables the distribution of LLMs across edge devices \cite{9qu2024mobileLLM}. Techniques such as model partitioning and collaborative inference can distribute the computational load among multiple devices. Li et al. proposed CoLLM, a collaborative inference framework that allows for participation of resource-constrained devices in LLM inference through tensor parallelism \cite{45li2024collm}. By dynamically distributing computational tasks across multiple devices and optimizing load balancing to reduce latency and energy consumption, CoLLM enables decentralized deployment of LLMs \cite{45li2024collm}. It also enables conducting inference on a distributed network of low-power devices without centralizing resources in data centers. A detailed analysis of studies on LLM-based edge intelligence is provided by Friha et al. \cite{83friha2024llm}.

Further, decentralized inference can leverage peer-to-peer networks to distribute inference tasks, enhancing scalability and robustness. Du et al. proposed a framework for distributed multi-modal foundation models (FMs) leveraging 6G networks to enable efficient and decentralized processing across wireless devices \cite{74du2024distributed}. By integrating data and pipeline parallelism and using FL enhanced with over-the-air computation, their proposed method facilitates collaborative training and inference of large FMs and enables  scalable decentralized deployment of LLMs in network environments with various device capabilities and constraints \cite{74du2024distributed}. OpenFedLLM, a framework for training and deploying LLMs on decentralized private data which is proposed by Ye et al.  \cite{21ye2024openfedllmLLM}  allows clients to fine-tune and deploy LLMs across distributed data sources without transferring data to a central server.

\section{Summary, Contributions, Challenges, and Future Directions of  Related  Studies on LLMs}
\label{summaryView}

\begin{figure*}
  \centering
  \includegraphics[width=0.99 \textwidth]{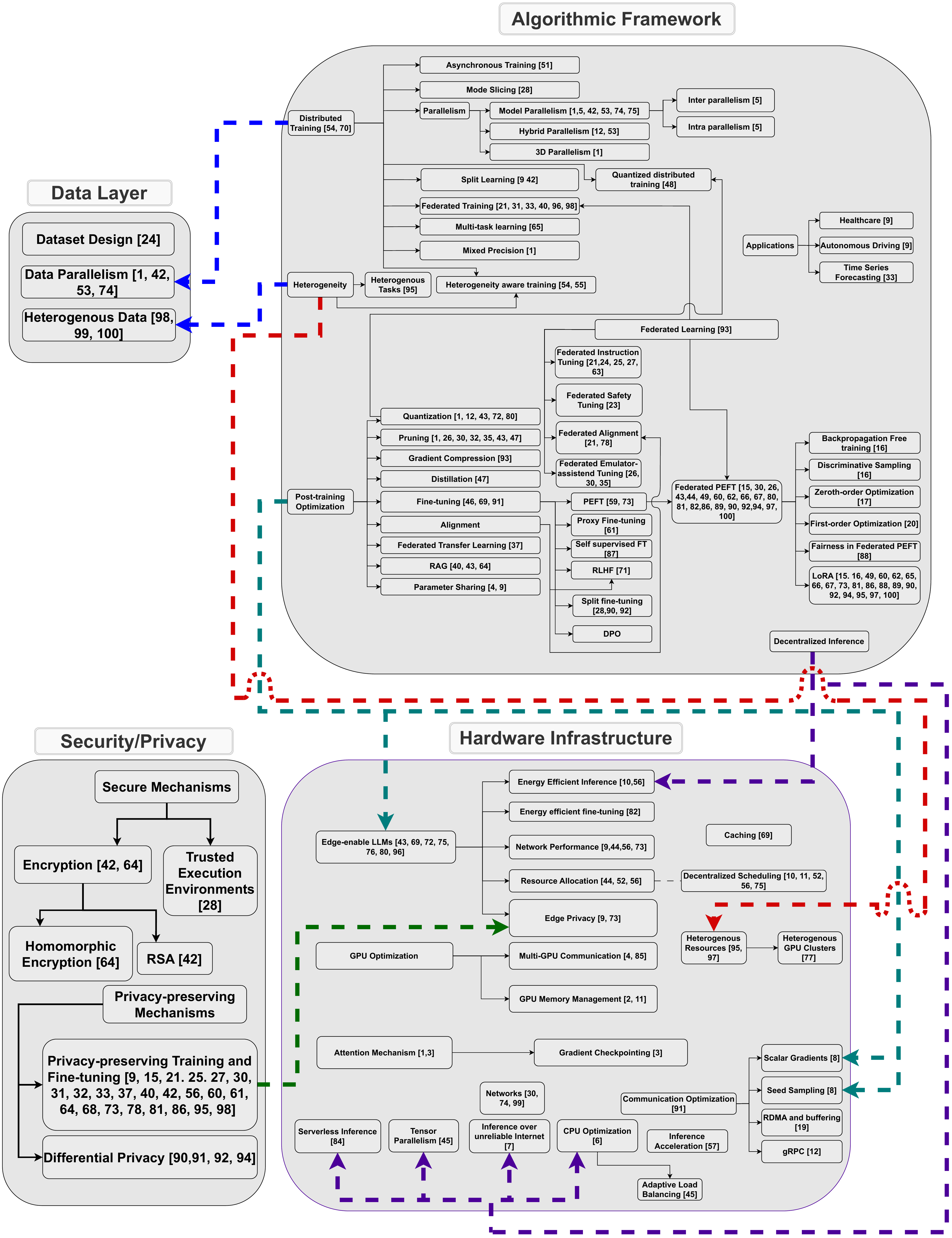} 
  \caption{Overview of key research areas and their connections in Distributed LLM pipelines.}

  \label{figint}  
\end{figure*}

\begin{figure*}
  \centering
  \includegraphics[width=0.99 \textwidth]{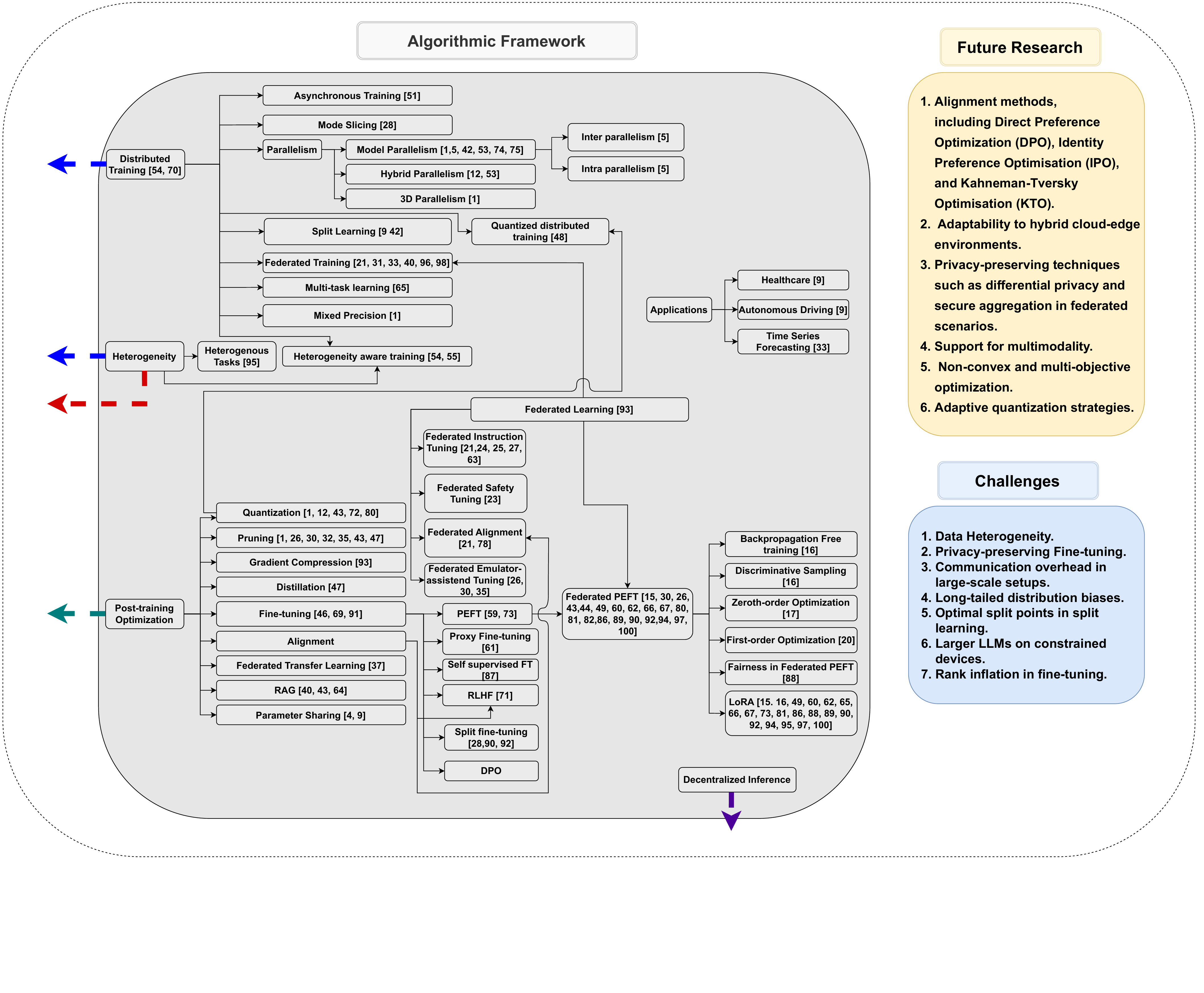} 
  \caption{Algorithmic Framework: pre-training, post-training, and application-focused techniques for DLLMs.}
  \label{figint1}  
\end{figure*}

\begin{figure*}
  \centering
  \includegraphics[width=0.99 \textwidth]{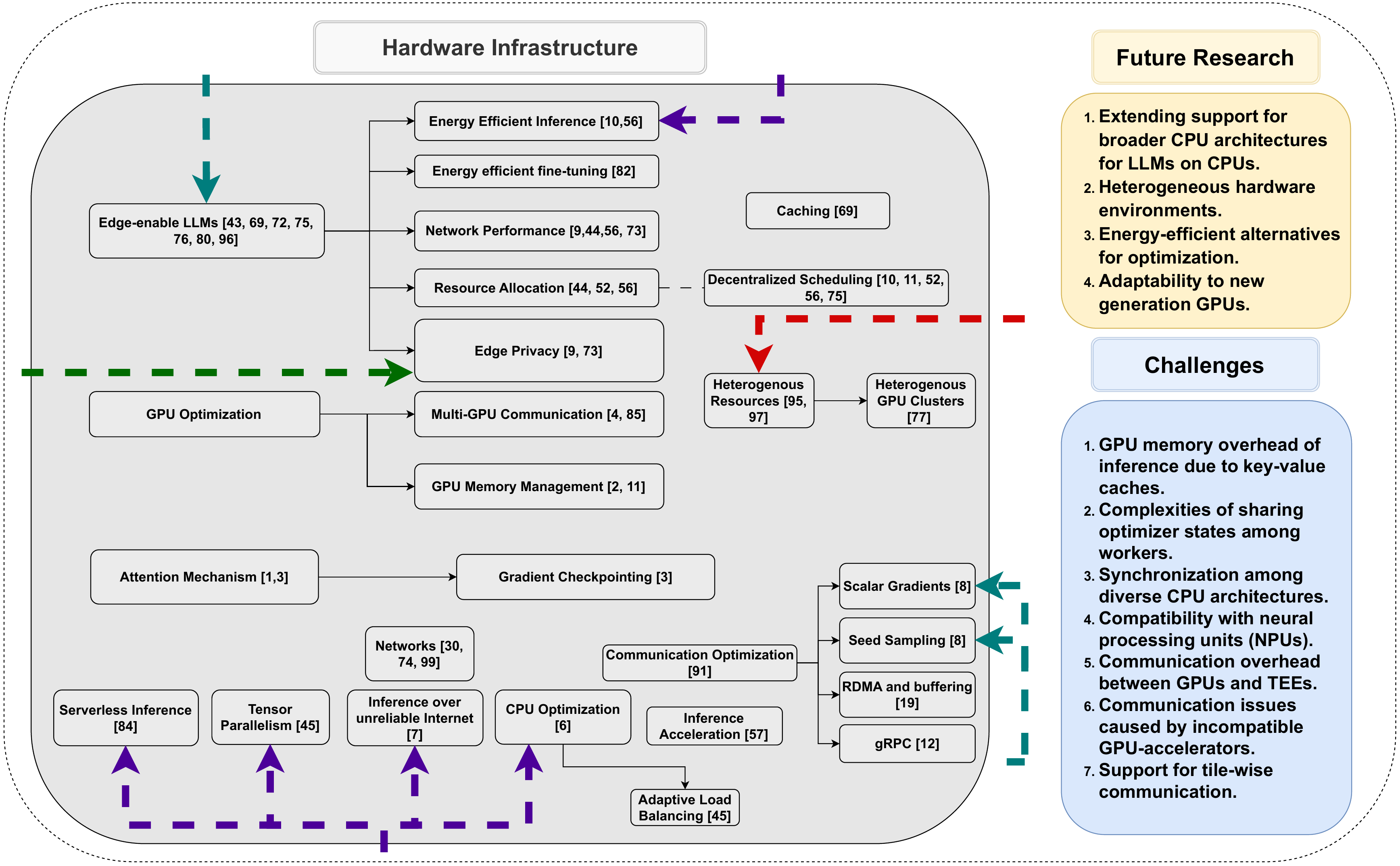} 
    \caption{Hardware Infrastructure: deployment challenges and hardware-based acceleration solutions for DLLMs.}
  \label{figint2}  
\end{figure*}

In this section, we present the core novelties of the selected related works. We discuss the contributions and challenges of each study, and outline potential future directions based on their findings. In this section,  our main focus is on non-survey articles. Please note that Table \ref{codes100} summarizes the links to the code for some of these papers. 
Further, Fig. \ref{figint} shows the connection of some of these papers and how they contribute to various aspects of decentralizing the LLM pipeline.This figure shows four primary areas of research focus: \textit{Algorithmic Framework}, \textit{Hardware Infrastructure}, \textit{Data Layer}, and \textit{Security and Privacy}. Within each of these categories, we further highlight the core challenges and technological solutions proposed by existing works. Further, we illustrate the relationships and interactions among the identified challenges and solutions both within and across these areas. Studies in the \textit{Algorithmic Framework} mainly follow two approaches: proposing novel pre-training or post-training methods tailored specifically for decentralized settings (e.g., intra-parallelism), or adapting and evaluating established pre-training or post-training approaches, such as retrieval-augmented generation, within decentralized environments. Furthermore, this category includes research on practical applications of distributed large language models (DLLMs). Research on \textit{Hardware Infrastructure} either examines deployment challenges encountered within specific hardware contexts, such as edge computing networks, or introduces hardware-level optimization strategies targeting computational and communication efficiency. Literature focusing on the \textit{Data Layer} primarily addresses decentralization during the data preparation and management stages. Finally, the \textit{Security and Privacy} category covers the integration and deployment of existing security and privacy-preserving techniques within decentralized learning contexts. More detailed versions of the algorithmic framework and hardware infrastructure categories, along with their corresponding future research directions and challenges, are provided in Fig. \ref{figint1} and Fig. \ref{figint2} respectively. It is worth mentioning that a comprehensive survey  focusing mainly on \textit{security and privacy challenge of LLMs} is provided by Das et al. \cite{SurveyLLM15das2024security}.

\begin{table*}[h]
\centering
\begin{tabular}{|l|l|}
\hline
\textbf{Paper } & \textbf{Code Link} \\
\hline
Wu et al. \cite{2wu2023fastLLM} & https://github.com/fast-inference-serving \\ \hline
Li et al. \cite{3li2024distflashattnLLM} & https://github.com/DISTFLASHATTN \\ \hline
Nabli et al. \cite{4nabli2024accoLLM} & https://github.com/acco-llm \\ \hline
Borzunov et al. \cite{7borzunov2024distributedLLM}   & https://github.com/LLM-distributed-inference \\ \hline
Qin et al. \cite{8qin2023federatedLLM} & https://github.com/federated-parameter-tuning \\ \hline
Khoshsirat et al. \cite{10khoshsirat2024decentralizedLLM} & https://github.com/edge-LLM-inference \\ \hline
Yao et al. \cite{12yao2024scalellmLLM} & https://github.com/ScaleLLM-framework \\ \hline
Kuang et al. \cite{15kuang2024federatedscopeLLM} & \url{https://github.com/alibaba/FederatedScope/tree/llm} \\ \hline
Xu et al. \cite{16xu2023fwdllmLLM} & \url{https://github.com/UbiquitousLearning/FwdLLM.git} \\ \hline
Shu et al. \cite{20shu2024ferretLLM} & https://github.com/Ferret-tuning \\ \hline
Ye et al. \cite{21ye2024openfedllmLLM} & \url{https://github.com/rui-ye/OpenFedLLM} \\ \hline
Zhang et al. \cite{27zhang2024fedpitLLM} & \url{https://github.com/UbiquitousLearning/FedPIT} \\ \hline
Huang et al. \cite{28huang2024frameworkLLM} & https://github.com/secure-distributed-LLM \\ \hline
Zhang et al. \cite{29pan2024cloudLLM} & https://github.com/CloudEdge-LLM \\ \hline
Peng et al. \cite{32peng2024fedpftLLM} & \url{https://github.com/pzp-dzd/FedPFT} \\ \hline
Wang et al. \cite{35wang2024cycleblackLLM} & \url{https://github.com/L3030/FedCyBGD} \\ \hline
Zheng et al. \cite{42zheng2024safely} & \url{https://github.com/TAP-LLM/SplitFedLLM} \\
\hline
Li et al. \cite{45li2024collm} & \url{https://github.com/zyang1580/CoLLM}\\
\hline
Liu et al. \cite{51liu2024asynchronous} & \url{https://github.com/google-deepmind/asyncdiloco}\\
\hline
Hagemann et al. \cite{53hagemann2023efficient} & \url{https://github.com/JohannesHa/COLM-submission-efficient-parallelization-layouts}\\
\hline
Li et al. \cite{56li2024tpi} & \url{https://github.com/Lizonghang/TPI-LLM}\\
\hline
Hu et al. \cite{59hu2023llm} & \url{https://github.com/AGI-Edgerunners/LLM-Adapters}\\
\hline
Sheng et al. \cite{71sheng2024hybridflow} & \url{https://github.com/volcengine/verl}\\
\hline
Shen et al. \cite{72shen2024edgeqat} & \url{https://github.com/shawnricecake/edge-qatl}\\
\hline
Fan et al. \cite{81fan2023fate} & \url{https://github.com/FederatedAI/FATE-LLM}\\
\hline
Fu et al. \cite{84fu2024serverlessllm} &\url{https://github.com/ServerlessLLM/ServerlessLLM}\\
\hline
Zeng et al. \cite{88zeng2024fair} &\url{https://github.com/huiminzeng/FF-DVP}\\
\hline
Lin et al. \cite{92lin2024splitlora} &\url{https://github.com/FDU-INC/Split_LoRA}\\
\hline
Bai et al. \cite{97wang2024flora} &\url{https://github.com/ATP-1010/FederatedLLM}\\
\hline
Nguyen et al. \cite{100nguyen2024flora} &\url{https://github.com/dphuongn/FLoRA}\\
\hline
\end{tabular}
\caption{Papers with code links for distributed and federated LLMs.}
\label{codes100}
\end{table*}

\paragraph*{\textbf{Wu et al. \cite{2wu2023fastLLM}, Fast Distributed Inference Serving for LLMs}}
\underline{Overview and Contributions:} Wu et al. \cite{2wu2023fastLLM} proposed FastServe, a distributed inference serving system designed to optimize low-latency requirements for LLMs. It addresses the shortcomings of traditional run-to-completion systems by leveraging preemptive scheduling and GPU memory management.

\begin{center}
\begin{tcolorbox}
\vspace{-0.05in}
\noindent \textbf{Contributions:}
Wu et al. \cite{2wu2023fastLLM} introduced three core contributions: i) a novel skip-join Multi-Level Feedback Queue scheduler designed to eliminate head-of-line blocking and reduce latency by leveraging semi information-agnostic scheduling; ii) a proactive GPU memory management mechanism that minimizes memory bottlenecks by dynamically offloading and uploading key-value tensors; and iii) a system prototype of FastServe that achieves substantial throughput improvements  compared to existing solutions under comparable latency constraints.
\end{tcolorbox}
\end{center}

\underline{Challenges:} The primary challenges addressed in this work include managing the variable job size characteristic of LLM inference (which introduces unpredictability in processing times) and the significant GPU memory overhead caused by maintaining key-value caches for both ongoing and preempted jobs. 

\underline{Future Directions:} Potential future directions include refining the scheduling algorithms to capture additional dynamic workload patterns and optimizing memory management techniques to further accommodate large-scale models with increasing context lengths;  expanding the framework to support real-time dynamic adjustments in parallelism strategies and enhancing integration with edge-based serving environments to extend its applicability; and minimizing energy consumption and improving sustainability for large-scale deployments.

\paragraph*{\textbf{Li et al. \cite{3li2024distflashattnLLM}, Distributed Memory-efficient Attention for Long-context LLMs Training}}
\underline{Overview and Contributions:} Li et al. \cite{3li2024distflashattnLLM} introduced DISTFLASHATTN which is a distributed attention mechanism optimized for memory efficiency in training long-context LLMs. By leveraging advanced parallelization and scheduling techniques, DISTFLASHATTN enables scalable and efficient training of large models with significantly extended sequence lengths.

\begin{center}
\begin{tcolorbox}
\vspace{-0.05in}
\noindent \textbf{Contributions:}
Li et al. \cite{3li2024distflashattnLLM} proposed DISTFLASHATTN with three major contributions: i) a load-balanced scheduling strategy that reduces idle time in causal modeling and enhances GPU utilization, leading to up to 2× throughput improvement; ii) a novel communication overlap mechanism that minimizes the overhead of token communication during distributed training, achieving a 1.32× end-to-end speedup; and iii) a re-materialization-aware gradient checkpointing technique that avoids redundant re-computations of FlashAttention, resulting in a 1.31× training speedup for long sequences.
\end{tcolorbox}
\end{center}

\underline{Challenges:} The work addresses critical challenges in distributed training of long-context LLMs including token-level workload imbalance inherent in causal language modeling which leads to suboptimal GPU utilization; excessive communication overhead caused by transferring key-value tensors and attention statistics, especially for long sequences; and inefficiencies in gradient checkpointing that result in redundant recomputation during backpropagation, consuming  computational resources.

\underline{Future Directions:} Future research directions include further optimizing distributed attention mechanisms to handle even longer sequences with minimal overhead and exploring new parallelization strategies that dynamically adapt to workload variations;  enhancing compatibility with sparse attention mechanisms and developing methods for energy-efficient training of LLMs; and extending DISTFLASHATTN to support real-time inference scenarios and hybrid cloud-edge environments to potentially expand its applicability in practical use cases. Further, alternative attention mechanisms, such as Mamba,  can scale to larger models and offer a promising approach to reducing the memory impact of long-context training and inference.

\paragraph*{\textbf{Nabli et al. \cite{4nabli2024accoLLM}, ACCO: Accumulate while you Communicate}}
\underline{Overview and Contributions:} Nabli et al. \cite{4nabli2024accoLLM} proposed ACCO, a novel memory-efficient optimization algorithm designed for distributed training of LLMs. ACCO effectively reduces communication costs by overlapping gradient computation and communication that also improves GPU utilization, and enables scalability by sharing optimizer states and dynamically adapting to heterogeneous hardware environments.

\begin{center}
\begin{tcolorbox}
\vspace{-0.05in}
\noindent \textbf{Contributions:}
Nabli et al. \cite{4nabli2024accoLLM} proposed ACCO with four major contributions: i) a shared optimization strategy enabling efficient memory utilization across distributed workers to overhead; ii) an overlapping gradient computation and communication mechanism that completely hides communication latency, ensuring high GPU utilization; iii) a delay compensation technique that eliminates the need for warm-up steps while maintaining convergence dynamics compared to standard distributed optimization; and iv) adaptability to heterogeneous hardware through dynamic workload adjustment.
\end{tcolorbox}
\end{center}

\underline{Challenges:} The study highlights critical challenges such as managing communication bottlenecks in large-scale distributed training, which can  impact wall-clock training time; and  complexities related to achieving memory efficiency while sharing optimizer states and ensuring compatibility with heterogeneous hardware.

\underline{Future Directions:} Future research can focus on extending ACCO to hybrid cloud-edge environments to further optimize real-world deployments; enhancing the algorithm to incorporate advanced gradient compression techniques for additional memory savings and exploring energy-efficient optimizations to minimize the carbon footprint of large-scale training tasks; and investigating compatibility with emerging transformer architectures and integrating ACCO with asynchronous decentralized frameworks to achieve more  scalability and efficiency.

\paragraph*{\textbf{Brakel et al. \cite{5brakel2024modelLLM}, Model Parallelism on Distributed Infrastructure: A Literature Review from Theory to LLM Case-Studies}}
\underline{Overview and Contributions:} Brakel et al. \cite{5brakel2024modelLLM} conducted a comprehensive review of model parallelism techniques, emphasizing their applicability to large neural networks, including LLMs. The paper systematically categorizes types of parallelism, identifies core challenges, and explores real-world implementations in state-of-the-art transformer architectures.

\begin{center}
\begin{tcolorbox}
\vspace{-0.05in}
\noindent \textbf{Contributions:}
Brakel et al. \cite{5brakel2024modelLLM} made three key contributions: i) a taxonomy of model parallelism strategies, differentiating between inter-operator and intra-operator parallelism, and their hybrid implementations; ii) an extensive analysis of the challenges of model partitioning, including  communication bottlenecks in distributed systems; and iii) a detailed evaluation of use cases, specifically the adaptation of model parallelism for large-scale transformer models such as GPT, Megatron, and PaLM.
\end{tcolorbox}
\end{center}

\underline{Challenges:} The primary challenges identified in this work include the complexities of optimizing inter-operator and intra-operator parallelism strategies, particularly in balancing compute and communication loads across heterogeneous hardware. Pipeline parallelism can lead to to synchronization delays and intra-operator parallelism leads to severe bandwidth and latency constraints, especially when scaling across multiple nodes. Further, achieving robust auto-parallelization strategies is still challenging due to the computational complexity of navigating a large search spaces with diverse performance metrics and the hardware dependency of these methods.

\underline{Future Directions:} Future research could focus on enhancing automated parallelization frameworks to better optimize hybrid parallelism strategies in complex neural networks; developing standard benchmarks and datasets for evaluating parallelization methods  to facilitate comparative studies; as well as addressing energy efficiency in distributed training, improving hardware utilization rates, and exploring new paradigms such as adaptive parallelism for dynamic workloads.

\paragraph*{\textbf{He et al. \cite{6he2024distributedLLM}, Distributed Inference Performance Optimization for LLMs on CPUs}}
\underline{Overview and Contributions:} He et al. \cite{6he2024distributedLLM} proposed an efficient distributed inference optimization solution tailored for running LLMs on CPUs. By leveraging advanced communication techniques and architectural optimizations, the solution addresses key challenges of memory usage and latency in resource-constrained environments.

\begin{center}
\begin{tcolorbox}
\vspace{-0.05in}
\noindent \textbf{Contributions:}
He et al. \cite{6he2024distributedLLM} made three significant contributions: i) a scalable synchronization mechanism that broadcasts token IDs instead of embeddings to reduce communication overhead and improve scalability; ii) a one-time synchronization approach that optimizes communication for decoder layers by parallelizing attention and feed-forward computations; and iii) a zero-copy memory optimization technique that directly writes computation results to communication buffers to eliminate redundant memory transfers.
\end{tcolorbox}
\end{center}

\underline{Challenges:} The key challenges tackled in this study include mitigating communication bottlenecks during distributed inference, especially in scenarios with high latency and constrained bandwidth;  efficiently mapping LLM computations onto CPU hardware with limited memory and managing synchronization across distributed nodes without compromising latency or throughput that further complicate the optimization process; and  ensuring compatibility with diverse CPU architectures that adds to the complexity of the solution.

\underline{Future Directions:} Future research could focus on extending the proposed techniques to support a broader range of CPU architectures and exploring adaptive scheduling mechanisms for heterogeneous hardware environments; incorporating advanced compression techniques to further reduce communication costs and expanding the framework to enable real-time applications in constrained settings; and developing open-source implementations to promote adoption and benchmarking.

\paragraph*{\textbf{Borzunov et al. \cite{7borzunov2024distributedLLM}, Distributed Inference and Fine-Tuning of LLMs Over the Internet}}
\underline{Overview and Contributions:} Borzunov et al. \cite{7borzunov2024distributedLLM} proposed a fault-tolerant decentralized system for distributed inference and fine-tuning of LLMs over the Internet. The proposed system addresses challenges in running LLMs on geo-distributed and unreliable hardware while achieving high performance and scalability.

\begin{center}
\begin{tcolorbox}
\vspace{-0.05in}
\noindent \textbf{Contributions:}
Borzunov et al. \cite{7borzunov2024distributedLLM} introduced three major contributions: i) a novel fault-tolerant autoregressive inference algorithm that is capable of handling server failures and dynamic participation in a distributed environment; ii) the PETALS system which enables decentralized hosting and execution of LLMs to provide correctness guarantees despite heterogeneous and unreliable nodes; and iii) an empirical evaluation demonstrating that PETALS achieves up to 10× speedup in autoregressive generation compared to local offloading, even in geo-distributed setups spanning multiple continents.
\end{tcolorbox}
\end{center}

\underline{Challenges:} The primary challenges addressed by this work include handling unreliable, geographically distributed devices prone to disconnections, achieving low-latency communication in high-latency networks, and optimizing load balancing across devices with heterogeneous capabilities. Further, there are additional complexities due to  ensuring fault tolerance while maintaining throughput and correctness in inference and fine-tuning processes.

\underline{Future Directions:} Future work can focus on improving data privacy within the PETALS system by exploring secure multiparty computation or homomorphic encryption methods; enhancing load balancing algorithms for even greater adaptability to device heterogeneity and network variability ; and integrating advanced compression techniques and extending support to more model architectures and hardware configurations to broaden the applicability of the system.

\paragraph*{\textbf{Qin et al. \cite{8qin2023federatedLLM}, Federated Full-Parameter Tuning of Billion-Sized Language Models with Communication Cost Under 18 Kilobytes}}
\underline{Overview and Contributions:} Qin et al. \cite{8qin2023federatedLLM} proposed FedKSeed which is  a novel approach for federated full-parameter tuning of LLMs that  reduces communication costs. By leveraging zeroth-order optimization and a finite seed paradigm, FedKSeed enables scalable FL while maintaining high accuracy and minimal resource utilization.

\begin{center}
\begin{tcolorbox}
\vspace{-0.05in}
\noindent \textbf{Contributions:}
Qin et al. \cite{8qin2023federatedLLM} have three major contributions: i) the development of FedKSeed, a method enabling full-parameter tuning of billion-parameter LLMs with communication costs under 18 kilobytes per round, using random seeds and scalar gradients to replace traditional parameter transmission; ii) a probability-differentiated seed sampling mechanism that prioritizes effective perturbations to enhance model accuracy and reduce synchronization time; and iii)  experimental evaluations showing significant performance improvements across multiple datasets and configurations, including an improvement in Rouge-L scores over existing federated fine-tuning methods.
\end{tcolorbox}
\end{center}

\underline{Challenges:} The primary challenges addressed in this work include managing the immense communication costs typically associated with full-parameter tuning in FL especially for billion-sized LLMs. Existing methods either rely on parameter-efficient tuning with suboptimal accuracy or add additional computational and memory costs due to extensive backpropagation. There are additional complexities caused by ensuring compatibility with resource-constrained devices and achieving convergence under statistically heterogeneous data conditions.

\underline{Future Directions:} Future research can explore integrating FedKSeed with advanced privacy-preserving techniques to enhance data confidentiality during  FL; optimizing the seed sampling strategy to dynamically adapt to varying data distributions and expanding the framework to support asynchronous federated settings; and reducing the reliance on zeroth-order optimization through hybrid approaches.

\paragraph*{\textbf{Qu et al. \cite{9qu2024mobileLLM}, Mobile Edge Intelligence for LLMs: A Contemporary Survey}}
\underline{Overview and Contributions:} Qu et al. \cite{9qu2024mobileLLM} provided a comprehensive survey of integrating LLMs with mobile edge intelligence (MEI) to address latency, bandwidth, and privacy challenges associated with deploying LLMs on edge devices. The paper emphasizes resource-efficient deployment techniques and presents a framework for edge-enabled LLMs.

\begin{center}
\begin{tcolorbox}
\vspace{-0.05in}
\noindent \textbf{Contributions:}
Qu et al. \cite{9qu2024mobileLLM} made three significant contributions: i) they outlined an MEI framework for LLMs (MEI4LLM) that integrates model caching, training, and inference with resource-efficient approaches tailored for edge environments; ii) they explored advanced techniques such as  parameter-sharing LLM caching and split learning to optimize communication, storage, and computation at the network edge; and iii) they identified critical application domains, such as healthcare and autonomous driving, highlighting the necessity of edge LLM deployment for latency-sensitive and privacy-critical tasks.
\end{tcolorbox}
\end{center}

\underline{Challenges:} The primary challenges discussed include the computational and memory overhead associated with deploying LLMs at the edge, as current edge devices lack the resources to support large-scale models effectively. Further, achieving efficient communication between distributed nodes while maintaining low latency and handling data heterogeneity presents significant obstacles. The work also underscores the complexity of integrating AI techniques with wireless communication systems in MEI frameworks.

\underline{Future Directions:} Future research can focus on enhancing green edge AI methods to reduce energy consumption during LLM training and inference at the edge. Secure edge AI for LLMs while integrating privacy-preserving techniques is another promising direction. Expanding support for multimodal LLMs and improving interoperability across diverse edge hardware and software ecosystems will be critical for broader adoption of MEI4LLM solutions.

\paragraph*{\textbf{Khoshsirat et al. \cite{10khoshsirat2024decentralizedLLM}, Decentralized LLM Inference Over Edge Networks with Energy Harvesting}}
\underline{Overview and Contributions:} Khoshsirat et al. \cite{10khoshsirat2024decentralizedLLM} proposed a decentralized approach to LLM inference tailored for energy-constrained edge networks with energy harvesting. The study introduces a semi-Markov model and scheduling algorithms to enhance resource utilization and task throughput while maintaining energy efficiency.

\begin{center}
\begin{tcolorbox}
\vspace{-0.05in}
\noindent \textbf{Contributions:}
Khoshsirat et al. \cite{10khoshsirat2024decentralizedLLM} made three primary contributions: i) the development of a semi-Markov model to characterize battery states and energy dynamics in edge devices for optimal scheduling; ii) the design of adaptive scheduling algorithms that minimize device downtimes and maximize throughput by leveraging energy availability; and iii) the implementation and evaluation of decentralized inference using commercial edge devices (e.g., Nvidia Jetson AGX Orin), demonstrating significant improvements in energy utilization and task performance.
\end{tcolorbox}
\end{center}

\underline{Challenges:} The study highlights several challenges, including managing energy constraints in battery-powered edge devices while maintaining reliable and efficient LLM inference. Addressing the heterogeneity of edge devices, such as varying energy arrival rates and processing capabilities, complicates scheduling. Further, ensuring robust performance under fluctuating energy conditions from renewable sources introduces significant design complexity.

\underline{Future Directions:} Future research can focus on integrating advanced energy prediction models to enhance the scheduling algorithm's adaptability to dynamic energy conditions. Extending the approach to incorporate more complex LLM architectures and multimodal models can broaden applicability.

\paragraph*{\textbf{Ren et al \cite{11ren2024taskLLM}, Task Scheduling for Decentralized LLM Serving in Heterogeneous Networks}}
\underline{Overview and Contributions:} Ren et al \cite{11ren2024taskLLM} proposed an efficient task scheduling algorithm for decentralized serving of LLMs across heterogeneous GPU networks. Their work focuses on minimizing the time per output token (TPOT) by leveraging model pipelining and adaptive scheduling in distributed settings.

\begin{center}
\begin{tcolorbox}
\vspace{-0.05in}
\noindent \textbf{Contributions:}
Ren et al \cite{11ren2024taskLLM} introduced three key contributions: i) a heuristic-based scheduling algorithm designed to optimize TPOT by balancing computational and communication overheads in heterogeneous GPU networks, ii) applying model pipelining to distribute inference workloads across decentralized GPUs to enable the deployment of consumer-grade devices for LLM inference, and iii) a comprehensive evaluation across various testbeds to show the superior latency performance compared to integer programming and random search baselines.
\end{tcolorbox}
\end{center}

\underline{Challenges:} The primary challenges addressed include managing the heterogeneity of GPU devices, which is different in memory capacity, computation speed, and network latency. It also ensures that inference tasks are scheduled without exceeding memory constraints or introducing significant communication delays further complicates the problem. Further, it addressed the issue of achieving robust performance in dynamic, real-world decentralized networks with fluctuating resource availability that can add complexity to the scheduling design.

\underline{Future Directions:} Future research directions can  explore integrating advanced energy-efficient strategies to minimize power consumption during decentralized LLM serving. Other promising future research directions include enhancing the algorithm's adaptability to real-time network fluctuations and supporting asynchronous task scheduling, expanding the framework to accommodate multimodal LLMs and heterogeneous hardware types, such as newer-generation GPUs or edge devices.

\paragraph*{\textbf{Yao et al \cite{12yao2024scalellmLLM}, ScaleLLM: A Resource-Frugal LLM Serving Framework by Optimizing End-to-End Efficiency}}
\underline{Overview and Contributions:} Yao et al \cite{12yao2024scalellmLLM} proposed ScaleLLM, a framework designed to optimize the end-to-end efficiency of LLM serving in commercial applications. By optimizing computational resource allocation and addressing end-to-end latency, ScaleLLM achieves high throughput, low latency, and efficient resource utilization.

\begin{center}
\begin{tcolorbox}
\vspace{-0.05in}
\noindent \textbf{Contributions:}
Yao et al \cite{12yao2024scalellmLLM} introduced three primary contributions: i) an end-to-end latency breakdown of LLM serving pipelines to identify bottlenecks in inference engines and gateways, ii) optimizations for the inference engine, including hybrid parallelism and quantization techniques, alongside gateway improvements using Rust-based routing and gRPC protocol for low-latency communication, and iii)  experimental evaluations showing a 4.3× speedup over vLLM and a 1.5× improvement in throughput compared to state-of-the-art solutions.
\end{tcolorbox}
\end{center}

\underline{Challenges:} This work addresses  challenges in optimizing LLM serving systems, including managing high latency caused by concurrent requests and computational overhead in routing gateways. However, balancing the computational loads across GPUs using tensor and expert parallelism while ensuring minimal resource consumption is still a complex task. 

\underline{Future Directions:} Future research could focus on developing adaptive load-balancing systems that are capable of dynamically optimizing resource allocation for varying workloads. Another future research direction is integrating more advanced energy-efficient techniques for inference. In order to enhance practicality of the proposed framework, it can be  expanded to support MLLMs and improve interoperability with various hardware infrastructures.

\paragraph*{\textbf{Kuang et al \cite{15kuang2024federatedscopeLLM}, FederatedScope-LLM: A Comprehensive Package for Fine-Tuning LLMs in FL}}
\underline{Overview and Contributions:} Kuang et al \cite{15kuang2024federatedscopeLLM} introduced FederatedScope-LLM (FS-LLM), an open-source framework designed for federated fine-tuning of LLMs.   It addresses communication and computation challenges in FL settings and provides a comprehensive solution for privacy-preserving, resource-efficient, and domain-specific fine-tuning of LLMs while integrating benchmarking, advanced algorithms, and extensible experimentation tools.

\begin{center}
\begin{tcolorbox}
\vspace{-0.05in}
\noindent \textbf{Contributions:}
Kuang et al \cite{15kuang2024federatedscopeLLM} made three core contributions: i) they developed FS-LLM with a benchmarking pipeline for federated fine-tuning, including automated dataset preparation, fine-tuning execution, and evaluation tasks, ii) implemented PEFT algorithms, including LoRA and prefix-tuning, to reduce communication and computation costs, and iii) equipped FS-LLM with resource-efficient operators and flexible interfaces to support interdisciplinary applications like personalized FL (pFL) and federated hyperparameter optimization (FedHPO).
\end{tcolorbox}
\end{center}

\underline{Challenges:} The work addresses major challenges in federated fine-tuning of LLMs, including high computational and communication demands, especially on resource-constrained clients. Ensuring robust performance when clients have heterogeneous data distributions and limited access to full model parameters introduces additional complexity. There is a still a gap in integrating privacy-preserving techniques while maintaining efficiency.

\underline{Future Directions:} Future research can focus on developing more efficient PEFT algorithms to reduce computation complexity for resource-constrained clients. It can also focus on optimizing pFL algorithms to improve compatibility with efficient training operators and exploring low-fidelity FedHPO methods to address hyperparameter sensitivity.

\paragraph*{\textbf{Xu et al \cite{16xu2023fwdllmLLM}, FwdLLM: Efficient FedLLM Using Forward Gradient}}
\underline{Overview and Contributions:} Xu et al \cite{16xu2023fwdllmLLM} proposed FwdLLM, an FL framework that uses backpropagation-free (BP-free) training for LLMs on resource-constrained mobile devices. FwdLLM  reduces memory and computation overhead and enables scalable and efficient federated fine-tuning of LLMs by using forward gradient methods and PEFT.

\begin{center}
\begin{tcolorbox}
\vspace{-0.05in}
\noindent \textbf{Contributions:}
Xu et al \cite{16xu2023fwdllmLLM} introduced three key contributions: i) the development of a BP-free training algorithm that replaces traditional gradient computation with memory-efficient perturbed inferences, ii) the integration of PEFT methods, such as LoRA and adapters, with forward gradient techniques to optimize resource usage, and iii) the design of variance-controlled perturbation pacing and discriminative perturbation sampling to enhance training efficiency and accelerate convergence.
\end{tcolorbox}
\end{center}

\underline{Challenges:} The primary challenges addressed by FwdLLM include the high memory and computational requirements of backpropagation-based training, which are incompatible with mobile neural processing units (NPUs) and other device constraints. Furthermore, balancing the trade-offs between computational cost and training accuracy, particularly in generating effective perturbations, posed significant design complexities. Ensuring scalability with a large number of devices and non-iid data distributions also presented challenges.

\underline{Future Directions:} Future work could focus on enhancing the scalability of FwdLLM for larger and more complex LLMs, integrating  privacy-preserving techniques such as differential privacy (DP) and secure aggregation. Exploring adaptive optimization strategies for dynamic task and device conditions, and extending the framework to multimodal LLMs for diverse downstream applications could also improve its applicability in real-world scenarios. Further, integrating energy-efficient mechanisms and optimizing compatibility with emerging edge hardware architectures are promising avenues.

\paragraph*{\textbf{Ling et al \cite{17ling2024convergenceLLM}, On the Convergence of Zeroth-Order Federated Tuning for LLMs}}
\underline{Overview and Contributions:} Ling et al \cite{17ling2024convergenceLLM} proposed FedMeZO, a memory-efficient federated tuning approach that integrates Zeroth-Order Optimization (ZOO) into LLMs. This method addresses computational challenges in federated fine-tuning by leveraging ZOO for reduced memory consumption while maintaining robust convergence properties.

\begin{center}
\begin{tcolorbox}
\vspace{-0.05in}
\noindent \textbf{Contributions:}
Ling et al \cite{17ling2024convergenceLLM} provided three primary contributions: i) they developed FedMeZO, introducing a two-point ZOO gradient estimator to minimize memory requirements while achieving effective tuning, ii) they established theoretical convergence bounds for both i.i.d. and non-i.i.d. federated settings while exploring the role of low-rank Hessian matrices, and iii) they proposed a personalized FL strategy, enabling adaptive learning rates tailored to heterogeneous client datasets.
\end{tcolorbox}
\end{center}

\underline{Challenges:} The work addresses key challenges in federated fine-tuning of LLMs, including high memory demands during gradient computation and the difficulty of ensuring convergence within heterogeneous client environments. The use of zeroth-order gradients introduces additional complexity in achieving reliable convergence, particularly when handling large parameter spaces typical of LLMs. It is still challenging to balance  personalization with computational efficiency.

\underline{Future Directions:} Future research directions include but not limited to  improving the efficiency of ZOO algorithms to further reduce memory consumption during LLM fine-tuning; exploring adaptive mechanisms for better personalization across diverse client data distributions and integrating secure aggregation techniques for enhanced privacy; and applying FedMeZO to multimodal and cross-device FL scenarios could broaden its applicability.

\paragraph*{\textbf{Zhang et al \cite{19zhang2024fedrdmaLLM}, FedRDMA: Communication-Efficient Cross-Silo Federated LLM via Chunked RDMA Transmission}}
\underline{Overview and Contributions:} Zhang et al \cite{19zhang2024fedrdmaLLM} proposed FedRDMA, a novel framework for optimizing communication efficiency in cross-silo FL of LLM by utilizing Remote Direct Memory Access (RDMA) with chunked data transmission. The approach addresses the limitations of RDMA in wide-area networks (WANs) and ensures robust performance under WAN constraints.

\begin{center}
\begin{tcolorbox}
\vspace{-0.05in}
\noindent \textbf{Contributions:}
Zhang et al \cite{19zhang2024fedrdmaLLM} made three major contributions: i) they developed FedRDMA, which integrates chunked RDMA transmission with advanced buffering and reassembly strategies, enabling efficient data transfer on WANs, ii) proposed an optimized FedRDMA-E variant that eliminates the need for reassembly and temporary storage by leveraging memory pools, reducing overhead, and iii) improves efficiency over TCP-based systems, integrates parameter-efficient tuning, and reduces energy usage for sustainable FL.
\end{tcolorbox}
\end{center}

\underline{Challenges:} The paper highlights significant challenges, such as overcoming RDMA’s dependency on lossless networks, which limits its applicability on WANs due to packet loss and retransmissions. Ensuring the stable operation of RDMA in dynamic WAN environments with varying bandwidth and latency, while maintaining compatibility with FL protocols, adds complexity. Further, the computational and memory overhead introduced by chunking and reassembly in FedRDMA requires careful optimization to avoid diminishing the performance benefits.

\underline{Future Directions:} Future research could focus on extending FedRDMA to support more complex WAN environments and further optimizing chunking strategies for diverse network conditions; integrating FedRDMA with privacy-preserving techniques; and  exploring compatibility with emerging RDMA technologies and extending the framework for larger-scale FL systems.

\paragraph*{\textbf{Shu et al \cite{20shu2024ferretLLM}, Ferret: Federated Full-Parameter Tuning at Scale for LLMs}} \underline{Overview and Contributions:} Shu et al \cite{20shu2024ferretLLM} proposed Ferret, a federated full-parameter tuning algorithm designed for large-scale deployment of LLMs in decentralized settings. This work addresses communication  and computational challenges by combining first-order methods for local updates with low-dimensional projections and shared randomness to reconstruct global updates. Ferret achieves scalability, competitive model accuracy, reduced communication overhead, and faster convergence which  make it a robust solution for FL of LLMs.

\begin{center} \begin{tcolorbox} \vspace{-0.05in} \noindent \textbf{Contributions:} Shu et al \cite{20shu2024ferretLLM} proposed a novel approach for federated full-parameter tuning of LLMs with three major contributions: i) integrating first-order methods with low-dimensional projections to enhance computational and communication efficiency; ii) utilizing shared randomness to achieve unbiased and efficient reconstruction of global updates, ensuring scalability and convergence; iii) complementing the approach with rigorous theoretical analyses and extensive experiments, demonstrating superior performance compared to existing methods. \end{tcolorbox} \end{center}

\underline{Challenges:} The main challenges in this work revolve around balancing computational efficiency, communication overhead, and model convergence in a FL setup. The algorithm must ensure that the high-dimensional updates from billions of model parameters are effectively represented in a low-dimensional space without losing critical information. Further, there are still challenges about maintaining privacy while facilitating global aggregation and minimizing computational cost. 

\underline{Future Directions:} Future research can explore extending the Ferret algorithm to support heterogeneous data distributions more effectively, as real-world federated environments often involve non-IID data; optimizing the trade-offs between communication overhead and reconstruction accuracy to improve scalability for even larger models; and  investigating the integration of advanced optimization techniques, such as adaptive learning rates or gradient compression to enhance its performance and convergence speed in different settings.

\paragraph*{\textbf{Ye et al \cite{21ye2024openfedllmLLM}, OpenFedLLM: Training LLMs on Decentralized Private Data via FL}}
\underline{Overview and Contributions:} Ye et al \cite{21ye2024openfedllmLLM} proposed OpenFedLLM, a  framework for training LLMs on decentralized private data through FL. This work emphasizes privacy-preserving methods for collaborative LLM training. OpenFedLLM integrates multiple FL algorithms, FedIT, and value alignment while being resource-efficient and flexible to be deployed  in  diverse domains. It is evaluated using 7 FL algorithms, 8 datasets, and over 30 metrics. The proposed framework demonstrates exceptional scalability, computational efficiency, and performance, even outperforming state-of-the-art LLMs such as  GPT-4 in certain domains.

\begin{center}
\begin{tcolorbox}
\vspace{-0.05in}
\noindent \textbf{Contributions:}
Ye et al \cite{21ye2024openfedllmLLM} proposed a novel framework for collaborative LLM training. Their proposed solution has three major contributions: i) the introduction of FedIT and federated value alignment (FedVA) for enhancing instruction-following and human-value alignment capabilities, ii) the development of a concise and integrated framework that bridges the FL and LLM communities by incorporating 7 FL algorithms, 8 datasets, and 30+ evaluation metrics, and iii) conducting an extensive empirical study showing that FL can outperform individual training, with significant improvements in specific domains, e.g.,  finance.
\end{tcolorbox}
\end{center}

\underline{Challenges:} OpenFedLLM faces challenges such as handling heterogeneous client preferences in FedVA, managing the decentralized nature of data to ensure efficient aggregation, and ensuring robust training under diverse private datasets. Further, there are some challenges regarding the balance between  computational and communication efficiency while maintaining model performance.

\underline{Future Directions:} Future research directions for OpenFedLLM include developing new FL algorithms tailored for LLM training, exploring robust mechanisms to address data heterogeneity, and enhancing privacy-preserving methods for secure decentralized training. Furthermore, advancements in personalized FL could help optimize model performance across diverse client-specific requirements.

\paragraph*{\textbf{Ye et al \cite{23ye2024safetyLLM}, Emerging Safety Attack and Defense in Federated Instruction Tuning of LLMs}}
\underline{Overview and Contributions:} Ye et al \cite{23ye2024safetyLLM} proposed a comprehensive study on vulnerabilities in the safety alignment of FedIT for LLMs. Their work identifies critical risks posed by malicious clients during FL, introduces an effective safety attack methodology, and provides a novel post-hoc defense mechanism. These contributions  advance the understanding and practical safeguarding of collaborative LLM training systems.

\begin{center}
\begin{tcolorbox}
\vspace{-0.05in}
\noindent \textbf{Contributions:}
Ye et al \cite{23ye2024safetyLLM} proposed a novel approach to enhance safety in FedIT, with three major contributions: i) they introduced a stealthy and effective safety attack method where malicious clients train on safety-unaligned data, compromising global model alignment, ii) they developed an automated post-hoc defense mechanism that generates and utilizes safety-aligned data for fine-tuning the global model, and iii) they demonstrated the effectiveness of their methods through extensive experiments, showing substantial improvements in safety metrics even under attack scenarios.
\end{tcolorbox}
\end{center}

\underline{Challenges:} The primary challenge addressed in this work lies in the stealthiness of the proposed safety attack, which makes it difficult for existing FL defense mechanisms to identify and mitigate malicious clients. The similarity in optimization objectives between benign and malicious training increases this issue and can deteriorate the performance of traditional model-level defenses. Further, relying on decentralized data introduces variability and heterogeneity that can  complicate the identification  of safety risks.

\underline{Future Directions:} Future research can focus on developing more robust real-time detection mechanisms for malicious activities during the FL process; expanding the defense framework to accommodate a wider variety of attack vectors and dataset types;  and exploring cross-domain applications and scaling up the defense mechanism for larger and more diverse federated systems to enhance its applicability.

\paragraph*{\textbf{Ye et al \cite{24ye2024fedllmLLM}, FedLLM-Bench: Realistic Benchmarks for FL of LLMs}}
\underline{Overview and Contributions:} Ye et al \cite{24ye2024fedllmLLM} proposed \textbf{FedLLM-Bench}, a realistic benchmark designed to evaluate FL for LLMs (FedLLM). This work introduces four datasets (Fed-Aya, Fed-ChatbotIT, Fed-WildChat, and Fed-ChatbotPA). Each of these datasets captures real-world diversities such as language, quality, quantity, and user preferences. It also integrates 8 baseline training methods, {4 datasets}, and 6 evaluation metrics to facilitate a comprehensive analysis. The benchmark emphasizes practical scenarios of FedIT and preference alignment to address the gaps in existing artificially partitioned datasets.

\begin{center}
\begin{tcolorbox}
\vspace{-0.05in}
\noindent \textbf{Contributions:}
Ye et al \cite{24ye2024fedllmLLM} proposed a comprehensive benchmark for FedLLM. Their proposed solution has three major contributions: i) they designed four realistic datasets capturing user-specific properties such as multilingualism, data heterogeneity, and preferences, ii) they implemented 8 representative baseline methods and 6 evaluation metrics to support performance evaluation and comparison, and iii) they conducted extensive experiments to benchmark existing methods and explore new research directions such as multilingual collaboration.
\end{tcolorbox}
\end{center}

\underline{Challenges:} The paper identifies several challenges, including the  data heterogeneity across clients, which complicates model aggregation and performance evaluation; as well as  the varying qualities, quantities, and preferences in datasets that introduce complexity in achieving consistent performance gains. Addressing these issues while ensuring fair comparisons across methods is still a significant challenge.

\underline{Future Directions:} Future research directions  include exploring more effective collaboration strategies for multilingual datasets, developing advanced {language} personalization techniques to balance local adaptation and global model performance;  integrating safety measures  in the training process; and expanding benchmarks to incorporate more datasets and evaluation metrics to support broader integration and deployment of  FedLLM.

\paragraph*{\textbf{Wang et al \cite{25wang2024federatedLLM}, Federated Instruction Tuning of LLMs with Domain Coverage Augmentation}}
\underline{Overview and Contributions:} Wang et al \cite{25wang2024federatedLLM} proposed  FedDCA to enhance the performance of FedIT in LLMs. This approach focuses on augmenting domain-specific instructions by leveraging a combination of client-private and server-public datasets. The method introduces techniques for maximizing domain coverage while maintaining privacy and computational efficiency, particularly via a variant, FedDCA*, which utilizes heterogeneous encoders. Extensive experiments demonstrate significant improvements across various domains.

\begin{center}
\begin{tcolorbox}
\vspace{-0.05in}
\noindent \textbf{Contributions:}
 Wang et al \cite{25wang2024federatedLLM} has three core contributions: i) they revealed that cross-client domain coverage  impacts model performance and  challenged the assumption that heterogeneity negatively correlates with model effectiveness, ii)
 They introduced FedDCA to employ greedy client center selection and retrieval-based instruction augmentation to optimize domain coverage and improve task-specific LLM performance, and iii)
 They developed FedDCA*, which reduces computational overhead on the client side by employing heterogeneous encoders and server-side feature alignment via contrastive learning.
\end{tcolorbox}
\end{center}

\underline{Challenges:} The work faces challenges in maintaining a balance between computational efficiency and model performance, particularly when scaling FedDCA* for broader applications. Ensuring privacy while optimizing instruction augmentation presents technical difficulties, especially against memory extraction attacks. Further, aligning heterogeneous encoder outputs without compromising semantic accuracy remains a complex task.

\underline{Future Directions:} Future research can explore enhancing the robustness of FedDCA against  privacy threats, such as adversarial attacks; developing dynamic domain coverage metrics to adapt to evolving client data distributions could improve model generalization; and investigating alternative methods for efficient instruction augmentation, such as leveraging synthetic data or advanced generative techniques to further optimize performance.

\paragraph*{\textbf{Wu et al \cite{26wu2024fedbiotLLM}, FedBiOT: LLM Local Fine-tuning in FL without Full Model}}
\underline{Overview and Contributions:} Wu et al \cite{26wu2024fedbiotLLM} proposed FedBiOT, a framework enabling resource-efficient fine-tuning of LLMs in FL setting. This method addresses computational and communication bottlenecks by compressing the LLM into two components: an emulator for general patterns and an adapter for domain-specific tasks. FedBiOT leverages bi-level optimization to align these components effectively under heterogeneous data distributions. {Experiments on tasks such as math, code generation, and QA presents the accuracy gains over baselines  (e.g., Offsite-tuning and FedOT).}

\begin{center}
\begin{tcolorbox}
\vspace{-0.05in}
\noindent \textbf{Contributions:}
  Wu et al \cite{26wu2024fedbiotLLM} proposed a novel algorithm, FedBiOT, for federated fine-tuning of LLMs. Their proposed solution has three major contributions: i) introducing a compressed model structure with an emulator and an adapter to minimize resource usage while maintaining performance, ii) formulating a bi-level optimization to ensure effective alignment between server-side and client-side components despite data heterogeneity, and iii) demonstrating significant improvements in computational and communication efficiency with comprehensive experiments on LLaMA-2 and highlighting its enhanced performance over existing baselines.
\end{tcolorbox}
\end{center}

\underline{Challenges:} This work faces challenges such as managing the distributional shift between server and client datasets, which complicates the alignment of the emulator and adapter components. Further, there is still a challenge in ensuring stable convergence of the bi-level optimization under heterogeneous data and client resource constraints.

\underline{Future Directions:} Future research could explore enhancing the alignment between the emulator and adapter through advanced optimization techniques or leveraging adaptive aggregation methods for dynamic client-server interactions; investigating the applicability of FedBiOT to more diverse FL scenarios, including those with extreme non-IID data distributions; and optimizing the framework for other resource-constrained devices and incorporating privacy-preserving mechanisms.
  
\paragraph*{\textbf{Zhang et al \cite{27zhang2024fedpitLLM}, FewFedPIT: Towards Privacy-Preserving and Few-Shot Federated Instruction Tuning}}
\underline{Overview and Contributions:} Zhang et al \cite{27zhang2024fedpitLLM} proposed FewFedPIT, an FL framework designed to enhance privacy and efficiency in few-shot instruction tuning of LLMs. This framework deploys synthetic data generation, parameter isolation training, and secure local aggregation mechanisms to address challenges of data scarcity, privacy preservation, and performance optimization in federated settings. FewFedPIT achieves significant improvements in utility and security via innovative approaches and extensive evaluations.

\begin{center}
\begin{tcolorbox}
\vspace{-0.05in}
\noindent \textbf{Contributions:}
 Zhang et al \cite{27zhang2024fedpitLLM} proposed  three major contributions: i) leveraging LLMs for task-specific synthetic data generation, addressing the scarcity of high-quality local data in federated settings, ii) they introduced parameter isolation training to separate the handling of synthetic and private data, reducing noise and improving robustness against federated model attacks, and iii) the framework incorporates a secure local aggregation sharing mechanism to protect private data by blending public and private model parameters before global aggregation, mitigating the risk of training data extraction attacks.
\end{tcolorbox}
\end{center}

\underline{Challenges:} The work identifies several key challenges in FL including the scarcity of high-quality and diverse local datasets limits model performance and domain coverage;  ensuring the privacy of decentralized data while maintaining performance that introduces significant complexity, especially in few-shot learning scenarios;  the computational and communication overhead associated with federated aggregation and synthetic data generation that can pose scalability challenges; and aligning the FL framework with human-centered design principles.

\underline{Future Directions:} Future research in FedIT includes  the integration of  privacy-preserving mechanisms to ensure data confidentiality; optimizing synthetic data generation techniques to improve the quality and diversity of local datasets that  enhance model performance; and exploring lightweight and scalable aggregation methods to reduce computational and communication overhead while expanding the framework to support multimodal federated tasks.

\paragraph*{\textbf{Huang et al \cite{28huang2024frameworkLLM}, A Fast, Performant, Secure Distributed Training Framework for LLM}}
\underline{Overview and Contributions:} Huang et al \cite{28huang2024frameworkLLM} proposed a secure and efficient distributed training framework for LLMs that combines model slicing, Trusted Execution Environments (TEEs), and lightweight encryption techniques. This framework addresses the dual challenges of protecting sensitive data and optimizing computational efficiency in distributed settings. Further, the authors introduced a split fine-tuning strategy to enhance downstream task performance while reducing computational overhead.

\begin{center}
\begin{tcolorbox}
\vspace{-0.05in}
\noindent \textbf{Contributions:}
Huang et al \cite{28huang2024frameworkLLM} proposed three major contributions: 
i) utilizing TEE-based model slicing to securely partition and process sensitive model components, ensuring data confidentiality during distributed training, 
ii) developing a split fine-tuning strategy to partition LLM layers between client and server, enhancing training efficiency and downstream task performance, and 
iii) introducing a sparsification parameter fine-tuning (SPF) technique that identifies critical parameters for fine-tuning, balancing computational cost and model accuracy.
\end{tcolorbox}
\end{center}

\underline{Challenges:} The work identifies several key challenges in distributed training of LLMs, including the immense computational and memory demands of TEEs, which constrain scalability; the numerical precision issues arising from processing encrypted data during training; the communication overhead caused by frequent data exchanges between GPUs and TEEs, which limits efficiency; and the difficulty in balancing security, efficiency, and accuracy in a unified framework.

\underline{Future Directions:} Future research in distributed training of LLMs includes the development of hardware accelerators tailored to enhance TEE performance while reducing energy consumption; exploring adaptive encryption techniques to lower the communication overhead of secure data transmission; and designing advanced sparsification and fine-tuning methods that further reduce computational complexity without compromising model accuracy.

\paragraph*{\textbf{Chua et al \cite{30chua2023fedpeatLLM}, FedPEAT: Convergence of 6G Enabled FL, Parameter-Efficient Fine Tuning, and Emulator Assisted Tuning for AI Foundation Models}}
\underline{Overview and Contributions:} Chua et al \cite{30chua2023fedpeatLLM} proposed FedPEAT, a framework integrating Emulator-Assisted Tuning (EAT), PEFT, and FL for efficient and private fine-tuning of foundation models. This approach utilizes adapters and emulators for efficient model adaptation while preserving user data privacy and addressing computational and communication bottlenecks. FedPEAT incorporates an adaptive control mechanism based on reinforcement learning to optimize hyper-parameters for scalable and resource-efficient model fine-tuning.

\begin{center}
\begin{tcolorbox}
\vspace{-0.05in}
\noindent \textbf{Contributions:}
Chua et al \cite{30chua2023fedpeatLLM} proposed three major contributions: 
i) introducing the FedPEAT framework that combines PEFT and EAT for federated fine-tuning while ensuring data privacy and minimizing computational overhead, 
ii) designing an adaptive control mechanism using a novel Single-Agent Action Branching Proximal Policy Optimization (SABPPO) algorithm for efficient resource allocation and device selection, and 
iii) extending the emulator-adapter approach for collaborative fine-tuning across multiple devices, including server-client collaborative scenarios, to optimize performance in diverse environments.
\end{tcolorbox}
\end{center}

\underline{Challenges:} The work identifies several key challenges, including ensuring the scalability of fine-tuning large models across devices with limited resources; addressing data privacy concerns while maintaining model performance; reducing the computational and communication overhead associated with emulator and adapter transmissions; and developing robust optimization mechanisms for federated environments with heterogeneous hardware capabilities and dynamic network conditions.

\underline{Future Directions:} Future research  includes optimizing the adaptive control mechanism to support more dynamic and heterogeneous scenarios; improving emulator compression techniques to reduce storage and computational demands further; and exploring the integration of multimodal datasets for expanding the applicability of the FedPEAT framework to diverse and complex tasks.

\paragraph*{\textbf{Sani et al \cite{31sani2024futureLLM}, The Future of LLM Pre-training is Federated}}
\underline{Overview and Contributions:} Sani et al \cite{31sani2024futureLLM} proposed Photon which is  {a scalable system enabling collaborative LLM pre-training across organizations using private data and computational resources while maintaining privacy.} This framework addresses key challenges of statistical and hardware heterogeneity in federated systems, enabling robust and scalable model training across geographically distributed nodes. Photon achieves competitive performance with centralized training while significantly reducing communication overhead and accommodating heterogeneous computational capabilities.

\begin{center}
\begin{tcolorbox}
\vspace{-0.05in}
\noindent \textbf{Contributions:}
Sani et al \cite{31sani2024futureLLM} proposed three major contributions: 
i) introducing Photon, a federated system for scalable collaborative pre-training of billion-scale LLMs that enables organizations to leverage private data without sharing it directly, 
ii) demonstrating that federated LLM training scales effectively with model size, with larger models achieving better consensus and generalization, and 
iii) addressing classical challenges of FL, such as statistical heterogeneity and partial participation, through innovative system design and optimization strategies.
\end{tcolorbox}
\end{center}

\underline{Challenges:} The work identifies several key challenges in FL for LLM pre-training, including the scarcity of high-quality decentralized data, which limits model performance and diversity; the need to maintain privacy during collaborative training, adding complexity to optimization; the high computational and communication demands of federated training, which pose scalability issues; and addressing the inherent heterogeneity of hardware and data sources across participants.

\underline{Future Directions:} Future research in federated LLM training includes developing  techniques to enable privacy-preserving training; optimizing federated systems for efficiency by reducing communication costs and improving hardware utilization; and expanding the applicability of FL frameworks to support multimodal data and cross-domain collaborations to enhance the generalizability and robustness of LLMs.

\paragraph*{\textbf{Peng et al \cite{32peng2024fedpftLLM}, FedPFT: Federated Proxy Fine-Tuning of Foundation Models}}
\underline{Overview and Contributions:} Peng et al \cite{32peng2024fedpftLLM} proposed FedPFT, a federated fine-tuning framework designed to enhance Foundation Models (FMs) for downstream tasks while addressing privacy concerns and computational challenges. {The framework employs a Sub-FM Construction Module with layer-wise compression to preserve critical neurons and a Sub-FM Alignment Module with a two-step knowledge distillation process to mitigate gradient errors and facilitate comprehensive FM adaptation.} FedPFT achieves superior performance across diverse NLP  tasks compared to existing methods.

\begin{center}
\begin{tcolorbox}
\vspace{-0.05in}
\noindent \textbf{Contributions:}
Peng et al \cite{32peng2024fedpftLLM} proposed three major contributions: 
i) a sub-FM construction module that uses layer-wise compression emphasizing neuron saliency to ensure effective fine-tuning of all FM layers, 
ii) a sub-FM alignment module utilizing layer-level and neuron-level distillations to minimize gradient discrepancies during federated training, and 
iii) comprehensive evaluations across multiple datasets demonstrating the superiority of FedPFT in privacy-preserving FM fine-tuning.
\end{tcolorbox}
\end{center}

\underline{Challenges:} The work identifies several key challenges, including the difficulty of maintaining comprehensive fine-tuning due to the inherent limitations of sub-FM constructions; the risk of gradient error accumulation, which can degrade model performance; the computational complexity associated with aligning sub-FMs with full FMs during training; and the challenge of ensuring scalability and efficiency in diverse and heterogeneous data settings.

\underline{Future Directions:} Future research in federated fine-tuning includes the exploration of advanced alignment techniques to further minimize gradient errors and improve convergence rates; developing more lightweight and scalable methods for sub-FM construction to reduce computational overhead; and expanding the framework to support multimodal and cross-domain tasks, enhancing its applicability in real-world FL scenarios.

\paragraph*{\textbf{Liu et al \cite{33liu2024timeFFMLLM}, TIME-FFM: Towards LM-Empowered Federated Foundation Model for Time Series Forecasting}}
\underline{Overview and Contributions:} Liu et al \cite{33liu2024timeFFMLLM} proposed TIME-FFM, a Federated Foundation Model for time series forecasting that leverages pretrained Language Models (LMs). This framework transforms time series data into text tokens, utilizes a dynamic prompt adaptation module for cross-domain reasoning, and employs a personalized federated training strategy. TIME-FFM addresses the challenges of cross-domain heterogeneity, modality alignment, and privacy preservation while achieving state-of-the-art forecasting performance in few-shot and zero-shot scenarios.

\begin{center}
\begin{tcolorbox}
\vspace{-0.05in}
\noindent \textbf{Contributions:}
Liu et al \cite{33liu2024timeFFMLLM} proposed three major contributions: 
i) introducing modality alignment by transforming time series data into text tokens for effective cross-modality adaptation using pretrained LMs, 
ii) developing a dynamic prompt adaptation module that constructs domain-specific prompts automatically, enhancing robustness and generalization across domains, and 
iii) implementing a personalized federated training strategy that balances global knowledge sharing with domain-specific predictions using local prediction heads.
\end{tcolorbox}
\end{center}

\underline{Challenges:} The work identifies several key challenges in time series forecasting, including the heterogeneity of cross-domain time series data, which complicates generalization; the necessity to maintain privacy while performing FL, which restricts access to raw data; the computational and communication overhead in federated training, posing scalability concerns; and the alignment of pretrained LMs to seamlessly adapt to the temporal characteristics of time series data.

\underline{Future Directions:} Future research in federated time series forecasting includes enhancing modality alignment techniques to better map time series data into the LM-compatible formats, improving the efficiency and scalability of federated training mechanisms to handle large-scale applications, and exploring the integration of multimodal data for comprehensive forecasting solutions that can further extend the capabilities of TIME-FFM.

\paragraph*{\textbf{Wang et al \cite{35wang2024cycleblackLLM}, Save It All: Enabling Full Parameter Tuning for Federated LLMs via Cycle Block Gradient Descent}}

\underline{Overview and Contributions:} Wang et al \cite{35wang2024cycleblackLLM} proposed FedCyBGD, an innovative training framework designed to enable full parameter tuning of LLMs in an FL   environment. This method addresses significant challenges such as computational, memory, and communication bottlenecks by cyclically updating specific blocks of the model while reducing resource consumption. It allows clients to train specific blocks of LLMs while maintaining the rest of the model in a compressed state, enabling full-parameter tuning on resource-constrained edge devices. The framework achieves competitive performance with centralized approaches while minimizing the associated costs through compression schemes and resource-efficient training paradigms.

\begin{center}
\begin{tcolorbox}
\vspace{-0.05in}
\noindent \textbf{Contributions:}
Wang et al \cite{35wang2024cycleblackLLM} proposed three major contributions: 
i) introducing the FedCyBGD framework, which leverages cyclic block updates to enable full parameter tuning on resource-constrained edge devices, 
ii) designing an effective compression mechanism to minimize the model download cost while preserving the integrity of updated blocks, and 
iii) conducting extensive evaluations across diverse LLMs and datasets, demonstrating substantial improvements in memory, computation, and communication efficiency.
\end{tcolorbox}
\end{center}

\underline{Challenges:} The work identifies several key challenges in FL for LLMs, including the prohibitive computational and memory costs of full parameter tuning on resource-limited edge devices; communication bottlenecks caused by large model transfers during training rounds; the complexity of balancing resource efficiency with maintaining high performance in federated settings; and ensuring privacy preservation without sacrificing model quality.

\underline{Future Directions:} Future research in FL for LLMs includes developing advanced compression and pruning techniques to further reduce resource usage while maintaining performance; enhancing the robustness of cyclic block updates to support diverse tasks and datasets; and exploring the applicability of FedCyBGD in multimodal and real-time FL scenarios to broaden its scope and impact.

\paragraph*{\textbf{Kang et al \cite{37kang2023groundingLLM}, Grounding Foundation Models Through Federated Transfer Learning: A General Framework}}
\underline{Overview and Contributions:} Kang et al \cite{37kang2023groundingLLM} proposed a Federated Transfer Learning-based framework, termed FTL-FM, to adapt Foundation Models (FMs) for domain-specific applications while ensuring privacy and efficiency. This framework formulates FTL-FM settings, categorizes state-of-the-art works, and outlines privacy-preserving and efficiency-enhancing techniques, addressing challenges such as constrained resources, data privacy, model ownership, and heterogeneity. It systematically provides a taxonomy of methods and discusses opportunities for future FTL-FM research.

\begin{center}
\begin{tcolorbox}
\vspace{-0.05in}
\noindent \textbf{Contributions:}
Kang et al \cite{37kang2023groundingLLM} proposed three major contributions: 
i) introducing a comprehensive FTL-FM framework that formulates federated settings, objectives, and knowledge transfer approaches, enabling a structured understanding of grounding FMs, 
ii) constructing a detailed taxonomy to classify and summarize state-of-the-art FTL-FM research based on challenges, techniques, and privacy concerns, and 
iii) reviewing  privacy-preserving and efficiency-improving methods and highlighting their applications in various FTL-FM scenarios.
\end{tcolorbox}
\end{center}

\underline{Challenges:} The work identifies several key challenges, including constrained computational and storage resources that hinder FM deployment; safeguarding data privacy across federated participants while preserving model performance; addressing model ownership issues due to heterogeneous architectures and sizes of server and client models; and maintaining efficient and secure knowledge transfer techniques to adapt FMs for domain-specific applications.

\underline{Future Directions:} Future research in federated transfer learning includes integrating privacy-aware techniques; optimizing efficiency-focused methods such as model compression and decentralized training strategies to reduce computational and communication costs; and exploring innovative co-optimization approaches to simultaneously benefit both FMs and domain models across heterogeneous federated participants.

\paragraph*{\textbf{Yu et al \cite{40yu2023federated}, Federated Foundation Models: Privacy-Preserving and Collaborative Learning for Large Models}}
\underline{Overview and Contributions:} Yu et al \cite{40yu2023federated} proposed Federated Foundation Models (FFMs), a novel paradigm integrating FL   with Foundation Models (FMs) to enable privacy-preserving, decentralized training of large models. This framework addresses challenges such as data privacy, computational costs, and heterogeneity in decentralized data environments. The approach achieves scalable and collaborative model training by incorporating continual learning, federated prompt tuning, and retrieval-augmented generation techniques.

\begin{center}
\begin{tcolorbox}
\vspace{-0.05in}
\noindent \textbf{Contributions:}
Yu et al \cite{40yu2023federated} proposed three major contributions: 
i) integrating FL into the lifespan of FMs to enable privacy-preserving pre-training, fine-tuning, and continual learning, thereby utilizing decentralized private data effectively, 
ii) introducing Federated Retrieval Augmented Generation to enhance FMs’ adaptability and responsiveness using both centralized and decentralized data sources, and 
iii) proposing methods for federated prompt tuning to craft more effective prompts while maintaining data confidentiality.
\end{tcolorbox}
\end{center}

\underline{Challenges:} The work identifies several key challenges in FL for large models, including data privacy concerns that restrict the utilization of sensitive decentralized data; significant computational costs required for FM optimization in FL settings; communication overhead due to frequent model updates between clients and the server; and handling non-IID data distributions across clients, which affect model convergence and performance.

\underline{Future Directions:} Future research in federated foundation models includes advancing edge hardware to support computational demands for FM optimization in FL settings; developing robust privacy-preserving mechanisms; and exploring federated multi-task learning frameworks to simultaneously optimize multiple learning objectives while leveraging distributed data and computational resources.

\paragraph*{\textbf{Zheng et al \cite{42zheng2024safely}, Safely Learning with Private Data: An FL Framework for LLMs}}
\underline{Overview and Contributions:} Zheng et al \cite{42zheng2024safely} proposed FL-GLM, an FL framework designed for securely training LLMs with private data. This framework employs model splitting, encrypted communication, and parallel acceleration techniques to address challenges in data privacy, computational efficiency, and training scalability. FL-GLM achieves comparable performance to centralized models while preserving data confidentiality and reducing training overhead.

\begin{center}
\begin{tcolorbox}
\vspace{-0.05in}
\noindent \textbf{Contributions:}
Zheng et al \cite{42zheng2024safely} proposed three major contributions: 
i) introducing a secure model split design, which keeps sensitive input and output layers on client devices while processing intermediate layers on a server, preventing gradient-based privacy attacks, 
ii) employing client-batch and server-hierarchical parallel training strategies to enhance training efficiency and scalability across diverse computational infrastructures, and 
iii) FL-GLM matches centralized ChatGLM performance on SuperGLUE and summarization tasks.
\end{tcolorbox}
\end{center}

\underline{Challenges:} The work identifies several key challenges, including the vulnerability of embedding gradients to reverse engineering attacks; risking private data leakage; the computational overhead associated with  training LLMs in resource-constrained client environments in a secure fashion; reducing efficiency due to sequential client training, which limits scalability; and the issue in ensuring robust performance under non-IID  data distributions across clients.

\underline{Future Directions:} Future research in FL for LLMs includes developing advanced encryption methods, such as homomorphic encryption and DP, to strengthen data protection during client-server interactions; exploring adaptive client-server training strategies to mitigate the effects of non-IID data and improve model generalization; and expanding the FL-GLM framework to support a wider range of LLM architectures, including multimodal and domain-specific models, to demonstrate versatility and broader applicability.

\paragraph*{\textbf{Qin et al \cite{43qin2024empirical}, Empirical Guidelines for Deploying LLMs onto Resource-constrained Edge Devices}}

\underline{Overview and Contributions:} Qin et al \cite{43qin2024empirical} proposed an empirical framework to address the challenges of deploying LLMs on resource-constrained edge devices. This framework focuses on systematically studying trade-offs among various design factors, such as model size, customization methods, compression techniques, and data constraints. Their findings offer actionable guidelines to optimize performance and efficiency for edge LLM applications, ensuring effective personalization and resource management.

\begin{center}
\begin{tcolorbox}
\vspace{-0.05in}
\noindent \textbf{Contributions:}
Qin et al \cite{43qin2024empirical} proposed three major contributions: 
i) conducting an extensive empirical study to evaluate the effects of customization techniques, including PEFT and retrieval-augmented generation (RAG), on LLM performance under resource constraints, 
ii) identifying the optimal trade-offs between model size, compression methods (distillation, quantization, and pruning), and learning efficiency for different task complexities, and 
iii) offering practical guidelines for fine-tuning strategies, data usage, and compression configurations tailored for edge devices.
\end{tcolorbox}
\end{center}

\underline{Challenges:} The work identifies several key challenges in deploying LLMs on edge devices, including ensuring high performance despite the computational limitations inherent to resource-constrained environments; balancing trade-offs among model size, compression, and task-specific customization methods; addressing overfitting risks associated with limited and non-diverse personalized data; and mitigating the potential challenges of  prolonged fine-tuning periods without improving performance .

\underline{Future Directions:} Future research in edge LLM deployment includes developing advanced compression techniques, such as hybrid strategies combining quantization and distillation, to enhance efficiency; exploring adaptive fine-tuning methods that dynamically adjust to task complexity and data constraints; and creating robust evaluation frameworks for real-time, edge-specific applications to guide the next generation of edge LLMs.

\paragraph*{\textbf{Liu et al \cite{44liu2024resource}, Resource Allocation for Stable LLM Training in Mobile Edge Computing}}
\underline{Overview and Contributions:} Liu et al \cite{44liu2024resource} proposed a collaborative framework to enable efficient and stable LLM training in mobile edge computing (MEC) environments. This framework integrates PEFT methods, distributed resource allocation, and stability optimization to address challenges of energy efficiency, latency, and model robustness. The approach balances computational loads between mobile users and edge servers, optimizing both energy and latency while ensuring model reliability in dynamic settings.

\begin{center}
\begin{tcolorbox}
\vspace{-0.05in}
\noindent \textbf{Contributions:}
Liu et al \cite{44liu2024resource} proposed three major contributions: 
i) designing a collaborative training framework where mobile devices handle initial LLM layers using PEFT techniques, and edge servers manage computationally intensive layers, optimizing energy and latency requirements, 
ii) formulating a multi-objective optimization problem incorporating energy efficiency, delay minimization, and model stability to enhance LLM performance in MEC environments, and 
iii) developing a novel fractional programming technique and a Concave-Convex Procedure (CCCP) for solving complex resource allocation and stability optimization problems, achieving improved performance metrics in simulations.
\end{tcolorbox}
\end{center}

\underline{Challenges:} The work identifies several key challenges in mobile edge computing and federated training. These challenges include balancing computational and communication loads between mobile users and edge servers without compromising latency or energy efficiency;  maintaining model stability when local fine-tuning introduces variability in model performance;  handling the non-convex nature of the optimization problem involving resource allocation, energy, delay, and stability; and  addressing scalability and heterogeneity in mobile environments to support diverse user and application requirements.

\underline{Future Directions:} Future research in MEC-based LLM training includes  developing more advanced PEFT methods tailored to highly dynamic and resource-constrained environments; exploring alternative optimization techniques to address non-convex multi-objective problems with greater efficiency and scalability; and  integrating adaptive mechanisms to dynamically adjust resource allocation and fine-tuning strategies in response to real-time user demands and network conditions.

\paragraph*{\textbf{Li et al \cite{45li2024collm}, CoLLM: A Collaborative LLM Inference Framework for Resource-Constrained Devices}}
\underline{Overview and Contributions:} Li et al \cite{45li2024collm} proposed CoLLM, a collaborative inference framework designed to address the challenges of running LLMs on resource-constrained devices. This framework utilizes tensor parallelism, minimum latency, and adaptive load balancing algorithms to optimize inference latency and energy consumption while enhancing scalability for distributed deployments. CoLLM demonstrates substantial improvements in efficiency, achieving up to 2.3x speedup in inference latency compared to hierarchical methods.

\begin{center}
\begin{tcolorbox}
\vspace{-0.05in}
\noindent \textbf{Contributions:}
Li et al \cite{45li2024collm} proposed three major contributions: 
i) introducing a tensor parallelism-based collaborative inference framework tailored for devices with constrained resources, enabling distributed execution of LLMs, 
ii) developing a minimum latency algorithm to optimize partitioning and workload distribution, minimizing inference delays across devices, and 
iii) designing an adaptive load balancing algorithm to dynamically adjust resource utilization based on device status, achieving efficient energy distribution and extended working times.
\end{tcolorbox}
\end{center}

\underline{Challenges:} The work identifies several key challenges in distributed LLM inference, including the computational and memory constraints of resource-limited devices, which hinder effective deployment of LLMs; managing the communication overhead in distributed systems, which can degrade performance; ensuring efficient load balancing among heterogeneous devices with varying resource capacities; and minimizing energy consumption while maintaining high inference speed and accuracy.

\underline{Future Directions:} Future research in collaborative LLM inference includes the exploration of advanced compression and quantization techniques to further reduce resource requirements; developing more robust and adaptive algorithms for dynamic workload balancing in diverse network conditions; and extending the framework to support real-time applications and multimodal LLMs, enhancing their applicability across various domains.

\paragraph*{\textbf{Pentyala et al \cite{46pentyala2024paft}, PAFT: A Parallel Training Paradigm for Effective LLM Fine-Tuning}}
\underline{Overview and Contributions:} Pentyala et al \cite{46pentyala2024paft} proposed PAFT, a novel parallel training framework for fine-tuning LLMs. This framework concurrently trains supervised fine-tuning (SFT) and preference alignment using distinct datasets and fuses their outputs to mitigate alignment tax and preserve model performance. PAFT incorporates sparsity-inducing techniques such as L1 regularization during SFT, ensuring efficient model merging without significant performance degradation. The approach demonstrates state-of-the-art results on prominent benchmarks, including the Open LLM Leaderboard and AlpacaEval.

\begin{center}
\begin{tcolorbox}
\vspace{-0.05in}
\noindent \textbf{Contributions:}
Pentyala et al \cite{46pentyala2024paft} proposed three major contributions: 
i) introducing a parallel training paradigm to concurrently train SFT and preference alignment, effectively reducing alignment tax and preserving task-specific capabilities, 
ii) developing an L1-norm regularization technique to enhance sparsity in delta parameters, mitigating parameter interference during model merging, and 
iii) conducting extensive evaluations on public benchmarks, achieving rank 1 for 7B and 70B models on the Open LLM Leaderboard.
\end{tcolorbox}
\end{center}

\underline{Challenges:} The work identifies several key challenges in LLM fine-tuning, including mitigating alignment tax, where sequential preference alignment degrades task-specific capabilities of the model;  reducing parameter interference during model merging, which can obscure critical updates and diminish performance;  ensuring scalability and efficiency in parallel training paradigms for large-scale LLMs; and  overcoming limitations in the diversity and quality of datasets used for SFT and preference alignment, which can constrain model generalizability.

\underline{Future Directions:} Future research in LLM fine-tuning includes  exploring advanced sparsity techniques beyond L1-norm regularization to further enhance merging efficiency and performance;  integrating more diverse and high-quality datasets for both SFT and preference alignment to improve robustness and applicability; and  investigating dynamic and adaptive merging strategies that account for task-specific requirements and reduce computational overhead in large-scale deployments.

\paragraph*{\textbf{Li et al \cite{47li2024unity}, Unity is Power: Semi-Asynchronous Collaborative Training of Large-Scale Models with Structured Pruning in Resource-Limited Clients}}

\underline{Overview and Contributions:}  
Li et al \cite{47li2024unity} proposed \textit{Co-S\(^2\)P}, a semi-asynchronous collaborative training framework tailored for large-scale models with structured pruning. This framework integrates data distribution-aware structured pruning and self-distillation mechanisms to optimize training efficiency and accuracy on resource-limited devices. It addresses challenges such as unstructured pruning, heterogeneous submodel architectures, knowledge loss, and stragglers. By leveraging structured pruning at both depth and width dimensions and semi-asynchronous aggregation, \textit{Co-S\(^2\)P} achieves superior convergence rates and resource utilization.

\begin{center}
\begin{tcolorbox}
\vspace{-0.05in}
\noindent \textbf{Contributions:}
Li et al \cite{47li2024unity} proposed three major contributions:  
i) A novel data distribution-aware structured pruning algorithm for both depth and width dimensions, ensuring balanced learning capabilities across heterogeneous devices;  
ii) A self-distillation mechanism enabling cross-block knowledge transfer to mitigate knowledge loss in resource-constrained clients; and  
iii) A semi-asynchronous aggregation strategy that alleviates the straggler problem and enhances convergence efficiency.
\end{tcolorbox}
\end{center}

\underline{Challenges:}  
The work identifies several key challenges in resource-limited collaborative training, including  unstructured pruning methods that fail to optimize memory constraints;  difficulty in adapting submodel architectures to heterogeneous data distributions;  significant knowledge loss due to sparse submodel structures; and training delays caused by computational disparities among clients (stragglers), which hinder convergence and resource utilization.

\underline{Future Directions:}  
Future research in resource-limited collaborative learning includes developing more adaptive and efficient pruning techniques that dynamically adjust to client resource variations; exploring advanced knowledge transfer methods to further minimize information loss in sparse submodels; and enhancing aggregation strategies to ensure scalability and robustness in larger and more heterogeneous client networks.

\paragraph*{\textbf{Markov et al \cite{48markov2023quantized}, Quantized Distributed Training of Large Models with Convergence Guarantees}}
\underline{Overview and Contributions:} Markov et al \cite{48markov2023quantized} proposed QSDP, a quantized extension of the Fully Sharded Data Parallel (FSDP) framework designed to address communication bottlenecks in distributed training of LLMs. This framework incorporates gradient and weight quantization techniques with convergence guarantees to enable efficient and scalable model training. QSDP reduces communication overhead without compromising model accuracy, achieving significant speedups for large-scale training tasks.

\begin{center}
\begin{tcolorbox}
\vspace{-0.05in}
\noindent \textbf{Contributions:}
Markov et al \cite{48markov2023quantized} proposed three major contributions: 
i) a novel weight and gradient quantization mechanism that ensures convergence even in non-convex domains, 
ii) a theoretical analysis establishing convergence guarantees for quantized stochastic gradient descent, and 
iii) an efficient PyTorch-based implementation validated through experiments on GPT-family models with up to 1.3 billion parameters.
\end{tcolorbox}
\end{center}

\underline{Challenges:} The work identifies several key challenges in distributed training, including the communication bottlenecks caused by the frequent transmission of large model weights; the difficulty of ensuring convergence in non-convex optimization when using quantized representations; the trade-off between compression efficiency and model accuracy; and practical challenges in implementing a scalable quantization scheme for large-scale distributed training.

\underline{Future Directions:} Future research in distributed training includes improving quantization techniques for further communication efficiency, exploring adaptive quantization strategies to dynamically optimize bit-width during training, investigating the impact of quantized training on other architectures beyond LLMs, and integrating QSDP with advanced distributed training frameworks to handle larger models and datasets effectively.

\paragraph*{\textbf{Koo et al \cite{49koo2024towards}, Towards Robust and Efficient Federated Low-Rank Adaptation with Heterogeneous Clients}}
\underline{Overview and Contributions:} Koo et al \cite{49koo2024towards} proposed LoRA-A2, an FL framework designed to enhance robustness and efficiency in LoRA  for LLMs. This framework addresses challenges such as data heterogeneity and communication constraints by introducing alternating freeze and adaptive rank selection strategies. LoRA-A2 achieves a substantial reduction in communication costs while maintaining or surpassing baseline performance across heterogeneous data scenarios.

\begin{center}
\begin{tcolorbox}
\vspace{-0.05in}
\noindent \textbf{Contributions:}
Koo et al \cite{49koo2024towards} proposed three major contributions: 
i) introducing LoRA-\( A^2 \), a robust algorithm addressing vulnerabilities in federated LoRA under high heterogeneity and low-rank settings, 
ii) developing an adaptive rank selection strategy to allocate communication resources effectively based on local data importance, and 
iii) enhancing communication efficiency by reducing  transmitted parameters without compromising model performance.
\end{tcolorbox}
\end{center}

\underline{Challenges:} The work identifies several key challenges in federated LoRA, including the aggregation discordance problem that arises during parameter updates in heterogeneous client environments; the difficulty of maintaining performance under low-rank constraints; ensuring communication efficiency without loss of model accuracy; and addressing data heterogeneity that leads to conflicting updates across clients with diverse datasets.

\underline{Future Directions:} Future research in federated LoRA includes extending LoRA-A2 to support complex tasks such as natural language generation to validate its generalizability; exploring scalability by applying the method to larger models such as GPT-style architectures;  and conducting experiments on real-world datasets to evaluate robustness and effectiveness under practical conditions.

\paragraph*{\textbf{Liu et al \cite{51liu2024asynchronous}, Asynchronous Local-SGD Training for Language Modeling}}
\underline{Overview and Contributions:} Liu et al \cite{51liu2024asynchronous} proposed an asynchronous Local-SGD framework tailored for training LLMs on heterogeneous devices. This method addresses the communication and computation challenges of synchronous distributed optimization by enabling model updates as soon as local training completes. Key contributions include strategies to mitigate staleness in gradient updates and optimize device-specific training schedules, achieving performance comparable to synchronous approaches while improving wall-clock efficiency.

\begin{center}
\begin{tcolorbox}
\vspace{-0.05in}
\noindent \textbf{Contributions:}
Liu et al \cite{51liu2024asynchronous} proposed three major contributions: 
i) introducing a delayed Nesterov momentum update mechanism to mitigate gradient staleness and stabilize asynchronous training, 
ii) designing a dynamic local update strategy that adjusts training steps based on device capabilities, thereby balancing computation across heterogeneous devices, and 
iii) integrating these techniques to achieve comparable perplexity performance to synchronous training while significantly reducing overall training time.
\end{tcolorbox}
\end{center}

\underline{Challenges:} The work identifies several key challenges in asynchronous distributed optimization, including the staleness of gradients due to asynchronous updates, which can slow convergence and reduce stability; managing the heterogeneity of device speeds, which causes imbalanced progress across workers; the difficulty in optimizing momentum updates in outer optimization loops; and maintaining efficient communication under varied computational loads.

\underline{Future Directions:} Future research in asynchronous Local-SGD includes developing theoretical guarantees for the proposed methods to better understand convergence behavior; exploring advanced optimization techniques to further reduce staleness and stabilize training under extreme heterogeneity; and expanding the framework to support larger models and more complex datasets, with a focus on minimizing communication overhead while ensuring robust performance.

\paragraph*{\textbf{Yang et al \cite{52yang2024perllm}, Personalized Inference Scheduling with Edge-Cloud Collaboration for Diverse LLM Services}}
\underline{Overview and Contributions:} Yang et al \cite{52yang2024perllm} proposed PerLLM, a personalized inference scheduling framework that leverages edge-cloud collaboration to optimize resource allocation and scheduling for diverse LLM services. The framework addresses challenges such as dynamic resource constraints and diverse service requirements by integrating a combinatorial multi-armed bandit approach with a constraint satisfaction mechanism. The solution achieves significant improvements in throughput, processing time, and energy efficiency, providing tailored responses for diverse user needs.

\begin{center}
\begin{tcolorbox}
\vspace{-0.05in}
\noindent \textbf{Contributions:}
Yang et al \cite{52yang2024perllm} proposed three major contributions: 
i) developing a novel edge-cloud collaborative framework that maximizes processing efficiency for large-scale LLM services while meeting diverse service requirements, 
ii) formulating the scheduling problem as a combinatorial multi-armed bandit problem and introducing a constraint satisfaction upper confidence bound algorithm for effective resource allocation, and 
iii) demonstrating the framework's efficiency through experimental evaluations and  achieving enhanced throughput and reduced energy cost.
\end{tcolorbox}
\end{center}

\underline{Challenges:} The work identifies several key challenges in edge-cloud collaborative inference scheduling, including  managing the diverse service requirements, such as varying response time and processing quality needs, addressing dynamic resource availability and network bandwidth fluctuations, which complicate real-time decision-making,  mitigating the high energy costs and delays associated with cloud processing, and  ensuring scalable and adaptive optimization for large-scale LLM deployments.

\underline{Future Directions:} Future research in edge-cloud collaborative LLM inference includes  exploring novel techniques for dynamic memory optimization to further enhance resource efficiency; integrating continuous learning mechanisms for real-time adaptation to changing workloads and service requirements; and extending the framework to support multi-dimensional resource optimization for hybrid edge-cloud architectures in diverse application scenarios.

\paragraph*{\textbf{Hagemann et al \cite{53hagemann2023efficient}, Efficient Parallelization Layouts for Large-Scale Distributed Model Training}}
\underline{Overview and Contributions:} Hagemann et al \cite{53hagemann2023efficient} proposed a comprehensive framework for parallelizing large-scale distributed model training. This framework is designed to optimize the efficiency of training LLMs by leveraging  FLASHATTENTION-2 and sequence parallelism. The method systematically analyzes various combinations of parallelization strategies, micro-batch sizes, and memory optimizations. Their approach achieves state-of-the-art training efficiency benchmarks across a range of large models.

\begin{center}
\begin{tcolorbox}
\vspace{-0.05in}
\noindent \textbf{Contributions:}
Hagemann et al \cite{53hagemann2023efficient} proposed three major contributions: 
i) a systematic exploration of parallelization strategies, including data, tensor, and pipeline parallelism, to identify optimal configurations, 
ii) the integration and evaluation of advanced memory and compute optimizations such as FLASHATTENTION-2 and activation checkpointing, and 
iii) actionable guidelines for achieving high Model FLOPs Utilization (MFU).
\end{tcolorbox}
\end{center}

\underline{Challenges:} The work identifies several key challenges in large-scale distributed training, including the complexity of balancing multiple parallelization strategies to optimize efficiency, the memory constraints imposed by increasing model sizes and sequence lengths, the need for compatibility between advanced optimizations such as FLASHATTENTION and activation checkpointing, and the manual effort required to fine-tune configurations for specific hardware setups.

\underline{Future Directions:} Future research in large-scale distributed training includes exploring the applicability of these findings to emerging hardware platforms such as NVIDIA H100 GPUs, investigating the benefits of selective activation checkpointing combined with FLASHATTENTION, and extending the analysis to other tasks and architectures, such as vision transformers and multi-modal models. Further, expanding the study to incorporate alternative frameworks and broader scaling laws could further refine the proposed guidelines.

\paragraph*{\textbf{Huang et al \cite{54huang2024distmm}, DistMM: Accelerating Distributed Multimodal Model Training}}
\underline{Overview and Contributions:} Huang et al \cite{54huang2024distmm} proposed DISTMM, a distributed training system tailored for multimodal models to address communication challenges and heterogeneity issues in submodules. This framework incorporates modality-aware partitioning, adaptive load balancing, and pipeline execution optimizations to  improve training efficiency and scalability. DISTMM achieves up to 3.27× speedup over Megatron-LM on various multimodal models by leveraging innovative parallelism strategies and reducing inter-device communication overheads.

\begin{center}
\begin{tcolorbox}
\vspace{-0.05in}
\noindent \textbf{Contributions:}
Huang et al \cite{54huang2024distmm} proposed three major contributions: 
i) a modality-aware partitioner that applies independent parallelism strategies to optimize submodule-level computation, 
ii) a heterogeneity-aware placement manager that minimizes communication overhead by aligning submodule placement with bandwidth availability, and 
iii) a novel pipeline parallelism schedule called DISTMM-Pipe, which supports large batch sizes required for multimodal model quality while avoiding dependency overheads.
\end{tcolorbox}
\end{center}

\underline{Challenges:} The work identifies several key challenges in distributed multimodal model training, including imbalanced computational efficiency due to submodule heterogeneity; inefficient utilization of existing parallelism strategies such as tensor and data parallelism; major communication overheads from inter-device interactions;  and limitations in current pipeline parallelism schedules that degrade model quality when batch sizes are constrained.

\underline{Future Directions:} Future research in distributed multimodal model training includes exploring adaptive algorithms for dynamic submodule partitioning to handle evolving workloads, integrating hardware-aware optimizations for diverse device architectures, developing techniques to support more complex multimodal tasks, and extending DISTMM's principles to integrate with edge-computing systems for real-time applications.

\paragraph*{\textbf{Wang et al \cite{55wang2024efficient}, Efficient Multi-Task Large Model Training via Data Heterogeneity-aware Model Management}}
\underline{Overview and Contributions:} Wang et al \cite{55wang2024efficient} proposed Spindle, a resource-efficient and high-performance system for multi-task multi-modal (MT-MM) large model training. This framework incorporates workload heterogeneity-aware parallelization and dependency-driven execution scheduling to address challenges in training complex MT-MM models. The system optimizes resource allocation and scheduling, minimizing execution overheads and improving GPU utilization. 

\begin{center}
\begin{tcolorbox}
\vspace{-0.05in}
\noindent \textbf{Contributions:}
Wang et al \cite{55wang2024efficient} proposed three major contributions: 
i) a heterogeneity-aware workload parallelization method for optimizing execution across multiple modalities and tasks, 
ii) a dependency-driven execution scheduling strategy to minimize resource wastage and improve efficiency, and 
iii) a general runtime engine to implement stage-based scheduling, addressing execution dependencies across heterogeneous data flows.
\end{tcolorbox}
\end{center}

\underline{Challenges:} The work identifies several challenges in MT-MM model training, including: workload heterogeneity from diverse data flows across modalities and tasks; execution dependency between shared and task-specific model components, which can lead to over-consuming computing resources; and  lack of existing systems that can effectively manage such dependencies while optimizing resource utilization.

\underline{Future Directions:} Future research in MT-MM model training could focus on improving scalability to handle even larger and more complex datasets, incorporating dynamic workload balancing to adapt to real-time changes in task demands, and exploring further optimizations in communication and memory management to reduce overheads in distributed training systems.

\paragraph*{\textbf{Li et al \cite{56li2024tpi}, TPI-LLM: Serving 70B-Scale LLMs Efficiently on Low-Resource Edge Devices}}
\underline{Overview and Contributions:} Li et al \cite{56li2024tpi} proposed TPI-LLM, a tensor-parallel inference framework designed to efficiently serve large-scale LLMs on low-resource edge devices. This framework addresses critical challenges such as memory limitations, high link latency, and privacy concerns in edge environments. TPI-LLM incorporates a sliding window memory scheduler to manage layer weights and a star-based allreduce algorithm to minimize latency. The proposed method achieves significant reductions in memory footprint and inference latency, enabling the deployment of 70B-scale models on edge devices with minimal resources.

\begin{center}
\begin{tcolorbox}
\vspace{-0.05in}
\noindent \textbf{Contributions:}
Li et al \cite{56li2024tpi} proposed three major contributions: 
i) the development of a memory-efficient tensor parallel inference system, TPI-LLM, tailored for low-resource edge devices, 
ii) the implementation of a star-based allreduce algorithm that minimizes link latency compared to conventional methods, and 
iii) the introduction of a sliding window memory scheduler to asynchronously load and unload layer weights, effectively reducing memory usage during inference.
\end{tcolorbox}
\end{center}

\underline{Challenges:} The work identifies several key challenges in serving large-scale LLMs on edge devices, including the severe memory constraints of edge devices lacking GPU support, high link latency that hinders communication efficiency, and the challenges of existing pipeline and tensor parallelism methods in single-user scenarios. Moreover, the need for safeguarding user data privacy while maintaining high computational performance adds further complexity.

\underline{Future Directions:} Future research in this domain includes optimizing computational latency through advanced tensor parallel techniques, enhancing the compatibility of TPI-LLM with a broader range of edge devices, and exploring adaptive scheduling methods to dynamically manage memory and computational resources. Further, refining the star-based allreduce algorithm for diverse network conditions and further reducing the memory footprint for even larger models are promising areas for exploration.

\paragraph*{\textbf{Yang et al \cite{57yang2024meta}, Meta-Learning for Speeding Up Large Model Inference in Decentralized Environments}}
\underline{Overview and Contributions:} Yang et al \cite{57yang2024meta} proposed MetaInf, a meta-learning framework designed to optimize inference acceleration in decentralized environments. This framework automates the selection of acceleration methods by leveraging historical performance data to adaptively choose the most efficient strategies. It addresses challenges such as computational resource constraints and heterogeneity in decentralized systems, significantly enhancing the efficiency and cost-effectiveness of deploying large models. The framework integrates meta-training and online selection mechanisms, providing a scalable and adaptive solution for inference optimization.

\begin{center}
\begin{tcolorbox}
\vspace{-0.05in}
\noindent \textbf{Contributions:}
Yang et al \cite{57yang2024meta} proposed three major contributions: 
i) the introduction of MetaInf, a meta-learning-based framework for dynamic selection of inference acceleration methods in decentralized systems, 
ii) the development of embeddings that encode dataset, model, and hardware characteristics for efficient prediction of optimal acceleration strategies, and 
iii) comprehensive experimental validation demonstrating significant improvements in efficiency and cost-effectiveness over traditional methods.
\end{tcolorbox}
\end{center}

\underline{Challenges:} The work identifies several key challenges in optimizing inference in decentralized environments, including the heterogeneity of hardware configurations, dynamic system conditions, and varying computational constraints. Further, the framework must effectively leverage historical performance data to generalize across diverse operational scenarios while minimizing latency and cost. Ensuring scalability and robustness in the presence of unpredictable workloads and resource limitations also presents significant challenges.

\underline{Future Directions:} Future research in this domain includes extending the MetaInf framework to handle more complex and dynamic decentralized environments, incorporating advanced embeddings for improved prediction accuracy, and exploring integration with newer acceleration techniques such as speculative decoding and adaptive quantization. Further, investigating the applicability of this framework to other domains, such as real-time edge computing and FL, could enhance its impact and scalability.

\paragraph*{\textbf{Hu et al \cite{59hu2023llm}, LLM-Adapters: An Adapter Family for Parameter-Efficient Fine-Tuning of LLMs}}
\underline{Overview and Contributions:} Hu et al \cite{59hu2023llm} proposed LLM-Adapters, a framework designed to integrate and evaluate various PEFT methods for LLMs. This framework supports state-of-the-art open-source LLMs and diverse adapter types, enabling fine-tuning with reduced computational and storage requirements. The approach incorporates adapters such as Series, Parallel, Prompt-based, and Reparameterization-based methods to address fine-tuning challenges, achieving task-specific performance while minimizing resource consumption.

\begin{center}
\begin{tcolorbox}
\vspace{-0.05in}
\noindent \textbf{Contributions:}
Hu et al \cite{59hu2023llm} proposed three major contributions: 
i) development of the LLM-Adapters framework, which integrates diverse PEFT methods into LLMs for a wide range of tasks, 
ii) comprehensive empirical evaluation of adapter placements, configurations, and performance across arithmetic and commonsense reasoning datasets, and 
iii) creation of high-quality fine-tuning datasets, Math10K and Commonsense170K, to enhance PEFT performance in reasoning tasks.
\end{tcolorbox}
\end{center}

\underline{Challenges:} The work identifies several key challenges in PEFT for LLMs, including determining optimal adapter placement and configurations, addressing performance gaps between smaller models and larger LLMs in complex tasks, and the limited availability of high-quality fine-tuning data for reasoning tasks. Further, the study highlights computational constraints in evaluating larger models and the extensive search space in combining multiple adapter methods.

\underline{Future Directions:} Future research in PEFT for LLMs includes exploring the combination of different adapter methods to leverage their complementary strengths, scaling evaluations to larger LLMs for enhanced understanding of adapter capabilities, and developing task-specific PEFT strategies tailored to varying dataset complexities and distributions. Further advancements may focus on refining fine-tuning datasets to address out-of-distribution scenarios and improving the interpretability of PEFT methods.

\paragraph*{\textbf{Gao et al \cite{60gao2024dlora}, Distributed Parameter-Efficient Fine-Tuning Solution for LLM}}
\underline{Overview and Contributions:} Gao et al \cite{60gao2024dlora} proposed DLoRA, a framework designed for distributed PEFT of LLMs across cloud and edge devices. This method addresses challenges in scalability, privacy, and computational efficiency by distributing workloads while protecting sensitive user data. The framework integrates a novel "Kill and Revive" algorithm to dynamically adjust active parameters during fine-tuning, significantly reducing communication and computational costs without compromising accuracy.

\begin{center}
\begin{tcolorbox}
\vspace{-0.05in}
\noindent \textbf{Contributions:}
Gao et al \cite{60gao2024dlora} proposed three major contributions: 
i) introduced the DLoRA framework, enabling distributed PEFT across cloud and edge devices while ensuring user data privacy and scalability, 
ii) developed the Kill and Revive (KR) algorithm to dynamically identify and fine-tune the most sensitive LLM parameters, reducing computational and communication loads, and 
iii) demonstrated  efficiency gains both in terms of reduction in computational load and reduction in communication overhead.
\end{tcolorbox}
\end{center}

\underline{Challenges:} The work identifies several key challenges in deploying distributed PEFT for LLMs, including privacy concerns associated with transmitting user data to cloud servers, scalability issues arising from increasing user-specific parameter sets, and the computational limitations of edge devices. Further, maintaining fine-tuning accuracy while reducing active parameter counts poses significant algorithmic and resource-management challenges.

\underline{Future Directions:} Future research in distributed PEFT includes optimizing the KR algorithm to adaptively balance accuracy and efficiency across a wider range of tasks, integrating quantization techniques to further reduce communication overhead, and extending the DLoRA framework to support real-time applications and multi-modal models. Furthermore, exploring its applicability to FL scenarios with multiple edge devices remains an open avenue.

\paragraph*{\textbf{Gao et al \cite{61gao2024fedpt}, FedPT: Federated Proxy-Tuning of LLMs on Resource-Constrained Edge Devices}}
\underline{Overview and Contributions:} Gao et al \cite{61gao2024fedpt} proposed Federated Proxy-Tuning (FedPT), a lightweight framework for the federated fine-tuning of LLMs in resource-constrained environments. This method addresses challenges of computation, communication, and memory overhead by leveraging small proxy models to collaboratively tune LLMs without requiring direct access to their parameters. FedPT achieves comparable performance to direct fine-tuning while preserving privacy and reducing resource demands through proxy tuning and knowledge distillation.

\begin{center}
\begin{tcolorbox}
\vspace{-0.05in}
\noindent \textbf{Contributions:}
Gao et al \cite{61gao2024fedpt} proposed three major contributions: 
i) a novel proxy-tuning approach for FL, enabling resource-constrained devices to fine-tune large black-box LLMs collaboratively, 
ii) an efficient framework that combines proxy tuning and knowledge distillation to mitigate the challenges of tuning  LLMs on edge devices, and 
iii) extensive experiments demonstrating the significant reductions in computational, memory, and communication overhead while maintaining competitive performance.
\end{tcolorbox}
\end{center}

\underline{Challenges:} The work identifies several key challenges in federated fine-tuning of LLMs, including the high memory requirements for tuning large models, which exceed the capacities of most edge devices; substantial computational overhead, causing extended fine-tuning sessions even with GPU-equipped devices; communication overhead during FL rounds; and the lack of white-box access to proprietary LLM parameters, limiting conventional tuning approaches.

\underline{Future Directions:} Future research  includes exploring advanced proxy-tuning strategies for improved scalability, incorporating more sophisticated privacy-preserving mechanisms, addressing the scalability of the FedPT framework for larger-scale deployments, and optimizing the trade-offs between tuning complexity and model performance across diverse application domains.

\paragraph*{\textbf{Ghiasvand et al \cite{62ghiasvand2024communication}, Communication-Efficient and Tensorized Federated Fine-Tuning of LLMs}}
\underline{Overview and Contributions:} Ghiasvand et al \cite{62ghiasvand2024communication} proposed FedTT and FedTT+, federated fine-tuning frameworks for LLMs, designed to enhance communication efficiency and robustness to data heterogeneity. These methods integrate tensorized adapters into client models, reducing the number of trainable parameters and communication overhead while maintaining competitive performance. FedTT+ further minimizes parameter updates and improves robustness under heterogeneous client data distributions.

\begin{center}
\begin{tcolorbox}
\vspace{-0.05in}
\noindent \textbf{Contributions:}
Ghiasvand et al \cite{62ghiasvand2024communication} proposed three major contributions: 
i) a tensorized adapter design that integrates tensor-train decomposition for parameter-efficient federated fine-tuning, 
ii) the FedTT framework, which reduces communication overhead by up to 10× compared to LoRA-based approaches, and 
iii) the FedTT+ extension, which adaptively freezes tensor factors to address data heterogeneity while further decreasing the number of trainable parameters.
\end{tcolorbox}
\end{center}

\underline{Challenges:} The work identifies several key challenges in federated fine-tuning of LLMs, including the communication overhead associated with transmitting model updates across clients, the computational costs of training LLMs on edge devices, and the significant performance degradation due to non-i.i.d. data distributions across clients. These challenges are particularly pronounced in large-scale and resource-constrained FL scenarios.

\underline{Future Directions:} Future research in federated fine-tuning includes exploring strategies to enhance system heterogeneity adaptability, such as dynamically adjusting tensor ranks to match client's computational resources. Investigating privacy preservation mechanisms to quantify and mitigate potential information leakage during communication rounds is also critical. Finally, extending the FedTT framework to support multi-task learning and generalization to diverse model architectures presents an exciting avenue for future work.

\paragraph*{\textbf{Qin et al \cite{63qin2024federated}, Federated Data-Efficient Instruction Tuning for LLMs}}
\underline{Overview and Contributions:} Qin et al \cite{63qin2024federated} proposed FedHDS, a federated hierarchical data selection framework aimed at improving data efficiency during instruction tuning of LLMs. This framework combines hierarchical clustering and coreset selection to reduce data redundancy at intra-client and inter-client levels, ensuring computational efficiency while maintaining model generalization. The approach achieves significant reductions in computational costs and data usage by fusing features from multiple Transformer layers and implementing a dual-layer clustering mechanism.

\begin{center}
\begin{tcolorbox}
\vspace{-0.05in}
\noindent \textbf{Contributions:}
Qin et al \cite{63qin2024federated} proposed three major contributions: 
i) the FedHDS framework, which introduces a two-layer data selection mechanism to reduce intra-client and inter-client data redundancy, 
ii) an innovative feature fusion technique that leverages multi-layer Transformer outputs for clustering, improving the quality of selected coresets, and 
iii)  evaluations demonstrating significant computational cost savings  and improved Rouge-L scores on unseen tasks compared to baseline methods.
\end{tcolorbox}
\end{center}

\underline{Challenges:} The work identifies several key challenges in FedIT of LLMs, including overcoming excessive computational overhead caused by traversing all local data during training, mitigating overfitting to non-IID client data, and addressing the limitations of existing coreset selection methods in FL contexts, such as suboptimal data representations and poor compatibility with privacy-preserving frameworks.

\underline{Future Directions:} Future research in FedIT includes exploring advanced clustering algorithms for more efficient coreset selection, integrating privacy-preserving mechanisms in a more robust fashion, and extending the framework to other modalities or heterogeneous client settings to enhance its applicability across diverse FL scenarios.

\paragraph*{\textbf{Zhao et al \cite{64zhao2024frag}, FRAG: Toward Federated Vector Database Management for Collaborative and Secure Retrieval-Augmented Generation}}
\underline{Overview and Contributions:} Zhao et al \cite{64zhao2024frag} proposed FRAG, a federated framework for secure and efficient Retrieval-Augmented Generation (RAG) systems. This framework allows mutually-distrusted parties to collaboratively perform encrypted Approximate \(k\)-Nearest Neighbor (ANN) searches without revealing sensitive data. The approach incorporates Single-Key Homomorphic Encryption (SK-MHE) for simplified key management and Multiplicative Caching (MC) for computational efficiency, achieving robust privacy guarantees and scalability.

\begin{center}
\begin{tcolorbox}
\vspace{-0.05in}
\noindent \textbf{Contributions:}
Zhao et al \cite{64zhao2024frag} proposed three major contributions: 
i) the development of the Federated Retrieval-Augmented Generation (FRAG) framework enabling secure collaborative ANN searches while preserving privacy; 
ii) introduction of the Single-Key Homomorphic Encryption (SK-MHE) protocol for simplifying encryption management with strong security guarantees; and 
iii) creation of the Multiplicative Caching (MC) protocol, significantly optimizing homomorphic encryption efficiency for large-scale federated environments.
\end{tcolorbox}
\end{center}

\underline{Challenges:} The work identifies several key challenges in federated database management, including maintaining strong privacy guarantees in environments with mutually-distrusted parties; ensuring computational efficiency despite the high overheads of homomorphic encryption; simplifying key management without compromising security; and achieving scalability to support large-scale real-time RAG systems.

\underline{Future Directions:} Future research in federated retrieval systems includes exploring more efficient cryptographic techniques for reducing computational and communication overhead; improving scalability for handling larger and more dynamic datasets; extending the framework to support additional query types and modalities; and addressing potential vulnerabilities in multi-party environments to enhance robustness against adversarial attacks.

\paragraph*{\textbf{Elbakary et al \cite{65elbakary2024mira}, MIRA: A Method of Federated Multi-Task Learning for LLMs}}
\underline{Overview and Contributions:} Elbakary et al \cite{65elbakary2024mira} proposed MIRA, a federated multi-task learning framework designed for efficient fine-tuning of LLMs across heterogeneous clients. This framework leverages the structure of client-specific tasks and data distributions while incorporating parameter-efficient techniques to mitigate computational and communication overhead. MIRA effectively balances local task performance and global model optimization through regularization mechanisms.

\begin{center}
\begin{tcolorbox}
\vspace{-0.05in}
\noindent \textbf{Contributions:}
Elbakary et al \cite{65elbakary2024mira} proposed three major contributions: 
i) a federated multi-task learning (FMTL) paradigm enabling personalized model tuning for heterogeneous clients while maintaining task similarities, 
ii) utilization of LoRA to reduce the computational and communication costs of LLM fine-tuning, and 
iii) comprehensive experimental evaluations demonstrating improved local and global performances using datasets such as Natural Instructions and Dolly-15k.
\end{tcolorbox}
\end{center}

\underline{Challenges:} The work identifies several key challenges in federated multi-task learning for LLMs, including the difficulty of aligning client models with varying task distributions, the high memory and computational requirements of gradient-based fine-tuning methods, and the communication overhead associated with parameter synchronization across clients. Further, managing task similarity effectively in a federated setting remains a complex issue due to the diverse nature of client tasks.

\underline{Future Directions:} Future research in federated fine-tuning of LLMs includes exploring advanced mechanisms for task alignment across heterogeneous clients, improving the scalability of the framework for larger networks and more diverse tasks, and integrating more efficient parameter-efficient fine-tuning techniques. Further, incorporating methods to dynamically adapt to client-specific requirements and reducing the communication cost further can enhance the usability of FL for LLMs.

\paragraph*{\textbf{Qi et al \cite{66qi2024fdlora}, FDLoRA: Personalized FL of LLM via Dual LoRA Tuning}}

\underline{Overview and Contributions:} Qi et al \cite{66qi2024fdlora} proposed FDLoRA, a framework that leverages dual LoRA modules in a personalized FL (PFL) setting to enhance the customization of LLMs. This framework integrates personalized and global LoRA modules to address challenges of data and system heterogeneity while maintaining low communication and computational overhead. It achieves enhanced client-specific and global knowledge fusion through an adaptive fusion algorithm, demonstrating robust performance in diverse non-IID data scenarios such as log-based anomaly detection and medical diagnosis.

\begin{center}
\begin{tcolorbox}
\vspace{-0.05in}
\noindent \textbf{Contributions:}
Qi et al \cite{66qi2024fdlora} proposed three major contributions:  
i) the introduction of FDLoRA, the first framework integrating dual LoRA modules with PFL for LLM customization,  
ii) the development of a gradient-free adaptive fusion approach to effectively combine global and personalized knowledge, and  
iii) demonstrated superior performance and stability in log analysis and medical diagnosis while reducing computation and communication costs.

\end{tcolorbox}
\end{center}

\underline{Challenges:} The work identifies several key challenges in the domain of FL for LLMs, including the heterogeneity of client data distributions (non-IID) that can lead to performance degradation, computational constraints posed by the large parameter size of LLMs, communication constraints in synchronizing personalized and global models, and the difficulty in aligning client-specific and global knowledge without overfitting or loss of generalization.

\underline{Future Directions:} Future research in this domain includes expanding the application of FDLoRA to broader industrial scenarios such as natural language understanding and sentiment analysis, optimizing the framework's robustness and scalability through advanced communication compression and model optimization techniques, and investigating hybrid personalization strategies that balance global and localized knowledge for improved client-specific outcomes.

\paragraph*{\textbf{Mahla et al \cite{67yao2024sharingLLM}, Why Gradient Subspace? Identifying and Mitigating LoRA's Bottlenecks in Federated Fine-Tuning of LLMs}}
\underline{Overview and Contributions:} Mahla et al \cite{67yao2024sharingLLM} proposed a comprehensive analysis and novel approach to improve federated fine-tuning of LLMs by identifying and mitigating the limitations of LoRA  techniques. This framework incorporates direct weight aggregation and gradient subspace optimization to address challenges such as rank inflation and suboptimal learning in federated environments. The proposed solution achieves enhanced generalization and reduced computational overhead through the use of the GaLore optimizer \cite{zhao2024galore} and targeted fine-tuning of model parameters.

\begin{center}
\begin{tcolorbox}
\vspace{-0.05in}
\noindent \textbf{Contributions:}
Yao et al \cite{67yao2024sharingLLM} proposed three major contributions: 
i) identified the limitations of LoRA-based FL, including rank inflation and issues of capturing heterogeneous client distributions, 
ii) introduced direct weight aggregation combined with the GaLore optimizer to achieve superior performance and tighter generalization bounds, and 
iii) developed the FedFTG framework, which leverages efficient gradient subspace learning for federated fine-tuning of both LLMs and Vision Transformers (ViTs).
\end{tcolorbox}
\end{center}

\underline{Challenges:} The work identifies several key challenges in federated fine-tuning of LLMs, including the rank inflation issue inherent in LoRA methods, which limits their ability to learn effectively in non-IID data settings; the quadratic risk bounds of certain LoRA frameworks that hinder their stability and scalability; and the computational inefficiencies associated with PEFT methods in resource-constrained environments. These issues collectively degrade performance and generalization across clients with diverse data distributions.

\underline{Future Directions:} Future research in federated fine-tuning includes developing adaptive aggregation strategies that can dynamically address client heterogeneity and mitigate rank inflation, designing more memory-efficient optimization algorithms for large-scale federated deployments, and exploring advanced gradient subspace learning techniques to further enhance generalization and reduce computational overhead. Further, expanding the applicability of the proposed methods to multimodal data and exploring their integration with privacy-preserving mechanisms are promising areas for further investigation.

\paragraph*{\textbf{Yang et al \cite{68yang2024research}, Research on Key Technologies for Cross-Cloud Federated Training of LLMs}}
\underline{Overview and Contributions:} Yang et al \cite{68yang2024research} proposed a comprehensive framework for cross-cloud federated training, a paradigm aimed at addressing the computational and resource limitations of single-cloud platforms for LLMs. This framework incorporates advanced data partitioning, communication optimization, and model aggregation techniques to enhance the efficiency and scalability of training across heterogeneous cloud environments. The proposed approach focused on data security and privacy and demonstrates improved training performance and cost efficiency via  experimental validation.

\begin{center}
\begin{tcolorbox}
\vspace{-0.05in}
\noindent \textbf{Contributions:}
Yang et al \cite{68yang2024research} proposed three major contributions: 
i) the development of dynamic data partitioning and distribution strategies to balance computational load and enhance efficiency across cloud platforms, 
ii) the introduction of optimized cross-cloud communication protocols and asynchronous mechanisms to minimize latency and improve scalability, and 
iii) the refinement of model aggregation algorithms, including dynamic weighting and gradient-based methods, to ensure robust global model convergence in heterogeneous environments.
\end{tcolorbox}
\end{center}

\underline{Challenges:} The work identifies several key challenges in cross-cloud federated training, including handling the heterogeneity of cloud platforms with varying computational capabilities, ensuring efficient and secure data partitioning and communication, and overcoming convergence and synchronization issues in model aggregation. These challenges are compounded by the need to maintain high training accuracy while minimizing resource and communication overheads.

\underline{Future Directions:} Future research in cross-cloud federated training includes exploring more advanced model aggregation algorithms to address extreme data heterogeneity, developing adaptive communication protocols for dynamic network environments, and integrating stronger data security measures such as post-quantum encryption.

\paragraph*{\textbf{Ouyang et al \cite{69ouyang2024pluto}, Pluto and Charon: A Time and Memory Efficient Collaborative Edge AI Framework for Personal LLMs Fine-Tuning}}
\underline{Overview and Contributions:} Ouyang et al \cite{69ouyang2024pluto} proposed Pluto and Charon (PAC), a collaborative edge AI framework designed for efficient fine-tuning of personal LLMs in resource-constrained environments. This framework incorporates Parallel Adapters and an activation cache mechanism to achieve parameter, time, and memory efficiency. PAC leverages hybrid data and pipeline parallelism, pooling edge devices into a collective resource to break the limitations of single-device setups. The approach achieves significant performance improvements in training speed and memory usage.

\begin{center}
\begin{tcolorbox}
\vspace{-0.05in}
\noindent \textbf{Contributions:}
{ Ouyang et al \cite{69ouyang2024pluto} proposed three major contributions: i) Conducted detailed studies on PEFT techniques for edge devices, revealing inefficiencies in resource utilization, ii) designed a resource-efficient fine-tuning method using Parallel Adapters, reducing backward passes through the LLM backbone and optimizing computational demands, and iii) proposed PAC, a collaborative edge AI framework combining activation caching and hybrid parallelism with empirical results showing  memory reduction compared to existing methods.}

\end{tcolorbox}
\end{center}

\underline{Challenges:} The work identifies several key challenges in edge-based fine-tuning of LLMs, including the limited computational capabilities of edge devices, the memory constraints that hinder hosting and training of large models, inefficiencies in existing PEFT techniques for edge scenarios, and the lack of robust frameworks to exploit collaborative edge resources for in-situ training.

\underline{Future Directions:} Future research in this domain includes optimizing PAC for broader LLM architectures beyond the tested models, reducing the latency introduced by inter-device communication in hybrid parallel setups, exploring adaptive methods to dynamically balance resource allocation across heterogeneous edge devices, and expanding PAC's capabilities to incorporate privacy-preserving mechanisms for user data during fine-tuning.

\paragraph*{\textbf{Tang et al \cite{70tang2024fusionllm}, FusionLLM: A Decentralized LLM Training System on Geo-distributed GPUs with Adaptive Compression}}
\underline{Overview and Contributions:} Tang et al \cite{70tang2024fusionllm} proposed FusionLLM, a decentralized training system aimed at training LLMs using geo-distributed GPUs. This system addresses challenges such as hardware heterogeneity, low-bandwidth communication, and system scalability by employing a directed acyclic graph (DAG) representation of models and adaptive compression mechanisms. FusionLLM ensures efficient workload distribution, seamless integration across diverse hardware, and improved communication efficiency, achieving significant speedups compared to existing methods.

\begin{center}
\begin{tcolorbox}
\vspace{-0.05in}
\noindent \textbf{Contributions:}
Tang et al \cite{70tang2024fusionllm} proposed three major contributions: 
i) the introduction of an OP-DAG-based model representation and runtime executor for flexible and heterogeneous system support, 
ii) the development of an OP-Fence scheduler and workload estimator to optimize resource allocation and throughput, and 
iii) the design of the AdaTopK compressor for adaptive communication compression, addressing low-bandwidth challenges.
\end{tcolorbox}
\end{center}

\underline{Challenges:} The work identifies several key challenges in decentralized training, including the need for remote automatic differentiation (RAD) support over the Internet, heterogeneity in hardware and software configurations, low network bandwidth causing communication bottlenecks, and the straggler problem caused by variable device performance. These challenges necessitate innovative system designs and optimizations to enhance efficiency and scalability.

\underline{Future Directions:} Future research in decentralized training systems includes exploring advanced compression algorithms to further reduce communication costs, improving robustness against network instability, enhancing security mechanisms to protect data privacy, and developing dynamic resource allocation strategies to optimize heterogeneous device utilization.

\paragraph*{\textbf{Sheng et al \cite{71sheng2024hybridflow}, HybridFlow: A Flexible and Efficient RLHF Framework}}
\underline{Overview and Contributions:} Sheng et al \cite{71sheng2024hybridflow} proposed HybridFlow, a hybrid framework designed to optimize RLHF in LLMs. This framework integrates single-controller and multi-controller paradigms to enhance flexibility in dataflow representation and efficiency in computation. It incorporates hierarchical APIs and a 3D-HybridEngine to tackle challenges such as intra-node computation, inter-node communication, and model resharding. The proposed approach demonstrates significant throughput improvements over state-of-the-art RLHF systems by optimizing GPU resource allocation and minimizing memory redundancy.

\begin{center}
\begin{tcolorbox}
\vspace{-0.05in}
\noindent \textbf{Contributions:}
Sheng et al \cite{71sheng2024hybridflow} proposed three major contributions: 
i) a hierarchical hybrid programming model that decouples data computation and transfer, allowing for efficient execution and modular design, 
ii) the 3D-HybridEngine that supports zero-redundancy transitions between training and generation phases, significantly improving throughput and memory efficiency, and 
iii) an auto-mapping algorithm for optimal GPU placement and workload distribution, enabling scalability and resource optimization in diverse RLHF workflows.
\end{tcolorbox}
\end{center}

\underline{Challenges:} The work identifies several key challenges in RLHF frameworks, including inflexibility in representing diverse RLHF dataflows, inefficiencies in resource utilization due to redundant memory usage and communication overhead, and the lack of modularity in current multi-controller paradigms. Further, balancing heterogeneous workloads across models and optimizing transitions between training and generation phases in RLHF workflows pose significant computational challenges.

\underline{Future Directions:} Future research in RLHF frameworks includes extending HybridFlow to support emerging RLHF algorithms, integrating advanced parallelism strategies for larger LLMs, and exploring adaptive resource allocation techniques to enhance efficiency in dynamic and heterogeneous computational environments. Furthermore, efforts could focus on improving ease of use for developers by refining APIs and automating configuration processes for complex RLHF workflows.

\paragraph*{\textbf{Shen et al \cite{72shen2024edgeqat}, EdgeQAT: Entropy and Distribution Guided Quantization-Aware Training for the Acceleration of Lightweight LLMs on the Edge}}
\underline{Overview and Contributions:} Shen et al \cite{72shen2024edgeqat} proposed EdgeQAT, an entropy and distribution-guided Quantization-Aware Training (QAT) framework designed to enhance the deployment of lightweight LLMs on edge devices. This framework tackles the challenges of performance degradation due to activation quantization by introducing innovative optimization techniques. It incorporates entropy maximization and distribution-based loss functions to mitigate quantization errors, achieving state-of-the-art accuracy with significant inference acceleration. Furthermore, the method dynamically allocates bit widths to tokens based on their importance, ensuring computational efficiency.

\begin{center}
\begin{tcolorbox}
\vspace{-0.05in}
\noindent \textbf{Contributions:}
Shen et al \cite{72shen2024edgeqat} proposed three major contributions: 
i) the entropy and distribution-guided quantization method, which minimizes information distortion in quantized query, key, and attention maps for enhanced accuracy, 
ii) a token importance-aware adaptive quantization technique, dynamically assigning bit widths to activations based on token significance for improved efficiency, and 
iii) practical deployment of EdgeQAT across various edge devices, achieving up to 2.37× speedup compared to FP16 counterparts.
\end{tcolorbox}
\end{center}

\underline{Challenges:} The work identifies several key challenges in deploying lightweight LLMs on edge devices, including significant performance degradation caused by activation quantization, particularly in the self-attention modules. These issues arise due to pronounced outliers in activation distributions and information distortion in quantized attention maps. Further, existing QAT methods often lack optimization for both weights and activations, leading to limited acceleration and compatibility with edge device constraints.

\underline{Future Directions:} Future research in this domain includes extending the EdgeQAT framework to larger LLMs while addressing the associated computational and data requirements. Enhancing support for mixed-precision quantization schemes to further improve accuracy and efficiency across diverse edge hardware platforms is another key direction. Finally, exploring alternative adaptive strategies for quantization-aware token importance evaluation could yield further performance gains and energy efficiency improvements.

\paragraph*{\textbf{Wang et al \cite{73wang2023privatelora}, PRIVATELoRA: For Efficient Privacy-Preserving LLM}}
\underline{Overview and Contributions:} Wang et al \cite{73wang2023privatelora} proposed PrivateLoRA, a novel 
 PEFT framework that enables efficient and privacy-preserving use of LLMs on edge devices. This framework employs low-rank residual transmission to achieve significant communication reduction, thereby addressing the challenges of bandwidth limitations and computational disparities between cloud and edge devices. PrivateLoRA preserves data locality by ensuring sensitive computations and personal data remain on edge devices while leveraging cloud resources for model scalability. Through extensive evaluations, the framework demonstrates both efficiency and adaptability in distributed LLM personalization tasks.

\begin{center}
\begin{tcolorbox}
\vspace{-0.05in}
\noindent \textbf{Contributions:}
Wang et al \cite{73wang2023privatelora} proposed three major contributions: 
i) a novel low-rank residual transmission technique that reduces communication overhead, 
ii) a privacy-preserving framework that maintains data locality by storing sensitive computations and personal data on edge devices, and 
iii) comprehensive empirical evaluations demonstrating improved throughput and task performance compared to device-only solutions and comparable results to GPU-based benchmarks.
\end{tcolorbox}
\end{center}

\underline{Challenges:} The work identifies several key challenges in distributed LLM fine-tuning, including balancing the computational workload between resource-constrained edge devices and powerful cloud infrastructure, minimizing communication overhead to adapt to bandwidth limitations, and maintaining competitive task performance while preserving data locality. The framework also faces the challenge of scaling to larger models without compromising efficiency or privacy guarantees.

\underline{Future Directions:} Future research in privacy-preserving LLM frameworks includes optimizing adaptive mechanisms for even larger-scale models, exploring integration with complementary techniques such as quantization to enhance resource efficiency, and expanding the applicability of PrivateLoRA to diverse edge device architectures. Further, addressing dynamic workload distribution and real-time adaptability in heterogeneous systems presents a promising direction.

\paragraph*{\textbf{Du et al \cite{74du2024distributed}, Distributed Foundation Models for Multi-Modal Learning in 6G Wireless Networks}}
\underline{Overview and Contributions:} Du et al \cite{74du2024distributed} proposed a distributed training architecture for multi-modal foundation models (FMs) in the context of 6G wireless networks. This framework leverages pipeline parallelism, data parallelism, and multi-modal learning to address the challenges of data scarcity and computational resource limitations. By integrating wireless communication technologies, it achieves efficient aggregation of distributed resources for training and inference, enabling the sustainable development of large-scale FMs within 6G networks.

\begin{center}
\begin{tcolorbox}
\vspace{-0.05in}
\noindent \textbf{Contributions:}
Du et al \cite{74du2024distributed} proposed three major contributions: 
i) a pipeline parallelism mechanism that compresses activations and gradients to overcome communication bottlenecks in unstable wireless links, 
ii) the incorporation of FL with over-the-air computation (AirComp) to integrate communication and computation for efficient gradient aggregation, and 
iii) the development of a multi-modal learning framework that aligns data modalities to achieve seamless integration of NLP and CV tasks in wireless networks.
\end{tcolorbox}
\end{center}

\underline{Challenges:} The work identifies several key challenges in the distributed training of foundation models, including the heterogeneity of data collected from wireless devices, non-IID and multi-modal data distribution, the instability of wireless links causing device disconnections, and the heterogeneity in computational, storage, and bandwidth resources across devices. These challenges necessitate advanced management strategies for resource allocation, model partitioning, and data aggregation.

\underline{Future Directions:} Future research in distributed training of multi-modal FMs includes the exploration of more efficient compression algorithms to minimize data transmission, the development of adaptive model partitioning techniques to achieve load balancing, and the optimization of communication resource management to improve channel quality.Advancing multi-modal learning to integrate diverse data modalities and addressing synchronization and interference issues in FL environments are critical areas for further research.

\paragraph*{\textbf{Chen et al \cite{75chen2023confidant}, Confidant: Customizing Transformer-based LLMs via Collaborative Edge Training}}
\underline{Overview and Contributions:} Chen et al \cite{75chen2023confidant} proposed \textit{Confidant}, a multi-backend collaborative training framework designed for fine-tuning transformer-based LLMs on mobile edge devices. This framework addresses the challenges of memory and computation constraints by partitioning LLMs into sub-models and employing pipeline parallel training. Confidant incorporates a novel backend scheduler to optimize resource utilization across heterogeneous hardware, achieving substantial memory reductions and speedups in training.

\begin{center}
\begin{tcolorbox}
\vspace{-0.05in}
\noindent \textbf{Contributions:}
Chen et al \cite{75chen2023confidant} proposed three major contributions: 
i) a pipeline-parallel training mechanism that partitions LLMs into sub-models for efficient distributed training across multiple devices, 
ii) a backend scheduler that allocates attention heads to heterogeneous compute backends, such as CPUs and GPUs, optimizing resource utilization and training speed, and 
iii) a practical implementation leveraging mobile frameworks such as MNN, with both memory reduction and speedup during LLM fine-tuning.
\end{tcolorbox}
\end{center}

\underline{Challenges:} The work identifies several challenges in deploying and fine-tuning LLMs on resource-constrained edge devices. Key challenges include high memory requirements during training, which exceed the capacity of most mobile devices, limited computational capabilities of mobile GPUs and CPUs, and inefficiencies in existing mobile frameworks that hinder multi-backend utilization. Ensuring convergence while distributing training across devices and addressing the dynamic resource availability further complicates the design.

\underline{Future Directions:} Future research in this domain includes developing adaptive memory management techniques to handle dynamically varying memory availability on mobile devices, designing energy-aware training algorithms with fault tolerance to ensure reliability on battery-powered devices, and exploring cross-framework implementations to support collaborative training across diverse edge platforms, such as integrating PyTorch for laptops and MNN for mobile devices.

\paragraph*{\textbf{Huang et al \cite{76huang2024edgellm}, EdgeLLM: A Highly Efficient CPU-FPGA Heterogeneous Edge Accelerator for LLMs}}
\underline{Overview and Contributions:} Huang et al \cite{76huang2024edgellm} proposed EdgeLLM, a CPU-FPGA heterogeneous acceleration framework tailored for LLMs to enhance computational efficiency on edge devices. The framework integrates universal data parallelism, specialized hardware operators, and dynamic compilation to address challenges such as computational complexity, operator diversity, and memory management. This design achieves significant improvements in throughput and energy efficiency over existing GPU and FPGA accelerators.

\begin{center}
\begin{tcolorbox}
\vspace{-0.05in}
\noindent \textbf{Contributions:}
Huang et al \cite{76huang2024edgellm} proposed three major contributions: 
i) a universal data format that streamlines operator execution and ensures compatibility across diverse AI algorithms, 
ii) custom-designed FP16*INT4 computational units with optimization methods including structured sparsity and group-vector systolic arrays to maximize throughput and minimize resource consumption, and 
iii) an end-to-end dynamic compilation scheme to efficiently map LLMs on a CPU-FPGA heterogeneous system, achieving superior performance over state-of-the-art solutions.
\end{tcolorbox}
\end{center}

\underline{Challenges:} The work identifies key challenges in deploying LLMs on edge devices, including managing diverse operator types that require different data formats, addressing the massive parameter sizes and computational complexities of LLMs, and optimizing the operator compilation system for seamless execution. These issues are compounded by the constraints of edge devices, such as limited memory and computational resources.

\underline{Future Directions:} Future research in this domain could explore further optimization of sparsity patterns to reduce computational overhead, integration of more advanced quantization techniques to enhance efficiency, and development of adaptive runtime systems to dynamically balance workloads between CPUs and FPGAs. Additional efforts could focus on generalizing the framework for multimodal AI models and extending compatibility with emerging hardware architectures.

\paragraph*{\textbf{Xu et al \cite{77xu2024hethub}, HETHUB: A Distributed Training System with Heterogeneous Cluster for Large-Scale Models}}
\underline{Overview and Contributions:} Xu et al \cite{77xu2024hethub} proposed HETHUB, a distributed training system designed to enable efficient training of large-scale models in heterogeneous GPU-accelerator clusters. This framework integrates a distributed unified communicator, a performance predictor, and an automatic parallel planner to overcome challenges in heterogeneous computing environments.

\begin{center}
\begin{tcolorbox}
\vspace{-0.05in}
\noindent \textbf{Contributions:}
Xu et al \cite{77xu2024hethub} proposed three major contributions: 
i) the introduction of a distributed unified communicator to facilitate communication between different types of GPU-accelerators, improving compatibility and efficiency; 
ii) a distributed performance predictor that enables efficient evaluation and strategy optimization for training in heterogeneous environments; and 
iii) an automatic parallel planner that identifies optimal distributed strategies to maximize resource utilization and minimize training time.
\end{tcolorbox}
\end{center}

\underline{Challenges:} The work identifies several key challenges in training large-scale models within heterogeneous clusters, including  communication issues caused by incompatible communication libraries across different GPU-accelerators;  the complexity of designing distributed training strategies due to computational and storage differences among hardware types; and  maintaining model accuracy across heterogeneous setups due to operator inconsistencies.

\underline{Future Directions:} Future research in heterogeneous distributed training systems could explore improving compatibility and integration across an even broader range of GPU-accelerators and accelerators;  enhancing the scalability of distributed training systems to accommodate clusters with thousands of heterogeneous nodes; and  developing novel algorithms that reduce communication overhead and further optimize the balance between computation and data transfer.

\paragraph*{\textbf{Shuai et al \cite{78shuai2024mitigating}, Align as Ideal Cross-Modal Alignment Binding for Federated Medical Vision-Language Pre-training}}
\underline{Overview and Contributions:} Shuai et al \cite{78shuai2024mitigating} proposed the Federated Align as IDeal (FedAID) framework, a novel strategy designed to address data heterogeneity in federated medical vision-language pre-training. This framework incorporates guidance-based regularization and distributionally robust optimization to bind local cross-modal alignments to an unbiased representation space. It achieves improved multimodal learning by reducing distortions in aggregated features and enabling robust cross-modal alignment across heterogeneous datasets.

\begin{center}
\begin{tcolorbox}
\vspace{-0.05in}
\noindent \textbf{Contributions:}
Shuai et al \cite{78shuai2024mitigating} proposed three major contributions: 
i) a robust framework for federated medical vision-language pre-training that mitigates data heterogeneity and ensures privacy preservation, 
ii) a guidance-based regularization approach to bind local cross-modal alignments to unbiased representations, and 
iii) the integration of distributionally robust optimization to tackle feature distortions and enhance downstream task performance.
\end{tcolorbox}
\end{center}

\underline{Challenges:} The work identifies several key challenges in federated vision-language pre-training, including addressing the significant data heterogeneity among client datasets, ensuring effective cross-modal alignment while maintaining privacy, mitigating distortions in aggregated representations caused by local training on heterogeneous datasets, and optimizing the model for diverse and worst-case data distributions.

\underline{Future Directions:} Future research in federated medical vision-language pre-training includes exploring more advanced methods to handle extreme heterogeneity in client datasets, integrating domain-specific knowledge to enhance cross-modal alignment, developing techniques to further improve the transferability of pre-trained models to diverse downstream tasks, and investigating privacy-preserving mechanisms to incorporate data diversity without compromising user data privacy.

\paragraph*{\textbf{Fang et al \cite{80fang2024automated}, Automated Federated Pipeline for Parameter-Efficient Fine-Tuning of LLMs}}
\underline{Overview and Contributions:} Fang et al \cite{80fang2024automated} proposed FedPipe, an automated federated pipeline designed to fine-tune LLMs efficiently for downstream tasks while accommodating heterogeneous edge servers. This framework integrates PEFT with FL   to address challenges related to resource constraints and performance variability across edge devices. FedPipe employs innovative techniques such as low-rank adapters, mixed-integer linear programming (MILP) optimization, and quantization-aware training to optimize training and communication overheads.

\begin{center}
\begin{tcolorbox}
\vspace{-0.05in}
\noindent \textbf{Contributions:}
Fang et al \cite{80fang2024automated} proposed three major contributions: 
i) the first automated federated pipeline for fine-tuning LLMs tailored to heterogeneous edge servers, 
ii) an MILP-based optimization strategy to identify critical trainable weights and adapt configurations dynamically, and 
iii) a quantization-aware training approach to reduce memory and computation costs while maintaining high accuracy.
\end{tcolorbox}
\end{center}

\underline{Challenges:} The work identifies several key challenges in federated fine-tuning of LLMs, including the straggler problem caused by heterogeneous computing resources, the difficulty of identifying and prioritizing critical trainable weights for adapter construction, and the challenge of aligning model configurations with varying memory budgets across edge servers. Addressing these challenges necessitates novel approaches in both optimization and system design.

\underline{Future Directions:} Future research in federated fine-tuning of LLMs includes exploring adaptive algorithms for dynamic resource allocation across heterogeneous servers, enhancing the scalability of the pipeline for larger LLMs, and integrating advanced quantization techniques to further reduce memory usage without compromising accuracy. Further, developing privacy-preserving mechanisms tailored for medical and other sensitive data applications remains an open avenue for investigation.

\paragraph*{\textbf{Fan et al \cite{81fan2023fate}, FATE-LLM: A Industrial Grade FL Framework for LLMs}}
\underline{Overview and Contributions:} Fan et al \cite{81fan2023fate} proposed FATE-LLM, an industrial-grade FL framework tailored for LLMs. This framework addresses critical challenges of high computational demands and data privacy in training LLMs. FATE-LLM incorporates parameter-efficient fine-tuning techniques and privacy-preserving mechanisms to enhance training efficiency while safeguarding intellectual property and data security. The approach is designed to facilitate FL for heterogeneous and homogeneous LLMs, thereby enabling broader adoption across industries with varying computational resources.

\begin{center}
\begin{tcolorbox}
\vspace{-0.05in}
\noindent \textbf{Contributions:}
Fan et al \cite{81fan2023fate} proposed three major contributions: 
i) enabling FL for heterogeneous and homogeneous LLMs using  methods such as LoRA and P-Tuning-v2; 
ii) incorporating federated intellectual property protection to secure model ownership during training; and 
iii) implementing privacy-preserving mechanisms to protect sensitive data and enhance the applicability of LLMs in industrial settings.
\end{tcolorbox}
\end{center}

\underline{Challenges:} The work identifies several key challenges, including the need to balance computational efficiency with the complexity of LLMs, the difficulty in reconciling varying model architectures during federated training, and the challenge of maintaining privacy while sharing model updates across diverse clients. Additional challenges include scalability to large industrial datasets and ensuring model ownership security in collaborative training settings.

\underline{Future Directions:} Future research in this domain includes improving methods for reconciling heterogeneous LLM architectures, enabling fine-tuning across organizations without compromising data and model privacy, developing efficient methods to protect user prompts during inference, and exploring the application of the framework in vertical FL scenarios to broaden its industrial impact.

\paragraph*{\textbf{Woisetschläger et al \cite{82woisetschlager2024federated}, Federated Fine-Tuning of LLMs on the Very Edge: The Good, The Bad, The Ugly}}
\underline{Overview and Contributions:} Woisetschläger et al \cite{82woisetschlager2024federated} proposed a comprehensive study on federated fine-tuning of LLMs at the network edge, focusing on computational, energy, and communication efficiencies. This framework incorporates state-of-the-art techniques such as PEFT and energy-efficient FL   optimizers to address challenges such as limited resources on edge devices and high communication costs. The approach achieves substantial improvements in system scalability, energy efficiency, and model convergence through rigorous benchmarking and novel optimization techniques.

\begin{center}
\begin{tcolorbox}
\vspace{-0.05in}
\noindent \textbf{Contributions:}
Woisetschlager et al \cite{82woisetschlager2024federated} proposed three major contributions: 
i) a systematic benchmarking of computational capabilities of embedded devices compared to data center accelerators, highlighting computational bottlenecks; 
ii) the introduction of energy efficiency metrics to complement traditional FLOP-based efficiency measures, enabling real-time system monitoring; and 
iii) an evaluation of FL optimizers for fine-tuning foundation models, showing significant energy savings and faster convergence with adaptive optimizers (e.g., FedAdamW).
\end{tcolorbox}
\end{center}

\underline{Challenges:} The work identifies several key challenges in federated fine-tuning of LLMs, including the limited memory bandwidth of edge devices compared to data center GPUs, which leads to severe computational bottlenecks; the high communication cost of transmitting large model updates in edge settings; the need for energy-efficient operations to comply with emerging AI regulations; and the difficulty in adapting optimization techniques designed for centralized systems to distributed and resource-constrained FL environments.

\underline{Future Directions:} Future research in federated fine-tuning of LLMs includes developing more communication-efficient FL methods, such as gradient compression techniques and selective update mechanisms, to reduce energy consumption and latency; designing adaptive parameter-efficient training strategies to mitigate data heterogeneity in FL; exploring lightweight hardware-specific optimizations for edge devices; and integrating regulatory-compliant energy monitoring systems to enhance the practicality and scalability of FL in real-world applications.

\paragraph*{\textbf{Fu et al \cite{84fu2024serverlessllm}, ServerlessLLM: Low-Latency Serverless Inference for LLMs}}
\underline{Overview and Contributions:} Fu et al \cite{84fu2024serverlessllm} proposed ServerlessLLM, a distributed system designed to support low-latency serverless inference for LLMs. This system leverages multi-tier storage hierarchies and novel checkpoint management strategies to minimize inference latency. By optimizing checkpoint storage and migration processes, ServerlessLLM enables efficient model execution in GPU clusters while addressing the challenges of cold-start latency and resource contention. The approach achieves significant improvements in inference efficiency and response time across various LLM workloads.

\begin{center}
\begin{tcolorbox}
\vspace{-0.05in}
\noindent \textbf{Contributions:}
Fu et al \cite{84fu2024serverlessllm} proposed three major contributions: 
i) fast multi-tier checkpoint loading, utilizing a novel format and pipeline to maximize GPU server storage bandwidth, 
ii) efficient live migration of LLM inference, reducing network traffic and maintaining minimal user disruption, and 
iii) startup-time-optimized model scheduling, incorporating cost models for latency-preserving and locality-aware server selection.
\end{tcolorbox}
\end{center}

\underline{Challenges:} The work identifies several key challenges in serverless inference for LLMs, such as prolonged cold-start latencies caused by the large size of LLM checkpoints and unpredictable inference durations. These issues can be further increased   due to  the interactive and resource-intensive nature of LLM workloads.

\underline{Future Directions:} Future research in serverless LLM inference includes exploring more scalable and cost-efficient multi-tier storage designs, improving live migration methods to handle dynamic workload distributions effectively, and developing advanced scheduling algorithms that integrate workload prediction and resource optimization for heterogeneous GPU clusters.

\paragraph*{\textbf{Xin et al \cite{85xinimmediate}, Immediate Communication for Distributed AI Tasks}}
\underline{Overview and Contributions:} Xin et al \cite{85xinimmediate} proposed DistFuse, a framework designed to optimize communication in distributed AI tasks, particularly for LLMs. This method addresses the significant communication overhead encountered in multi-GPU systems by facilitating immediate communication during computation. DistFuse leverages fine-grained overlapping of computation and communication, improving hardware utilization and reducing inference latency. 

\begin{center}
\begin{tcolorbox}
\vspace{-0.05in}
\noindent \textbf{Contributions:}
Xin et al \cite{85xinimmediate} proposed three major contributions: 
i) They introduced the concept of immediate communication to reduce GPU communication latency by initiating communication as soon as part of the data is ready, 
ii) They developed a tile-wise communication strategy for GeMM and All-reduce operations, enabling fine-grained overlapping of dependent operations, and 
iii) They implemented DistFuse as a prototype that showed significant performance improvements both by reducing communication latency and a speedup in Llama 3 inference latency.
\end{tcolorbox}
\end{center}

\underline{Challenges:} The work identifies several key challenges in distributed AI tasks, including the difficulty in applying fine-grained overlapping techniques when data dependencies exist between operations; the challenge of achieving efficient scheduling for triggering communication without excessive overhead; the complexity of adapting communication libraries such as NCCL to support tile-wise communication; and the lack of compiler support to enable the simultaneous execution of dependent operations across different hardware resources.

\underline{Future Directions:} Future research in distributed AI tasks includes extending the approach to multi-node systems to address communication bottlenecks in large-scale environments; developing automated systems for detecting overlap opportunities and dynamically selecting tile sizes based on model execution plans; and exploring the co-design of computation and communication kernels to improve instruction-level parallelism and maximize model FLOPs utilization (MFU).

\paragraph*{\textbf{Shabani et al \cite{86shabani2024harnessing}, Harnessing FL for LLM Fine-Tuning: A Distributed Approach}}
\underline{Overview and Contributions:} Shabani et al \cite{86shabani2024harnessing} proposed a distributed framework that integrates FL   with PEFT techniques to fine-tune LLMs while preserving data privacy. Their method incorporates LoRA to reduce computational load and mitigate the need for extensive data sharing. The framework, tested with the T5 model on a summarization task, ensures data privacy and significantly reduces computational costs while achieving comparable performance to centralized training.

\begin{center}
\begin{tcolorbox}
\vspace{-0.05in}
\noindent \textbf{Contributions:}
Shabani et al \cite{86shabani2024harnessing} proposed three major contributions: 
i) They introduced an FL framework integrated with LoRA, which allows fine-tuning LLMs while keeping data localized and preserving privacy, 
ii) They demonstrated the use of PEFT techniques, specifically LoRA, to achieve efficient fine-tuning by modifying only a small fraction of the model’s parameters, and 
iii) They conducted extensive experiments on the T5 model in both centralized and federated settings, showing that their approach effectively reduces computational demands without sacrificing performance.
\end{tcolorbox}
\end{center}

\underline{Challenges:} The work identifies several key challenges in FL for LLM fine-tuning, including the difficulty of ensuring model convergence when data is distributed and non-iid across clients; the challenge of managing communication overhead due to frequent updates and large model sizes; the issue of maintaining privacy while exchanging model updates without leaking sensitive information; and the need for effective incentive mechanisms to ensure active client participation in FL.

\underline{Future Directions:} Future research in FL for LLM fine-tuning includes exploring more efficient communication protocols to reduce the bandwidth cost; investigating  privacy-preserving techniques  to mitigate information leakage during model updates; and developing adaptive FL strategies that can handle diverse data distributions and varying client resources to improve system scalability and fairness.

\paragraph*{\textbf{Yan et al \cite{87yan2024lightweight}, Lightweight Unsupervised FL with Pretrained VLM}}
\underline{Overview and Contributions:} Yan et al \cite{87yan2024lightweight} proposed a novel approach for lightweight unsupervised FL using pretrained VLMs, specifically CLIP. Their method addresses the significant challenges of data annotation and resource limitations in FL by leveraging zero-shot predictions from CLIP's pretrained image and text encoders. This approach enables model training with minimal client-side computation and communication overhead, while also improving model performance compared to standard zero-shot predictions. Through a self-training mechanism and class-balanced data generation, their method outperforms traditional supervised FL methods even under challenging data heterogeneity.

\begin{center}
\begin{tcolorbox}
\vspace{-0.05in}
\noindent \textbf{Contributions:}
Yan et al \cite{87yan2024lightweight} proposed three major contributions: 
i) They introduced a lightweight unsupervised FL framework by utilizing the pretrained CLIP model, which performs zero-shot predictions on unlabeled data and enables efficient FL with minimal resource usage, 
ii) They developed a self-training strategy with evolving pseudo-labels to refine the initial low-confidence predictions from the CLIP model, significantly improving model performance, and 
iii) They proposed a class-balanced data generation approach to address data heterogeneity and class imbalance, enhancing the convergence of FL models under non-i.i.d. data distributions.
\end{tcolorbox}
\end{center}

\underline{Challenges:} The work identifies several key challenges in unsupervised FL, including the generation of high-quality pseudo-labels from low-confidence initial predictions made by the pretrained model, the inherent heterogeneity of data across clients that leads to biased local models, and the difficulties in maintaining computational and communication efficiency when training on edge devices with limited resources. Further, the need to balance the model's performance while addressing class imbalances in local data presents a significant challenge in ensuring robust learning across clients with diverse datasets.

\underline{Future Directions:} Future research in unsupervised FL could explore enhancing the robustness of pseudo-label generation by integrating more sophisticated self-training techniques or semi-supervised learning approaches. Further investigations could focus on improving the scalability of the framework for deployment on larger federated networks and exploring alternative pretrained VLMs for domain-specific applications. Moreover, additional work is needed to refine data generation techniques to mitigate the effects of severe data heterogeneity and improve the efficiency of synthetic instance sampling methods.

\paragraph*{\textbf{Zeng et al \cite{88zeng2024fair}, Fair Federated Learning with Biased VLMs}}
\underline{Overview and Contributions:} Zeng et al \cite{88zeng2024fair} proposed a fairness-aware FL framework called Fair Federated Deep Visual Prompting (FF-DVP), designed to address inherent bias in pretrained VLMs such as  CLIP. This framework specifically targets the bias amplified by data heterogeneity in FL   applications. FF-DVP integrates deep visual prompting to debias the CLIP model while maintaining domain-generalized feature extraction and learning client-specific fairness constraints. The proposed framework enhances model fairness without compromising accuracy and can be adapted to various parameter-efficient fine-tuning methods such as LoRA and adapter-based approaches.

\begin{center}
\begin{tcolorbox}
\vspace{-0.05in}
\noindent \textbf{Contributions:}
Zeng et al \cite{88zeng2024fair} proposed three major contributions: 
i) They introduced FF-DVP, a fairness-aware framework that debiases pretrained VLMs   in the context of FL, addressing bias due to data heterogeneity; 
ii) They designed fairness-aware deep visual prompting (DVP) and modality-fused classification heads to learn client-specific knowledge and fairness constraints, ensuring group fairness in FL models; 
iii) They demonstrated the effectiveness of FF-DVP on face attribute recognition (FAR) tasks, showing that it significantly improves fairness and model convergence compared to existing baselines.
\end{tcolorbox}
\end{center}

\underline{Challenges:} The work identifies several key challenges in FL with biased VLMs, including mitigating the demographic bias inherent in pretrained models, which can be exacerbated by the non-i.i.d. data distributions in FL. Further, addressing the computational cost and communication overhead associated with the large scale of pretrained VLMs, while still ensuring fairness across clients, presents significant hurdles. The need to balance model fairness with performance and ensure the stability of federated training under high data heterogeneity further complicates the implementation of fair FL systems.

\underline{Future Directions:} Future research in fair FL includes developing more robust debiasing techniques for other types of bias in VLMs, such as stereotypical or malicious content biases, to ensure fairness in broader applications. Further work could explore more efficient methods for integrating fairness constraints into FL while minimizing the computational and communication overhead. Further, extending FF-DVP to larger federated networks with diverse client data distributions, while ensuring scalability and maintaining fairness, remains a critical area for future research.

\paragraph*{\textbf{Raje  \cite{89raje2024communication}, Communication-Efficient LLM Training for FL}}
\underline{Overview and Contributions:} Raje et al \cite{89raje2024communication} proposed FLoSS, a framework designed to improve the communication efficiency of FL   when fine-tuning LLMs. The approach combines LoRA with sparsity to reduce both local computation and communication costs during federated training. By focusing sparsity on model updates during the download and upload phases, FLoSS reduces communication costs by up to 10x compared to vanilla LoRA, achieving comparable or better utility across several tasks. The framework also recommends heuristics to optimize LoRA rank and sparsity configurations based on communication budgets.

\begin{center}
\begin{tcolorbox}
\vspace{-0.05in}
\noindent \textbf{Contributions:}
Raje  \cite{89raje2024communication} proposed three major contributions: 
i) They introduced the novel concept of applying unstructured sparsity to LoRA during FL, enhancing communication efficiency without significant loss in model utility, 
ii) They proposed FLoSS, a method that applies top-k sparsity only to the communication phases of federated training, reducing communication costs by up to 10x while preserving model performance, and 
iii) They developed and tested heuristics to select optimal LoRA rank and sparsity ratios, providing guidelines for communication-efficient federated training under varying network conditions and communication budgets.
\end{tcolorbox}
\end{center}

\underline{Challenges:} The work identifies several key challenges in FL with LLMs, including the difficulty of balancing communication efficiency and model performance, particularly with resource-constrained clients. Sparsity must be carefully applied to avoid significant degradation in model accuracy, and the variability in communication bandwidth (such as faster download speeds and slower upload speeds) complicates the optimal selection of sparsity ratios. Further, ensuring the scalability of FL when dealing with heterogeneous data distributions across clients poses a challenge in maintaining model convergence and fairness.

\underline{Future Directions:} Future research in communication-efficient FL for LLMs includes exploring methods to automate the selection of optimal sparsity configurations and LoRA rank values, reducing the need for manual tuning. Further studies could also examine the integration of additional compression techniques, such as quantization or more sophisticated pruning strategies, to improve the efficiency of FL with large models. Moreover, expanding the FLoSS framework to support privacy-preserving FL to enhance its applicability in data-sensitive settings.

\paragraph*{\textbf{Sadeepa et al \cite{90sadeepa2024disllm}, DisLLM: Distributed LLMs for Privacy Assurance in Resource-Constrained Environments}}
\underline{Overview and Contributions:} Sadeepa et al \cite{90sadeepa2024disllm} introduced DisLLM, a distributed learning framework that enhances privacy preservation and computational efficiency for fine-tuning LLMs in resource-constrained environments. The approach combines Splitfed Learning (SFL), FL , and LoRA to efficiently manage model training while safeguarding sensitive client-side data. DisLLM splits the model into client-side and server-side components, allowing sensitive data to remain local, and incorporates Local DP (LDP) for added security. Experimental evaluations show that DisLLM provides comparable accuracy to centralized models, with enhanced privacy protection and optimized resource utilization.

\begin{center}
\begin{tcolorbox}
\vspace{-0.05in}
\noindent \textbf{Contributions:}
{ Sadeepa et al \cite{90sadeepa2024disllm} proposed three major contributions: i) Introduced a novel framework, DisLLM, integrating Split Learning (SL) and FL for privacy preservation and effective computational load distribution in LLMs, ii) developed a resource-efficient method that splits LLMs at a predefined cut layer, optimizing computational balance across devices and servers while safeguarding data privacy, and iii) performed comprehensive evaluations to demonstrate the method’s effectiveness in maintaining performance, efficiency, and privacy across diverse application scenarios.}

\end{tcolorbox}
\end{center}

\underline{Challenges:} The work identifies several key challenges in distributed LLM training, including the computational complexity of fine-tuning large models on resource-constrained client devices, the difficulty of preserving privacy while processing sensitive data, and the challenge of balancing model accuracy with computational efficiency. Further, selecting the appropriate model split point and optimizing the trade-off between privacy levels and model performance pose significant challenges in maintaining a scalable and effective solution for real-world applications.

\underline{Future Directions:} Future research in distributed LLM training includes exploring dynamic resource allocation techniques that adjust the client-server model split based on resource availability and privacy needs. Further improvements could focus on adaptive privacy mechanisms that vary based on the sensitivity of the data being processed, ensuring optimal privacy protection. Moreover, optimizing the communication efficiency between clients and serverscab benefit in terms of scaling DisLLM to larger distributed networks with more clients.

\subsection{\textbf{Popov et al \cite{91popov2018distributed}, Distributed Fine-Tuning of Language Models on Private Data}}
\underline{Overview and Contributions:} Popov et al \cite{91popov2018distributed} proposed a distributed fine-tuning framework for language models, designed to adapt general models to private user data while preserving the quality on the original dataset and minimizing communication costs. This framework integrates techniques such as random rehearsal and model averaging to address challenges such as  catastrophic forgetting and privacy. The proposed approach achieves substantial improvements in user language modeling, with notable reductions in perplexity and gains in keystroke saving rates through efficient on-device and server-side updates.

\begin{center}
\begin{tcolorbox}
\vspace{-0.05in}
\noindent \textbf{Contributions:}
Popov et al \cite{91popov2018distributed} proposed three major contributions: 
i) an efficient distributed fine-tuning procedure resilient to catastrophic forgetting, 
ii) a comparative analysis of communication-efficient strategies for on-device training and model updates, and 
iii) an experimental framework to evaluate the DP of distributed LM training.
\end{tcolorbox}
\end{center}

\underline{Challenges:} The work identifies several key challenges in distributed fine-tuning of language models, including addressing catastrophic forgetting during on-device updates; ensuring communication efficiency under resource-constrained environments; achieving balance between personalization and maintaining general model quality; and preserving user privacy in the presence of adversarial risks.

\underline{Future Directions:} Future research in distributed fine-tuning includes enhancing communication efficiency by combining computation-heavy and gradient-compression strategies,  exploring robust architectures for low-resource devices, and extending experimental frameworks to evaluate privacy in diverse adversarial scenarios.

\paragraph*{\textbf{Lin et al \cite{92lin2024splitlora}, SplitLoRA: A Split Parameter-Efficient Fine-Tuning Framework for LLMs}}
\underline{Overview and Contributions:} Lin et al \cite{92lin2024splitlora} proposed SplitLoRA, a novel framework designed to enable efficient fine-tuning of LLMs in resource-constrained environments. The framework integrates Split FL (SFL) with the PEFT technique LoRA to balance computational efficiency and model accuracy. SplitLoRA partitions the model into client-side and server-side components, offloading most of the computational workload to the central server while minimizing the data transmitted between clients and servers. Experimental results demonstrate that SplitLoRA achieves high training performance with significantly reduced computation and communication costs compared to traditional LLM fine-tuning paradigms.

\begin{center}
\begin{tcolorbox}
\vspace{-0.05in}
\noindent \textbf{Contributions:}
Lin et al \cite{92lin2024splitlora} proposed three major contributions: 
i) They introduced SplitLoRA, a framework that combines Split Learning and FL to optimize the training efficiency of LLMs, 
ii) They incorporated LoRA, a PEFT method, into the SplitLoRA framework, enabling resource-constrained devices to fine-tune LLMs with minimal computational overhead, and 
iii) They demonstrated the effectiveness of SplitLoRA through extensive experiments, showing that it outperforms both centralized and FL frameworks in terms of accuracy, convergence speed, and resource efficiency.
\end{tcolorbox}
\end{center}

\underline{Challenges:} The work identifies several key challenges in distributed fine-tuning of LLMs, including the difficulty of selecting optimal model splitting points that balance the workload between client and server, and the challenge of handling data heterogeneity across clients without degrading model performance. Furthermore, ensuring the efficient communication of model updates and activations, while minimizing the communication overhead in resource-constrained environments, presents a significant challenge. There is also the issue of privacy preservation, as sensitive data from clients must not be exposed during training.

\underline{Future Directions:} Future research in SplitLoRA and distributed LLM fine-tuning includes exploring optimal strategies for selecting model splitting points that account for varying client resources and data distributions. Further work could focus on extending SplitLoRA to support heterogeneous computing environments, where clients have vastly different computational capabilities.

\paragraph*{\textbf{Wu et al \cite{93wu2024cg}, CG-FedLLM: How to Compress Gradients in Federated Fine-Tuning for LLMs}}
\underline{Overview and Contributions:} Wu et al \cite{93wu2024cg} introduced CG-FedLLM, a federated fine-tuning framework designed to reduce communication costs for LLMs. The framework employs an AutoEncoder to compress and reconstruct gradients, enabling efficient communication between clients and the server in federated settings. The approach includes two stages: Temporal-ensemble Gradient-Aware Pre-training (TGAP) and Federated AutoEncoder Fine-tuning (FAF), which together ensure accurate gradient reconstruction and enhanced model performance. Experimental evaluations demonstrate that CG-FedLLM achieves superior performance on benchmarks while maintaining privacy and communication efficiency.

\begin{center}
\begin{tcolorbox}
\vspace{-0.05in}
\noindent \textbf{Contributions:}
Wu et al \cite{93wu2024cg} proposed three major contributions: 
i) They developed CG-FedLLM, a novel FL framework that integrates an AutoEncoder for gradient compression and reconstruction, significantly reducing communication costs, 
ii) They introduced TGAP, a pre-training strategy that identifies characteristic gradients for efficient AutoEncoder training, avoiding the need for dynamic retraining during FL, and 
iii) They validated the effectiveness of CG-FedLLM through extensive evaluations, showing improvements in both performance and robustness over traditional centralized and federated fine-tuning approaches.
\end{tcolorbox}
\end{center}

\underline{Challenges:} The work identifies several key challenges, including the high communication costs of transmitting large gradients in federated fine-tuning of LLMs, the difficulty in designing an AutoEncoder that can accurately reconstruct gradients with minimal loss, and ensuring compatibility with privacy-preserving techniques such as DP. Further, balancing gradient compression rates with performance retention remains a non-trivial task, requiring careful tuning of the AutoEncoder's structure.

\underline{Future Directions:} Future research in federated fine-tuning of LLMs includes exploring more efficient AutoEncoder architectures to further reduce training and inference costs and investigating adaptive compression strategies that dynamically adjust to varying client and network conditions. Moreover, extending CG-FedLLM to support heterogeneous client environments and diverse model architectures could enhance its applicability in real-world FL scenarios.

\paragraph*{\textbf{Sun et al \cite{94sun2024improving}, Improving LoRA in Privacy-Preserving FL}}
\underline{Overview and Contributions:} Sun et al \cite{94sun2024improving} introduced FFA-LoRA (Federated Freeze A LoRA), an efficient and privacy-preserving modification to the LoRA framework for FL. This approach addresses challenges related to data heterogeneity, noisy gradients due to DP, and hyper-parameter sensitivity, particularly the scaling factor. FFA-LoRA reduces communication costs by halving trainable parameters and improves compatibility with FL aggregation methods. Experimental results demonstrate superior performance and stability compared to standard LoRA across diverse privacy-preserving tasks.

\begin{center}
\begin{tcolorbox}
\vspace{-0.05in}
\noindent \textbf{Contributions:}
Sun et al \cite{94sun2024improving} proposed three major contributions: 
i) They identified key discordances in applying LoRA within privacy-preserving FL, including issues with noisy gradients, data heterogeneity, and hyper-parameter optimization, 
ii) They developed FFA-LoRA, which freezes one of the low-rank matrices during training to address these discordances and halve communication costs, and 
iii) They validated the effectiveness of FFA-LoRA through extensive experiments, demonstrating improved performance and stability over traditional LoRA.
\end{tcolorbox}
\end{center}

\underline{Challenges:} The work identifies several key challenges in federated fine-tuning with LoRA, including the amplification of noise when combined with DP-SGD due to LoRA's semi-quadratic nature, the susceptibility of performance to hyper-parameters such as the scaling factor, and significant aggregation errors caused by heterogeneous data across federated clients. These issues lead to instability and suboptimal convergence in privacy-preserving FL environments.

\underline{Future Directions:} Future research in federated fine-tuning includes exploring advanced initialization methods for low-rank matrices to further enhance stability, investigating adaptive strategies for dynamically tuning hyper-parameters in federated environments, and extending FFA-LoRA to other PEFT methods. Further, integrating FFA-LoRA with novel privacy-preserving mechanisms and supporting heterogeneous architectures across clients remain important directions.

\paragraph*{\textbf{Bai et al \cite{95bai2024federated}, Federated Fine-Tuning of LLMs under Heterogeneous Tasks and Client Resources}}

\underline{Overview and Contributions:} Bai et al \cite{95bai2024federated} proposed FlexLoRA, a parameter-efficient federated fine-tuning framework tailored for LLMs. This framework addresses resource and task heterogeneity among clients, enhancing generalization while preserving privacy. FlexLoRA employs a novel aggregation mechanism based on Singular Value Decomposition (SVD) to integrate client contributions of varying computational capabilities. Experimental validation demonstrates superior generalization and scalability compared to state-of-the-art FL   approaches.

\begin{center}
\begin{tcolorbox}
\vspace{-0.05in}
\noindent \textbf{Contributions:}
{Bai et al \cite{95bai2024federated} offered three major contributions: i) Proposes FlexLoRA, a scalable method leveraging client resources to enhance global model generalization, backed by theoretical and empirical evidence, ii) demonstrates federated fine-tuning of billion-sized LLMs across diverse NLP tasks in large-scale, heterogeneous settings, and iii) explores the relationship between LoRA rank adjustments, client heterogeneity, and resource distributions, providing insights for cross-device FL.}

\end{tcolorbox}
\end{center}

\underline{Challenges:} The work identifies several key challenges in federated fine-tuning of LLMs, including the ``bucket effect'' limiting contributions from resource-rich clients; ensuring effective task generalization despite client heterogeneity; managing computational overhead introduced by dynamic rank adjustments; and maintaining scalability across thousands of clients with diverse resource and data distributions.

\underline{Future Directions:} Future research in federated fine-tuning of LLMs includes exploring lightweight alternatives to Singular Value Decomposition to reduce computational costs; investigating adaptive strategies for rank tuning to further enhance scalability; extending FlexLoRA to support multi-modal federated tasks; and addressing data and resource imbalances to maximize the participation of low-resource clients without compromising model performance.

\paragraph*{\textbf{Gao et al \cite{96gao2024efficient}, Efficient Adapting for Vision-language Foundation Model in Edge Computing Based on Personalized and Multi-Granularity FL}}
\underline{Overview and Contributions:} Gao et al \cite{96gao2024efficient} proposed PMG-FL, a novel personalized and multi-granularity FL framework designed for adapting vision-language foundation models (FM) in edge computing environments. PMG-FL addresses challenges such as data heterogeneity, limited resources, and multi-granularity data structures by leveraging personalized prompt training, distance-based aggregation, and cross-granularity guidance mechanisms. Experimental results demonstrate that PMG-FL significantly outperforms state-of-the-art methods in both IID and non-IID scenarios, achieving robust performance improvements across various tasks.

\begin{center}
\begin{tcolorbox}
\vspace{-0.05in}
\noindent \textbf{Contributions:}
Gao et al \cite{96gao2024efficient} proposed three major contributions: 
i) They introduced a personalized prompt training mechanism to adapt FM at the edge device level, effectively reducing computation and communication pressure, 
ii) They developed a distance-based aggregation mechanism to capture similarities among same-granularity edge devices, addressing non-IID challenges, and 
iii) They designed a bi-directional cross-granularity guidance mechanism to enhance knowledge fusion between fine- and coarse-granularity edge devices, improving overall model performance.
\end{tcolorbox}
\end{center}

\underline{Challenges:} The work identifies several challenges in adapting foundation models to edge environments, including handling non-IID data distributions caused by device diversity, managing limited computational and communication resources, and ensuring effective knowledge fusion across multi-granularity data. Furthermore, balancing the personalization of models for individual devices while achieving global generalization remains a complex task.

\underline{Future Directions:} Future research directions for PMG-FL include exploring more efficient and lightweight aggregation mechanisms to further reduce computational costs, developing adaptive strategies for dynamic granularity-aware model updates, and extending the framework to support diverse foundation models and modalities.

\paragraph*{\textbf{Wang et al \cite{97wang2024flora}, FLoRA: Federated Fine-Tuning LLMs with Heterogeneous Low-Rank Adaptations}}
\underline{Overview and Contributions:} Wang et al \cite{97wang2024flora} proposed FLoRA, a federated fine-tuning framework that addresses the limitations of existing methods by enabling noise-free aggregation of heterogeneous LoRA modules. The framework introduces a stacking-based aggregation mechanism, ensuring accurate global updates and seamless integration of heterogeneous LoRA ranks. Experimental results demonstrate the superiority of FLoRA over state-of-the-art methods in both homogeneous and heterogeneous federated fine-tuning scenarios, achieving improved performance and convergence efficiency.

\begin{center}
\begin{tcolorbox}
\vspace{-0.05in}
\noindent \textbf{Contributions:}
Wang et al \cite{97wang2024flora} proposed three major contributions:  
i) They developed a stacking-based aggregation mechanism for federated fine-tuning that eliminates noise introduced by naive averaging methods, ensuring faster convergence,  
ii) They introduced support for heterogeneous LoRA ranks across clients, enabling effective participation of devices with diverse computational capacities and data heterogeneity, and  
iii) They validated FLoRA through extensive experiments on multiple benchmarks, showcasing superior performance over FedIT and other baselines in both homogeneous and heterogeneous settings.
\end{tcolorbox}
\end{center}

\underline{Challenges:} The work identifies key challenges in federated fine-tuning of LLMs, including the aggregation noise introduced by traditional averaging methods, the inability to handle heterogeneous LoRA configurations across clients, and the computational overhead associated with communication and storage in large-scale federated setups. These issues hinder convergence and limit the applicability of existing federated fine-tuning approaches.

\underline{Future Directions:} Future research in federated fine-tuning of LLMs includes exploring lightweight aggregation techniques to further reduce communication overhead, enhancing privacy-preserving mechanisms to ensure robust security in federated settings, and extending FLoRA to support multi-modal and cross-domain federated fine-tuning tasks. Further, investigating dynamic rank adaptation strategies for LoRA to optimize resource utilization and performance across diverse client environments presents a promising direction.

\paragraph*{\textbf{Li et al \cite{98li2024mllm}, MLLM-FL: Multimodal LLM Assisted FL on Heterogeneous and Long-tailed Data}}
\underline{Overview and Contributions:} Li et al \cite{98li2024mllm} proposed MLLM-FL, an innovative framework that integrates multimodal LLMs (MLLMs) to address the challenges of data heterogeneity and long-tailed distributions in FL. MLLM-FL leverages server-side computational power and open-source data to perform global multimodal pretraining, federated fine-tuning, and global alignment. This approach enhances FL model performance while maintaining privacy and minimizing computational overhead on client devices.

\begin{center}
\begin{tcolorbox}
\vspace{-0.05in}
\noindent \textbf{Contributions:}
Li et al \cite{98li2024mllm} proposed three major contributions:  
i) They introduced the first integration of multimodal LLMs into FL to improve performance on heterogeneous and long-tailed data distributions,  {ii) introduces a methodology that improves privacy protection and reduces client-side computational overhead compared to state-of-the-art approaches for handling data heterogeneity in FL, and}  
iii) They demonstrated significant improvements over state-of-the-art FL approaches in both privacy preservation and computational efficiency through extensive experiments on benchmarks.
\end{tcolorbox}
\end{center}

\underline{Challenges:} The work identifies key challenges, including managing non-IID data distributions among clients, ensuring scalability and efficiency under resource-constrained environments, and mitigating the long-tailed distribution biases that degrade model performance. The reliance on high-quality server-side resources for global alignment further emphasizes the need for optimized data preprocessing and aggregation techniques.

\underline{Future Directions:} Future research directions include enhancing the dynamic weighted pretraining mechanism to reduce reliance on high-resource server setups, exploring more efficient global alignment strategies to tackle extreme class imbalances, and extending the framework to support multi-modal tasks beyond image-text pairings. Further, incorporating advanced privacy-preserving techniques to address potential vulnerabilities in server-side processing offers a promising avenue for further development.

\paragraph*{\textbf{Zhang et al \cite{99zhang2024distributed}, Distributed Foundation Models for Multi-Modal Learning in 6G Wireless Networks}}
\underline{Overview and Contributions:} Zhang et al \cite{99zhang2024distributed} proposed a novel framework that integrates distributed foundation models (FMs) for multi-modal learning within 6G wireless networks. The framework addresses key challenges in training and inference of large FMs by leveraging advanced communication and computation techniques, including pipeline and data parallelism. The approach is designed to enable sustainable AI development by distributing computational workloads across wireless devices, reducing energy consumption, and mitigating the bottlenecks associated with data heterogeneity and communication.

\begin{center}
\begin{tcolorbox}
\vspace{-0.05in}
\noindent \textbf{Contributions:}
Zhang et al \cite{99zhang2024distributed} proposed three major contributions:  
i) They introduced a distributed training architecture that incorporates pipeline and data parallelism to overcome computational and communication constraints in 6G wireless networks,  
ii) They developed methods for gradient compression and adaptive resource allocation to address bandwidth limitations and device heterogeneity, and  
iii) They demonstrated the integration of multi-modal learning techniques, such as image-text retrieval and audio captioning, to enhance the performance and applicability of foundation models in wireless environments.
\end{tcolorbox}
\end{center}

\underline{Challenges:} The work identifies critical challenges in distributed training of foundation models, including handling non-IID and heterogeneous data collected from wireless devices, managing the instability of wireless links that affect communication reliability, and addressing computational disparities across devices. The compression of activations and gradients also introduces potential biases that may impact model convergence, necessitating sophisticated optimization strategies.

\underline{Future Directions:} Future research could focus on advancing adaptive model partitioning and scheduling algorithms to improve load balancing across devices, exploring more robust communication compression techniques to mitigate aggregation errors, and extending multi-modal learning frameworks to support additional modalities and more complex tasks. Further, enhancing privacy-preserving mechanisms and integrating semantic communication paradigms hold promise for addressing data security concerns and optimizing training in distributed environments.

\paragraph*{\textbf{Nguyen et al \cite{100nguyen2024flora}, FLoRA: Enhancing VLMs with Parameter-Efficient FL}}
\underline{Overview and Contributions:} Nguyen et al \cite{100nguyen2024flora} proposed FLoRA, a parameter-efficient FL   framework designed to adapt VLMs, specifically CLIP, for distributed settings. This framework leverages LoRA  to fine-tune pre-trained VLMs with minimal communication overhead and enhanced adaptability to heterogeneous data distributions. The method ensures improved efficiency in communication and computational demands, achieving superior performance across various benchmarks. More details of this paper will be discussed in decentralized VLM section.

\section{Summary, Contributions, Challenges, and Future Directions of  Related  Studies on VLMs} \label{vlms}

In this section, we present the core novelties of the selected related works on VLMs and decentralized VLMs. We discuss the contributions and challenges of each study, and outline potential future directions based on their findings. In this section, although our main focus is on non-survey articles, we may cover some relevant survey papers as well.

\subsection{Introduction to VLMs}

\begin{figure}
  \centering
  \includegraphics[width= 0.48 \textwidth]{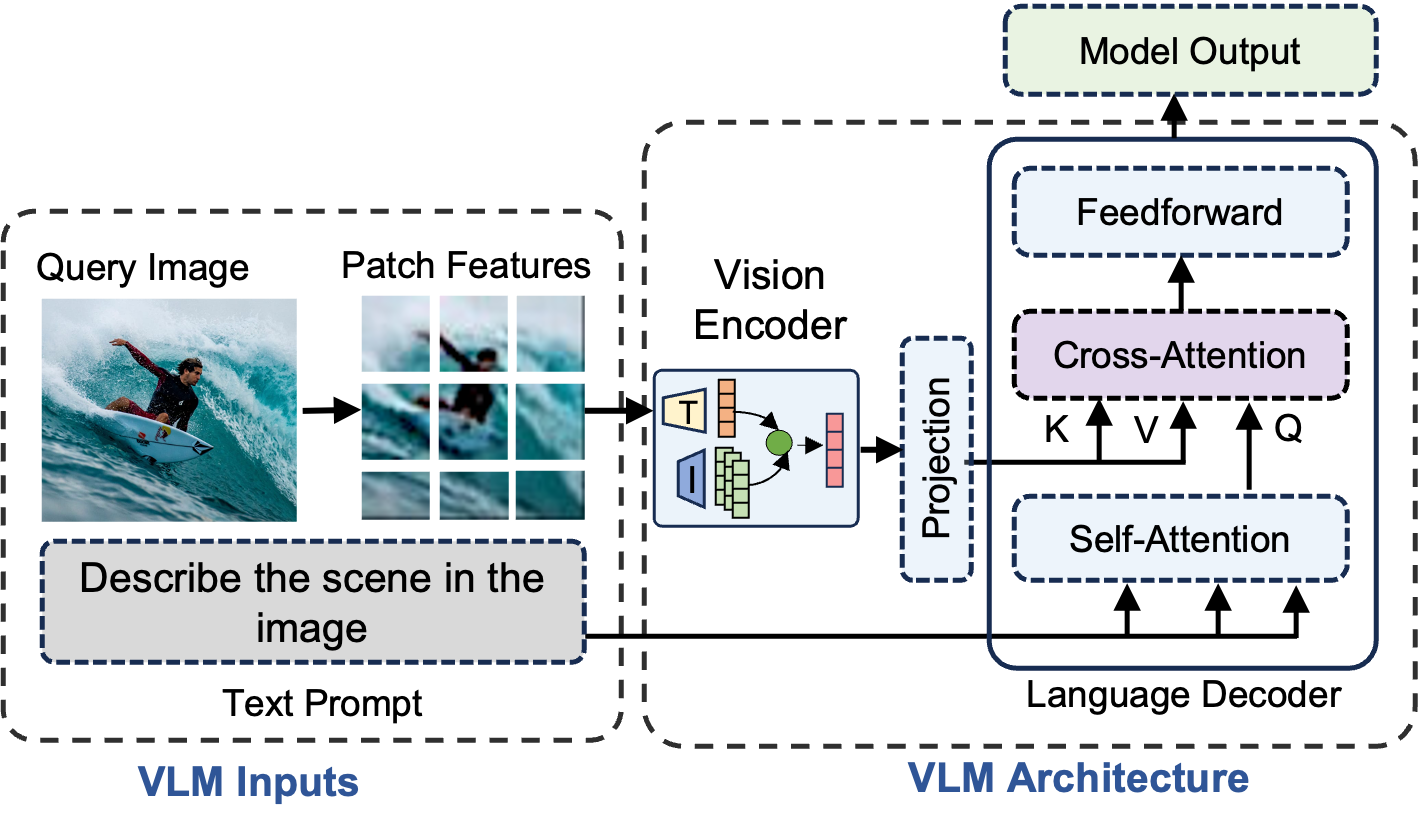} 
  \caption{Overview of the VLM Pipeline.}
  \label{vlmpip}  
\end{figure}

VLMs integrate visual and textual modalities to understand and generate multimodal content. These models leverage their ability to align visual perception with linguistic reasoning to tackle diverse tasks such as image captioning, where descriptive text is generated for images \cite{135vlmchen2022visualgpt}; visual question answering (VQA), where models provide answers to queries about visual content \cite{134vlmye2024mplug}; and cross-modal retrieval, which involves retrieving relevant images based on textual queries \cite{108vlmradford2021learning}. The workflow of VLMs relies on modality-specific encoders and sophisticated mechanisms to align and fuse visual and textual information. Vision encoders, such as convolutional neural networks (CNNs) or ViTs, extract spatial and semantic features from images, while language encoders such as GPT process textual inputs. To bridge these modalities, linear projection layers are commonly used to transform features from both encoders into a shared embedding space. This projection ensures that the visual and textual representations are compatible and facilitates effective cross-modal interaction.
Depending on the VLM architecture, multimodal fusion can occur at different stages. Early fusion combines raw features from both modalities at the initial stages, while intermediate fusion integrates processed representations through attention mechanisms, enabling deeper cross-modal interactions \cite{136vlmramesh2021zero, 137vlmli2019visualbert}. In contrast, late fusion merges modality-specific outputs at the decision level, preserving independent processing before final integration \cite{108vlmradford2021learning}.  Foundational pre-trained models such as CLIP  use contrastive learning to align image-text pairs in a shared space, while frameworks such as BLIP-2 \cite{114vlmli2022blip} employ modules such as the Querying Transformer (Q-Former) to bridge visual and textual information. An alternative approach is introduced by the Perceiver Resampler in Flamingo \cite{110vlmalayrac2022flamingo}, which reduces high-dimensional visual features into a fixed number of outputs using latent input queries and cross-attention. This design significantly reduces computational complexity while retaining the most essential visual information, making it particularly effective for downstream tasks. In Figure \ref{vlmpip}, we illustrate the pipeline of a VLM, where modality-specific encoders process multimodal inputs, followed by linear projection and cross-attention mechanisms to effectively align and integrate features from both modalities.

The effectiveness of VLMs in handling multimodal tasks relies not only on their architectural design but also on the strategies employed during their training and fine-tuning. In the subsequent sections, we discuss the fundamentals of VLM training and fine-tuning: 

\subsubsection{VLM Training}
 VLMs are often pretrained on large-scale multimodal datasets to learn robust and generalizable representations of visual and textual information. The process begins with an image Encoder, such as ViT \cite{113vlmdosovitskiy2020image} or foundational pre-trained model such as CLIP, which processes the input images and encodes them into visual embeddings. Simultaneously, a Text Decoder, often a pre-trained LLM like GPT or BERT, processes textual inputs. A crucial component of the training pipeline is the Multimodal Projector, which bridges the gap between the modalities by aligning and integrating features from the image encoder and text decoder into a shared space \cite{102vlmliu2024visual}. VLM pretraining typically involves tasks such as image-text matching and caption generation, where paired datasets with image and text pairs serve as ground truth. During this phase, the image encoder is often frozen to retain its pre-learned visual features, while the multimodal projector is fine-tuned to optimize the alignment of visual and textual embeddings.

\subsubsection{VLM Fine-tuning}
Fine-tuning VLMs involves selectively adapting specific components of the model to optimize performance on domain specific tasks. The decision of which layers to fine-tune depends on the task’s requirements and the differences between the pre-training and target domains. For tasks involving visual data from a new domain, such as medical images or satellite imagery \cite{132vlmhe2024parameter}, fine-tuning the vision encoder layers is essential to help the model adapt to the unique features of the visual input. Similarly, when the textual data includes domain-specific language, fine-tuning the LLM layers ensures accurate interpretation and generation of text. The projection or cross-attention layers, which align and integrate visual and textual features, are particularly crucial for tasks that demand strong multimodal reasoning, such as visual question answering or image captioning. Full model fine-tuning allows complete adaptation but is computationally intensive, while partial fine-tuning freezes certain layers, such as the lower layers of the encoders, to retain general features and reduce costs. Parameter-efficient approaches, like LoRA (Low-Rank Adaptation) \cite{hu2021lora}, introduce lightweight modifications to specific layers, enabling task-specific adaptation with minimal computational overhead. These strategies allow VLMs to effectively balance performance and efficiency, adapting to a wide range of downstream applications.

\subsection{Pretrained Foundation Models}

\paragraph*{\textbf{Radford et al. \cite{108vlmradford2021learning}, Learning transferable visual models from
natural language supervision}} \underline{Overview and Contributions:}
In this work, the authors introduced CLIP (Contrastive Language-Image Pretraining), a foundational pre-trained model that leverages contrastive learning and natural language supervision to learn highly transferable visual representations. CLIP aligns image and text embeddings by learning to predict correct (image, text) pairs from a massive dataset of 400 million examples. This design allows the model to integrate multimodal information effectively, making it a versatile tool for various tasks. CLIP has become a widely adopted vision encoder in large Vision-Language Models (LVLMs) due to its ability to generalize across diverse domains with minimal task-specific fine-tuning.

\begin{center}
\begin{tcolorbox}
\vspace{-0.05in}
\noindent \textbf{Contributions:}
The key contributions of this work include:
i) the introduction of a scalable contrastive pretraining method that aligns image and text representations by training on an extensive dataset of 400 million (image, text) pairs. This approach enables the learning of highly transferable multimodal embeddings, which serve as a robust foundation for numerous downstream applications; and
ii) the integration of textual understanding with visual representation learning, setting CLIP apart from traditional vision encoders such as ViT \cite{113vlmdosovitskiy2020image}, which focus solely on visual features. By aligning image and text embeddings within a shared semantic space through contrastive pretraining, CLIP achieves effective generalization across diverse tasks and domains without requiring extensive task-specific fine-tuning.
\end{tcolorbox}
\end{center}

\underline{Challenges:} Developing CLIP involved several challenges. First, scaling natural language supervision to a dataset of 400 million (image, text) pairs required significant computational resources and optimization of training methods. Second, ensuring robustness and generalization across diverse tasks and datasets demanded careful architectural and loss design. Finally, the inherent noisiness and variability of web-sourced data presented difficulties in maintaining the quality and relevance of training examples, requiring filtering and curation techniques.

\underline{Future Directions:} Future research could focus on extending CLIP to handle more complex multi-modal interactions, such as video understanding or multi-step reasoning tasks. Enhancing the scalability of the model to larger datasets and exploring its adaptability to fine-tuned applications (e.g., captioning and retrieval) could further broaden its applicability. Additionally, integrating advanced mechanisms to defend against adversarial attacks and examining its fairness across diverse data distributions will be crucial for real-world deployments.

\paragraph*{\textbf{Li et al. \cite{114vlmli2022blip}, Blip: Bootstrapping language-image pre-training for unified vision-language understanding and generation}} \underline{Overview and Contributions:}

Li et al. introduced BLIP \cite{114vlmli2022blip}, a unified framework designed to address both vision-language understanding and generation tasks. BLIP employs a dual-stage design with an encoder for aligning vision and language representations and a decoder for generative tasks such as captioning. Additionally, BLIP incorporates a captioner to generate high-quality captions and a filter to remove noisy image-text pairs, enhancing the quality of the training data. The authors demonstrate that bootstrapping captions and leveraging diverse captions enable BLIP to achieve substantial improvements across various downstream tasks.

\begin{center}
\begin{tcolorbox}
\vspace{-0.05in}
\noindent \textbf{Contributions:}
The key contributions of this work include:
i) the development of a dual vision-language encoder and a decoder framework that efficiently handles both understanding and generation tasks, addressing limitations of prior models which focused exclusively on one domain; and
ii) the introduction of \textbf{CapFilt}, a novel mechanism combining a captioner to generate high-quality pseudo-captions and a filter to remove noisy or irrelevant image-text pairs, significantly enhancing the quality of the pretraining dataset;
\end{tcolorbox}
\end{center}

\underline{Challenges:} 
BLIP addresses several challenges in vision-language pretraining. First, unifying understanding and generation tasks within a single framework required carefully balancing task-specific needs, such as optimizing encoders for retrieval and decoders for caption generation. Second, ensuring the quality of the pretraining dataset posed significant difficulties due to the prevalence of noisy and inconsistent captions, many of which were derived from alt text and lacked sufficient detail for effective training. BLIP tackled this issue by introducing CapFilt, a mechanism that removes noisy image-text pairs and generates high-quality pseudo-captions, significantly enhancing the dataset’s relevance and informativeness.

\underline{Future Directions:}
Future work could focus on extending BLIP’s capabilities to more complex scenarios, such as video understanding, temporal reasoning, or multi-modal dialogue systems. Exploring adaptive data curation techniques, such as dynamic filtering or domain-specific bootstrapping, may improve the robustness and reliability of pretraining. Additionally, integrating BLIP into specialized domains, such as medical imaging or autonomous navigation, could expand its applicability. Addressing issues of fairness, bias mitigation, and scalability in handling larger datasets will also be critical for improving its deployment in real-world settings.

\subsection{Centralized VLM}

\paragraph*{\textbf{Alayrac et al. \cite{110vlmalayrac2022flamingo}, Flamingo: a Visual Language Model
for Few-Shot Learning}} \underline{Overview and Contributions:}
Alayrac et al. \cite{110vlmalayrac2022flamingo} proposed Flamingo, a framework designed to advance few-shot learning in multimodal tasks by integrating pretrained vision and language models through Perceiver Resampler and Gated Cross-Attention Dense layers. This architecture allows seamless processing of interleaved text and visual inputs, enabling state-of-the-art performance across diverse benchmarks with minimal task-specific data. 

\begin{center}
\begin{tcolorbox}
\vspace{-0.05in}
\noindent \textbf{Contributions:}
Alayrac et al. \cite{110vlmalayrac2022flamingo} provides three significant contributions: i) a novel architecture that bridges pretrained vision and language models, enabling them to handle arbitrarily interleaved sequences of text and images for open-ended text generation; ii) the development of a Perceiver Resampler and Gated Cross-Attention Dense layers, ensuring efficient and scalable visual token processing while preserving pretrained knowledge; and iii) state-of-the-art performance in few-shot learning across 16 multimodal benchmarks, surpassing fine-tuned methods on several downstream tasks.
\end{tcolorbox}
\end{center}
\underline{Challenges:} 
In this work, the authors addressed key challenges in multimodal learning, including the alignment of pretrained vision and language models while preserving their knowledge during integration. Managing computational overhead in cross-attention mechanisms and ensuring scalability for high-resolution inputs posed significant constraints. Furthermore, the model's performance in classification tasks was limited compared to contrastive methods, which are specifically optimized for retrieval-based objectives.

\underline{Future Directions:} Future research could focus on enhancing the efficiency and scalability of Flamingo's architecture, such as optimizing the Perceiver Resampler and cross-attention mechanisms. Developing unified approaches that combine the strengths of generative and contrastive methods might improve performance across a broader range of tasks. Furthermore, exploring ways to mitigate the limitations of in-context learning, such as adaptive prompt optimization or hybrid learning paradigms, could expand its applicability.

\paragraph*{\textbf{Liu et al \cite{102vlmliu2024visual}, Visual Instruction Tuning}} 
 \underline{Overview and Contributions:} LLaVA, an open-source multimodal framework
designed to enhance LLMs for understanding both language
and images. It utilizes decoder only transformers (e.g., vicuna \cite{101vlmchiang2023vicuna}) or LLMs (e.g., Llama \cite{}) to generate text responses
for instruction-following tasks in a multimodal context. LLaVA
integrates a vision encoder from CLIP with the LLM, enabling
it to process visual information alongside language. The model
undergoes pre-training on image-text pairs and fine-tuning for
end-to-end multimodal understanding, resulting in a versatile
multimodal model. 
\begin{center}
\begin{tcolorbox}
\vspace{-0.05in}
\noindent \textbf{Contributions:}
 Liu et al. \cite{102vlmliu2024visual} introduced three core contributions: i) a GPT-assisted multimodal data generation pipeline designed to convert image-text pairs into high-quality instruction-following datasets, facilitating the alignment of vision and language representations; ii) an end-to-end multimodal architecture integrating CLIP's vision encoder with language decoders, optimized through a two-stage training process for robust instruction-following; and iii) the development of LLaVA-Bench, a benchmark with diverse tasks, alongside open-source models, datasets, and tools, enabling substantial advancements in multimodal research and real-world visual reasoning.
\end{tcolorbox}
\end{center}

\underline{Challenges:} The primary challenges in  LLaVA indclude the scarcity of high-quality multimodal instruction-following datasets and the complexities of aligning visual and language representations. Balancing the training of large multimodal models while retaining generalization capabilities proved demanding. Moreover, limitations in current visual tokenization techniques led to suboptimal representation of complex image semantics.

\underline{Future Directions:} Future research could focus on improving multimodal alignment with advanced tokenization and fusion strategies. Exploring techniques for scalable dataset generation, possibly leveraging synthetic data or enhanced machine-annotation pipelines, can further enhance performance.

\paragraph*{\textbf{Zhu et al. \cite{104vlmzhuminigpt}, Minigpt-4: Enhancing vision-language understanding with advanced large language models}} 
 \underline{Overview and Contributions:} MiniGPT-4 is a VLM designed to align a pretrained vision encoder with an advanced LLM, Vicuna, using a single projection layer. It achieves advanced multimodal capabilities similar to GPT-4, including detailed image descriptions, website generation from handwritten drafts, and creative tasks like poem writing and meme interpretation. The model demonstrates efficient training with minimal computational resources while addressing limitations in vision-language alignment through a two-stage training process.

\begin{center}
\begin{tcolorbox}
\vspace{-0.05in}
\noindent \textbf{Contributions:}
MiniGPT-4 \cite{104vlmzhuminigpt} provides three significant contributions: i) a novel architecture that aligns visual features from a pretrained vision encoder with the Vicuna large language model using a single projection layer, enabling efficient and scalable multimodal learning; ii) a two-stage training process that addresses early-stage language generation issues by fine-tuning on a curated dataset of detailed image descriptions, significantly enhancing language output quality and usability; and iii) the demonstration of advanced multimodal capabilities, including detailed image descriptions, contextual humor interpretation, and creative writing, outperforming prior models on a variety of vision-language tasks with minimal computational resources.
\end{tcolorbox}
\end{center}

\underline{Challenges:}The primary challenges identified in this work include preserving pretrained knowledge while aligning vision and language components through a single projection layer. Early-stage training using short image captions led to fragmented and incoherent language outputs. Furthermore, the model showed limitations in understanding spatial relationships and exhibited hallucinations when generating longer image descriptions. These issues were addressed through the curation of a high-quality dataset and targeted fine-tuning.

\underline{Future Directions:} Future research can explore integrating reinforcement learning with AI feedback to mitigate hallucinations in image descriptions. Expanding datasets tailored to spatial reasoning and contextual understanding, such as those focused on visual scene layouts, could improve spatial comprehension. Moreover, investigating the use of multimodal datasets generated by advanced LLMs like GPT-4V and developing methods for scalable vision-language alignment hold promise for enhancing generalization and performance across diverse tasks.

\subsection{Summary, Contributions, Challenges, and Future Directions of  Related  Studies on Decetnralized VLMs }

\paragraph*{\textbf{Nguyen et al.\cite{100nguyen2024flora}, FLoRA: Enhancing Vision-Language Models with
Parameter-Efficient Federated Learning}} 
 \underline{Overview and Contributions:} FLoRA is a framework designed to enhance VLMs within FL settings by integrating Low-Rank Adaptation (LoRA) for parameter-efficient fine-tuning. By focusing on lightweight model updates, FLoRA addresses privacy concerns, reduces communication costs, and accelerates training while preserving the performance of pretrained models. The methodology demonstrates significant gains in efficiency and accuracy across diverse datasets, making it a robust solution for federated VLM training.

\begin{center}
\begin{tcolorbox}
\vspace{-0.05in}
\noindent \textbf{Contributions:}
Nguyen et al. \cite{100nguyen2024flora} provides three significant contributions:
i) the adaptation of LoRA to the CLIP model's text encoder, enabling efficient fine-tuning with minimal parameter updates while maintaining high performance;
ii) a FL strategy that achieves reduction in communication overhead compared to full model aggregation approaches;
iii) rigorous evaluations across various datasets with increased accuracy over traditional federated baselines.
\end{tcolorbox}
\end{center}

\underline{Challenges:} The development of FLoRA presented several challenges, including the integration of LoRA adapters into the CLIP architecture while preserving model performance. Addressing the complexities of non-IID data distributions across federated clients necessitated sophisticated optimization of training and aggregation strategies. Additionally, balancing data privacy with computational efficiency posed significant difficulties, particularly in minimizing communication overhead in distributed learning environments.

\underline{Future Directions:} Future research could explore extending LoRA adaptations to additional components of the CLIP model, such as image encoder layers and transformer blocks, to further improve flexibility and generalization. Investigating adaptive federated optimization techniques to dynamically address client-specific data heterogeneity and integrating advanced privacy-preserving mechanisms like differential privacy could enhance FLoRA's robustness. Additionally, expanding evaluations to include real-world applications and low-resource devices would provide valuable insights into the framework’s scalability and practical deployment potential.

\paragraph*{\textbf{Shi et al.\cite{103vlmshi2023clip}, CLIP-Guided Federated Learning on Heterogeneous and Long-Tailed Data}} 
 \underline{Overview and Contributions:} The paper presents CLIP2FL, a FL framework designed to tackle challenges arising from data heterogeneity and long-tailed distributions. This approach utilizes the vision-language capabilities of the CLIP model to enhance learning at both the client and server levels. On the client side, knowledge distillation transfers CLIP’s extensive vision-language prior knowledge to local models, enhancing feature representation, and reducing biases toward dominant classes. On the server side, prototype contrastive learning is employed to create federated features guided by CLIP’s text encoder, enabling the retraining of a balanced server classifier.

\begin{center}
\begin{tcolorbox}
\vspace{-0.05in}
\noindent \textbf{Contributions:}
The key contributions of this work include the novel integration of CLIP into FL to connect client-side and server-side training, leveraging CLIP’s rich semantic knowledge to generate meaningful federated features, and achieving state-of-the-art performance on benchmark datasets such as CIFAR-10/100-LT and ImageNet-LT.
\end{tcolorbox}
\end{center}

\underline{Challenges:} CLIP2FL faces several key challenges that underline the complexity of its implementation. The heterogeneity of client data remains a significant issue, as variations in data distributions across clients can lead to inconsistencies in the global model's performance. Similarly, addressing long-tailed data distributions continues to be difficult, especially when achieving balanced performance across head and tail classes. Scalability to larger datasets and more complex FL scenarios presents additional hurdles, including computational and communication bottlenecks. Furthermore, the framework's dependence on pre-trained models like CLIP assumes access to pretrained weights, which may not always be feasible in certain FL deployments. Finally, while the evaluation demonstrates impressive results, a heavier reliance on classification accuracy as a metric limits the scope of analysis, and broader metrics such as efficiency and fairness remain underexplored.

\underline{Future Directions:} The framework can be expanded to support diverse modalities, such as audio, text, and video, enabling its application to multimodal FL scenarios. Real-world applications in domains like healthcare, autonomous vehicles, and smart cities could test its robustness and practical utility. Adaptive mechanisms to handle dynamic client participation and varying computational resources during training would enhance its scalability. Furthermore, optimizing the computational and communication costs associated with CLIP-based FL would improve its efficiency. Broader evaluation metrics, including fairness, robustness against adversarial attacks, and energy efficiency, should be explored to provide a more comprehensive assessment of the framework's capabilities. Experimenting with other vision-language models or foundation models would help generalize the approach, while integrating advanced privacy-preserving techniques like differential privacy or homomorphic encryption could further strengthen data confidentiality.

\paragraph*{\textbf{Imteaj et al.\cite{105vlmimteaj2024tripleplay}, TriplePlay: Enhancing Federated Learning with CLIP for Non-IID Data and Resource Efficiency}} 
 \underline{Overview and Contributions:} The paper explores the integration of the CLIP foundation model into FL systems to address challenges such as non-IID data distributions, skewed class representation, and resource constraints in distributed learning environments. By leveraging CLIP's robust feature extraction capabilities and applying techniques like quantization and low-rank adaptation, the paper proposes a novel framework, TriplePlay, to improve FL performance while optimizing resource usage.

\begin{center}
\begin{tcolorbox}
\vspace{-0.05in}
\noindent \textbf{Contributions:}
The main contributions of the paper include the development of an adaptive FL framework that incorporates CLIP as an attention-based adapter for personalized and generalized learning. It introduces GAN-based synthetic data generation to tackle the issue of long-tail data distributions, enabling better representation of underrepresented classes. Furthermore, the framework employs quantization and QLoRA techniques to reduce computational demands, ensuring efficient local model training and communication.
\end{tcolorbox}
\end{center}

\underline{Challenges:} The challenges addressed are significant, including the difficulty of achieving equitable learning across heterogeneous client datasets, the high computational cost of training large models like CLIP in resource-constrained environments, and the communication overhead associated with exchanging large model updates in FL settings. The imbalance in class distributions across clients further complicates learning, requiring innovative solutions to ensure fairness.
The experimental results demonstrate that TriplePlay achieves faster convergence, reduced GPU usage, and improved accuracy compared to baseline methods such as FedCLIP. By generalizing effectively across diverse datasets (PACS and Office-Home), the framework validates its scalability and adaptability.

\underline{Future Directions:} Future directions for this research include extending the proposed approach to other foundation models and multimodal learning scenarios, further improving resource optimization for more constrained edge devices, and exploring the integration of advanced privacy-preserving techniques to enhance security in FL systems. These advancements could pave the way for broader applications in domains like healthcare, smart cities, and finance, where data privacy and resource efficiency are critical.

\paragraph*{\textbf{Lu et al. \cite{106vlmlu2023fedclip}, FedCLIP: Fast generalization and personalization for clip in federated learning}} 
 \underline{Overview and Contributions:} The paper presents FedCLIP, a FL framework designed to achieve fast generalization and personalization for the CLIP model in distributed environments. It employs an attention-based adapter that focuses on specific tasks without fine-tuning the entire model, thereby significantly reducing computational and communication costs. By leveraging pretrained CLIP \cite{108vlmradford2021learning} features and lightweight updates, FedCLIP addresses the challenges of resource constraints and data heterogeneity, demonstrating superior performance in both generalization and personalization across multiple datasets.

\begin{center}
\begin{tcolorbox}
\vspace{-0.05in}
\noindent \textbf{Contributions:}
The key contributions of this work include: i) the development of an attention-based adapter for the CLIP image encoder, enabling efficient task-specific fine-tuning while preserving pretrained knowledge; and ii) a lightweight FL framework that reduces computational costs and minimizes communication overhead by focusing on adapter parameter updates rather than full model aggregation.
\end{tcolorbox}
\end{center}

\underline{Challenges:} FedCLIP faces several challenges, including the integration of lightweight adapters into the CLIP architecture, which requires careful design to preserve its pretrained capabilities while enabling effective task-specific adaptation. The heterogeneity of data distributions across clients introduces additional complexity, necessitating strategies to ensure consistent optimization and convergence. Furthermore, balancing the reduction of computational and communication costs with maintaining high model performance in federated settings poses significant technical hurdles that demand innovative solutions.

\underline{Future Directions:} Future research can explore expanding FedCLIP’s framework to include adapters for text encoders in addition to image encoders, further enhancing its adaptability. Furthermore, this adapter-based architecture could be integrated with advanced foundation models to further improve performance and scalability.

\paragraph*{\textbf{Guo et al. \cite{109vlmguo2023pfedprompt}, 
pFedPrompt: Learning Personalized Prompt for Vision-Language Models in Federated Learning}} 
 \underline{Overview and Contributions:} In this work the authors introduced pFedPrompt, a FL framework designed to address user heterogeneity in VLMs by introducing personalized prompt learning. It combines global user consensus in the linguistic space with local feature adaptation in the visual space. By leveraging the multimodal capabilities of pre-trained models like CLIP, pFedPrompt achieves effective personalization and robust performance across diverse datasets, even under non-IID data conditions.

\begin{center}
\begin{tcolorbox}
\vspace{-0.05in}
\noindent \textbf{Contributions:}
The key contributions of this work include:
i) identifying the limitations of existing prompt training in FL, particularly its inability to personalize for heterogeneous users;
ii) introducing pFedPrompt, which combines global user consensus from the linguistic space with local feature attention from the visual space to address user heterogeneity effectively;
iii) demonstrating that this dual-modality approach improves performance and personalization by dynamically adapting to user-specific data distributions; and
iv) conducting extensive evaluations across multiple datasets, showing that pFedPrompt outperforms state-of-the-art personalized FL methods in accuracy and robustness under non-IID scenarios.
\end{tcolorbox}
\end{center}

\underline{Challenges:} pFedPrompt faces several key challenges that underscore the complexities of personalized prompt learning in federated settings. Addressing data heterogeneity in non-IID scenarios is critical to ensuring consistent model performance across clients. Additionally, balancing global user consensus with local adaptation requires effective integration of linguistic and visual modalities without compromising pretrained capabilities. Furthermore, scaling to larger federated networks introduces significant computational and communication bottlenecks. The framework's reliance on pretrained models like CLIP assumes accessibility, which may not always be feasible in resource-constrained or privacy-sensitive environments.

\underline{Future Directions:} Future research could explore extending pFedPrompt to other vision-language models and incorporating text encoder adaptation for further flexibility. Enhancing scalability to handle larger federated networks and more complex data distributions is another promising direction. Additionally, integrating advanced privacy-preserving techniques, such as differential privacy, could strengthen the framework's applicability in sensitive domains like healthcare and finance.

\paragraph*{\textbf{Chen et al. \cite{111vlmchen2024feddat}, 
FedDAT: An Approach for Foundation Model Finetuning in Multi-Modal Heterogeneous Federated Learning}} 

\underline{Overview and Contributions:} The paper explores the complexities of PEFT for foundation models in heterogeneous multi-modal FL settings, particularly addressing challenges such as data heterogeneity across clients, communication efficiency, and computational constraints. To overcome these issues, it introduces FedDAT, a FL framework designed for Vision-Language (VL) tasks. FedDAT employs a Dual-Adapter Teacher (DAT) module to balance client-specific and client-agnostic knowledge and leverages Mutual Knowledge Distillation (MKD) for effective knowledge transfer, ensuring robust learning and scalability. 

\begin{center}
\begin{tcolorbox}
\vspace{-0.05in}
\noindent \textbf{Contributions:}
The key contributions of this work include:
i) the introduction of the Federated Dual-Adapter Teacher (FedDAT) framework, the first to address PEFT of foundation models for heterogeneous multi-modal FL tasks;
ii) the design of a Dual-Adapter Teacher (DAT) module that combines client-specific and client-agnostic knowledge through parallel adapters, enabling effective personalization and generalization;
iii) the application of Mutual Knowledge Distillation (MKD) to ensure efficient knowledge transfer and robust learning in data-heterogeneous environments; and
iv) extensive evaluations on four diverse multi-modal benchmarks, demonstrating superior performance, scalability, and convergence compared to existing PEFT methods.
\end{tcolorbox}
\end{center}

\underline{Challenges:} FedDAT faces several challenges that highlight the complexities of fine-tuning foundation models in heterogeneous multi-modal federated settings. First, managing the heterogeneity of multi-modal data across clients, including variations in both vision and language modalities, presented significant optimization challenges. Second, ensuring a balance between client-specific and client-agnostic knowledge required a carefully designed architecture to mitigate model drift. Finally, maintaining computational efficiency and minimizing communication overhead while scaling to large federated networks posed substantial technical hurdles that necessitated innovative solutions.

\underline{Future Directions:} Future research could focus on expanding FedDAT’s framework to incorporate task-specific adaptations to enhance its flexibility. Investigating more advanced strategies for balancing client-specific and global knowledge, such as adaptive adapter mechanisms, may improve scalability and robustness. Additionally, exploring FedDAT’s performance on downstream tasks beyond Visual Question Answering (VQA), such as image captioning and other Vision-Language tasks, could provide deeper insights into its versatility and generalization capabilities.

\paragraph*{\textbf{Wang et al \cite{25wang2024federatedLLM}, Federated Instruction Tuning of LLMs with Domain Coverage Augmentation}}
\underline{Overview and Contributions:} Wang et al \cite{25wang2024federatedLLM} proposed  FedDCA to enhance the performance of federated instruction tuning in LLMs. This approach focuses on augmenting domain-specific instructions by leveraging a combination of client-private and server-public datasets. The method introduces techniques for maximizing domain coverage while maintaining privacy and computational efficiency, particularly via a variant, FedDCA*, which utilizes heterogeneous encoders. Extensive experiments demonstrate significant improvements across various domains.

\underline{Challenges:} The work faces challenges in maintaining a balance between computational efficiency and model performance, particularly when scaling FedDCA* for broader applications. Ensuring privacy while optimizing instruction augmentation presents technical difficulties, especially against memory extraction attacks. Further, aligning heterogeneous encoder outputs without compromising semantic accuracy remains a complex task.

\underline{Future Directions:} Future research can explore enhancing the robustness of FedDCA against advanced privacy threats, such as adversarial attacks; developing dynamic domain coverage metrics to adapt to evolving client data distributions could improve model generalization; and investigating alternative methods for efficient instruction augmentation, such as leveraging synthetic data or advanced generative techniques to further optimize performance.

\section{ A Brief Overview of Small Language Models (SLMs)}
\label{slm}

In distributed settings, there are scenarios that the local device has resource limitations. This makes running LLMs locally challenging due to limited hardware resources. However, there are several lightweight language models or SLMs that can be run efficiently even on resource-constrained edge devices. Here, we provide a list of SLMs that are designed to operate on less powerful hardware \cite{lu2024small}. In this section, we provide a curated list of such lightweight language models, along with their respective links to repositories or Hugging Face model hubs for quick access. Each entry is accompanied by PyTorch code snippets, presented in gray boxes, to guide users in deploying these models efficiently. These resources aim to empower developers to integrate LLM capabilities into low-resource devices, enabling broader adoption in real-world, distributed applications. Table \ref{lightweight1} summarizes some of SLMs. More details are provided in \textit{Appendix A}.

\onecolumn
\small
\captionof{table}{Overview of Selected Small Language Models}
\label{lightweight1}
\begin{longtable}{|p{3cm}|p{2cm}|p{5cm}|p{5cm}|}
\hline
\textbf{Name} & \textbf{Parameters} & \textbf{Highlights} & \textbf{Challenges} \\ \hline
GPT-2 Small \cite{radford2019language} & 124M & Compact GPT-2 variant; good for general NLP tasks and text generation. & Lacks instruction-following capabilities and limited multilingual support. \\ \hline
DistilGPT-2 \cite{sanh2019distilbert} & 82M & Distilled version of GPT-2; faster and smaller with 97\% performance retention. & Limited for nuanced understanding and lacks bidirectional capabilities. \\ \hline
DistilBERT \cite{sanh2019distilbert} & 66M & Compact BERT; versatile for classification and question answering. & Requires fine-tuning for domain-specific tasks; limited multilingual support. \\ \hline
TinyBERT \cite{jiao2019tinybert} & 14.5M & the Most lightweight LLM; efficient for on-device NLP. & Performance drops on complex tasks; limited generalization. \\ \hline
MobileBERT \cite{sun2020mobilebert} & 25.3M & Optimized for mobile devices; fast inference and strong general NLP performance. & Fine-tuning complexity; lacks extensive multilingual capabilities. \\ \hline
T5 Small \cite{raffel2020exploring} & 60M & General-purpose text-to-text transformer; versatile and efficient. & Limited multilingual support; weaker on extremely complex tasks. \\ \hline
Flan-T5 Small \cite{FlanT5-small} & 80M & Instruction-tuned version of T5; excels in task generalization. & Requires more resources than T5 Small. \\ \hline
ByT5 Small \cite{xue2022byt5} & 300M & Token-free architecture; handles noisy and multilingual data well. & Larger size compared to other small LLMs; slower inference. \\ \hline
OPT-350M \cite{zhang2022opt} & 350M & Open-source GPT-3-like model; strong in text generation tasks. & Larger size for constrained devices; lacks multilingual capabilities. \\ \hline
MiniCPM 230M \cite{yao2024minicpm}& 230M & Balanced performance; optimized for general NLP tasks. & Requires fine-tuning for domain-specific tasks. \\ \hline
MiniCPM 1.2B \cite{yao2024minicpm} & 1.2B & Highly efficient for larger tasks; competitive with larger LLMs. & Resource-intensive for very constrained devices. \\ \hline
Reformer \cite{kitaev2020reformer}& 110M & Memory-efficient; handles long sequences well. & Approximation errors in LSH; requires specific use-case tuning. \\ \hline
Gemma 2B \cite{team2024gemma} & 2B & Extended context (8k tokens); instruction-tuned variant available. & Larger size than many small LLMs; fine-tuning complexity. \\ \hline
Cerebras-GPT 256M \cite{dey2023cerebras} & 256M & Compute-optimal training; efficient and scalable. & Lacks instruction-tuning and advanced multilingual support. \\ \hline
TinyLlama 1.1B \cite{zhang2024tinyllama} & 1.1B & Compact GPT variant; excellent for generative tasks. & Relatively larger size compared to other 'small' LLMs. \\ \hline
GPT-Neo \cite{black2022gpt} & 125M & Open-source GPT variant; good for text generation and reasoning tasks. & Requires fine-tuning for specific tasks; limited multilingual capabilities. \\ \hline
LLaMA-2 7B \cite{touvron2023llama} & 7B & Powerful small LLM; instruction-tuned and highly generalizable. & Not open-source for commercial use; resource-intensive. \\ \hline
ALBERT-Base \cite{lan2019albert} & 11M & Parameter-efficient BERT variant with cross-layer parameter sharing. & Slower inference compared to TinyBERT; requires task-specific fine-tuning. \\ \hline

Phi-2 \cite{javaheripi2023phi} & 2.7B & Optimized for technical tasks like reasoning and coding; highly efficient. & Not fully open-source; limited general-purpose task training. \\ \hline
Gemini Nano \cite{team2023gemini}& 300M & Efficient multimodal LLM optimized for edge devices. & Multimodal tasks require extra fine-tuning; relatively new, limited documentation. \\ \hline
Claude Instant & 500M  & Fast and lightweight variant of Claude for interactive tasks. & Not open-source; limited accessibility for research. \\ \hline
BLOOMZ 560M \cite{muennighoff2022crosslingual} & 560M & Multilingual and instruction-tuned; excels in cross-lingual tasks. & Larger size compared to other small LLMs; slower inference. \\ \hline
Galactica 125M \cite{GALACTICA} & 125M & Specialized in scientific text and reasoning tasks. & Domain-specific; limited generalization to non-scientific domains. \\ \hline
Pythia \cite{biderman2023pythia} 160M & 160M & Transparent training process; versatile across general NLP tasks. & Lacks instruction-tuning and multilingual capabilities. \\ \hline
\end{longtable}
\twocolumn

\section{Highlights of Future Research Directions}
\label{frd}
\begin{figure*}
  \centering
\includegraphics[height=0.99\textheight]{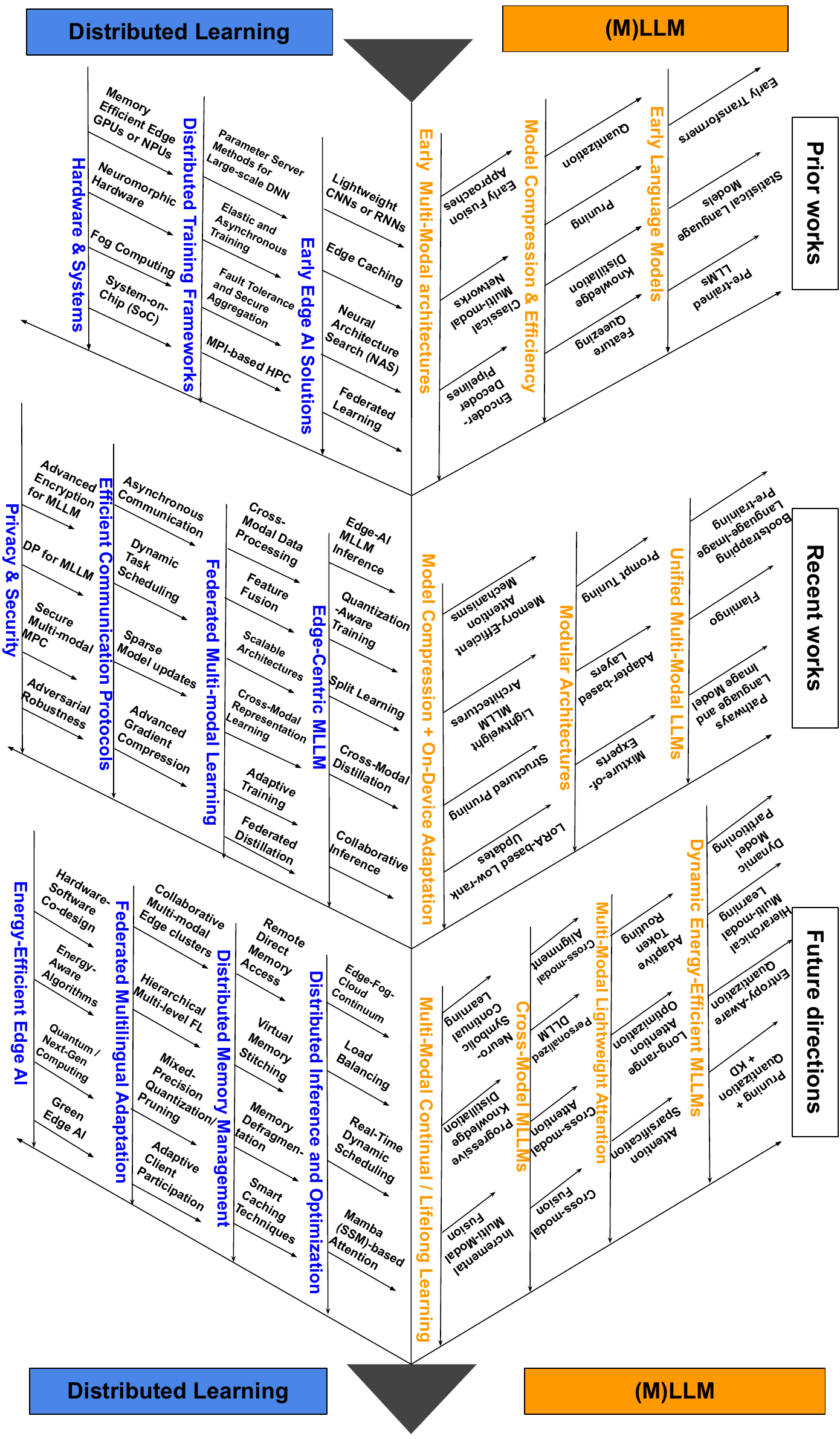}
  \caption{Highlights of selected prior, current, and future research directions.}
  \label{figfuture}  
\end{figure*}

As the field of distributed MLLMs is a very wide area of research, we focus on future directions in specific directions. Figure \ref{figfuture} shows an overview of selected prior, current, and  future research directions.

\noindent $\diamondsuit$ \textbf{Distributed Inference and Optimization}
\begin{itemize}
\item \textbf{Memory-Efficient Inference Mechanisms} 
Since self-attention requires quadratic time complexity for inference, exploring memory-efficient alternatives is a promising future direction. Mamba \cite{gu2023mamba} and other State Space Models (SSMs), such as LSSL and S4 \cite{gu2021efficiently}, offer linear-time inference, optimizing memory usage while improving the inference speed of longer sequences without compromising performance.

\item \textbf{Automated Real-Time Parallelism} 
Evolving hybrid parallelization methods can integrate context-aware scheduling and sparse attention mechanisms \cite{3li2024distflashattnLLM,5brakel2024modelLLM} to ensure balanced workloads across heterogeneous hardware capabilities \cite{6he2024distributedLLM}. Given the continuous and dynamic nature of workloads in distributed systems, real-time parallelism (or Automated Parallelism) strategies are essential to efficiently manage task distribution and system behavior from a top-down perspective \cite{2wu2023fastLLM, 3li2024distflashattnLLM, 5brakel2024modelLLM}. 
 
\item \textbf{Real-Time Dynamic Scheduling} Hybrid edge-cloud platforms require inference systems that dynamically reallocate tasks based on time constraints and the resource availability of each edge device. Due to the complexities of resource-constrained distributed settings, novel adaptive scheduling mechanisms are required to manage dynamic shifts in data, network stability, and available computational resources \cite{10khoshsirat2024decentralizedLLM,11ren2024taskLLM,2wu2023fastLLM}. 
Integrating energy prediction models that adaptively update scheduling strategies based on energy constraints \cite{10khoshsirat2024decentralizedLLM} or employing online scheduling mechanisms that adjust to network fluctuations \cite{11ren2024taskLLM} could significantly reduce latency and optimize resource utilization, which is essential for resource-limited edge devices. Exploiting priority-based scheduling mechanisms, such as those proposed in \cite{fahad2022multi}, could also ensure the timely processing of high-priority tasks for DLLM. On the other hand, Joint training and inference paradigms \cite{69ouyang2024pluto,73wang2023privatelora} leveraging both cloud and edge resources can balance local autonomy with global optimization and could be further explored as another future research direction. 

\end{itemize}

\noindent $\diamondsuit$ \textbf{Memory Management Optimization}
\begin{itemize}

\item \textbf{Efficient Memory Allocation for Load Balancing}  
Optimizing memory allocation is crucial for efficiently deploying Distributed LLMs in hybrid edge-cloud environments. Future research could explore \textit{adaptive memory allocation techniques}, where each node dynamically adjusts its memory usage based on its current workload and hardware constraints \cite{7borzunov2024distributedLLM, 12yao2024scalellmLLM}. Techniques such as \textit{memory defragmentation} and \textit{virtual memory stitching} \cite{guo2024gmlake} could enhance memory efficiency in multi-node LLM execution.
Further, \textit{model offloading} to trusted, available participating devices presents another promising direction. For instance, active inference-based offloading could be applied in cloud-based systems to minimize latency and optimize resource utilization \cite{fang2023large}. This approach could allow for \textit{dynamic task redistribution}, reducing the memory burden on edge devices while maintaining efficient inference speed.

\item \textbf{Remote Direct Memory Access (RDMA)}  
Using secure RDMA approaches, one computing node can directly access the memory of another node without involving the CPU, operating system, or intermediate buffering. This significantly reduces latency, computational overhead, and memory bandwidth usage, making it highly suitable for distributed computing \cite{19zhang2024fedrdmaLLM}.  Future research could explore optimized RDMA-based memory sharing strategies for heterogeneous distributed infrastructures, particularly in low-power edge computing scenarios where traditional memory management techniques introduce significant latency. The next future path could be combining RDMA with PCIe (Peripheral Component Interconnect Express) \cite{gangidi2024rdma} to further expedite the data movement process and apply light-weight encryption approaches which ensure a safe transmission process.

\item \textbf{Smart Caching Techniques:}  
Optimizing caching mechanisms is crucial for reducing memory and computational overhead, and improving inference speed, which is all essential in LLM-based Distributed settings. Future research could explore \textit{advanced KV cache compression and allocation} strategies to optimize memory usage without performance degradation. For instance, the integration of Dynamic Memory Compression \cite{nawrot2024dynamic} in distributed settings enhances inference throughput without accuracy loss, making it particularly effective for low-resource hardware. Further, utilizing MiniCache, which applies depth-wise KV cache compression across model layer stacks \cite{liu2024minicache} is a potential future research direction.
Beyond compression, \textit{dynamic KV cache allocation} can further optimize memory usage by adjusting cache allocation based on prompt context. Methods such as FINCH \cite{corallo2024finch} implement a context-aware KV cache mechanism, keeping only the most relevant information, could hugely improve the efficiency for DLLMs. 
Another promising future direction is KV cache encoding and compression which transforms KV cache data into a compact bit-stream representation e.g., CacheGen \cite{liu2024cachegen} 

\item \textbf{FL-based Memory Management}:  
Memory management techniques for DLLMs can be further extended to FL environments, presenting a promising approach for optimizing distributed model execution. FL techniques such as adaptive client participation, federated pruning, mixed-precision quantization for training and quantization-aware model aggregation \cite{chen2024mixed} , \cite{zhu2024efficient} can enhance memory efficiency without compromising model performance. Research into intelligent memory-aware FL strategies, such as PEFT and forward-mode auto-differentiation \cite{panchal2024thinking}, could significantly reduce memory consumption during training. These approaches may further minimize the memory footprint of LLM-based FL environments while ensuring efficient training across heterogeneous edge devices. 
\end{itemize}

\noindent $\diamondsuit$ \textbf{Model Optimization}\\
Although several SLMs discussed in this survey  have been used and optimized for small devices, the bottleneck could still be the memory. Memory management is essential for reducing energy costs, latency, and improving computational efficiency \cite{2wu2023fastLLM,4nabli2024accoLLM}.
\begin{itemize}

\item \textbf {Model Size Compression:} 
Reducing model size is essential for deploying DLLMs on resource-constrained devices. A promising future direction is the integration of hybrid model compression techniques, such as pruning combined with quantization algorithms, including QSDP \cite{markov2023quantized} and FPTQ \cite{li2023fptq}, to further reduce model parameters \cite{pisarchyk2020efficient}. Further, advancements in activation quantization and Quantization-Aware Training (QAT), along with Hardware-Algorithm Co-Design, could significantly enhance model compression efficiency. Extending Knowledge Distillation (KD) to additional LLM architectures also remains an active research area  worth exploring. Moreover, PEFT, which fine-tunes only a subset of model parameters, provides an effective solution for scaling LLMs under strict resource constraints. Techniques such as LoRA and adapter modules \cite{46pentyala2024paft,59hu2023llm,15kuang2024federatedscopeLLM} could be further explored to optimize for decentralized resource-limited environments such as FL.

\item \textbf{Advanced Model Partitioning}:
Model partitioning is a promising approach for memory offloading in resource-limited distributed systems. By splitting very large LLMs across multiple trusted devices based on their resource availability \cite{hu2024blockllm}, memory and computation can be efficiently distributed, enabling inference on constrained hardware.
Future research could explore novel memory-efficient, fast, or secure partitioning strategies that optimize load balancing and data locality \cite{6he2024distributedLLM,56li2024tpi}, improving training throughput while mitigating communication overhead in extremely large models. Moreover, adaptive model partitioning, where models dynamically adjust their partitioning scheme based on real-time device constraints and workload demands, could further optimize performance. Another promising future research could explore energy-aware partitioning \cite{wilkins2024offline}, where memory-intensive computations are dynamically offloaded based on device power constraints. Further, hierarchical partitioning, which distributes model components across cloud, edge, and IoT tiers, could optimize both latency and memory efficiency \cite{13duan2024efficientLLM}.

\item \textbf{Advanced Memory-Efficient Attention Mechanisms} \\
Another future work for DLLMs can focus on enhancing memory-efficient attention mechanisms to mitigate the high memory overhead of standard Transformer-based architectures. Several existing strategies could be extended and adapted to improve the efficiency of attention mechanisms in resource-limited distributed LLM settings. 
One promising direction is the integration of dimensionality reduction techniques such  as Linformer \cite{wang2020linformer} into attention computation could significantly reduce memory consumption. Further, investigating alternatives such as Performer \cite{choromanski2020rethinking} (favoring kernelized attention) and LongFormer (utilizing sparse global-local attention) \cite{beltagy2020longformer} could lead to more scalable and adaptable attention mechanisms for distributed systems.
Another critical aspect is optimizing data movement within memory hierarchies. Recent advancements such as DistFlashAttn \cite{3li2024distflashattnLLM} have integrated FlashAttention \cite{dao2022flashattention} into distributed systems, enabling the handling of longer sequences without additional memory overhead. Future research could expand upon these techniques by improving workload balancing across attention tokens, reducing communication overhead, and designing adaptive attention mechanisms that adjust based on real-time resource availability.

\item \textbf{Sparsification}  
Exploring sparsification techniques could further improve the efficiency of Distributed LLMs in cloud-edge settings by reducing unnecessary computations in the attention mechanism. \textit{Sparse attention mechanisms} allow Transformer models to maintain high accuracy while significantly lowering memory and compute costs. 
One promising direction is the \textit{Mixture of Sparse Attention (MoA)} framework \cite{fu2024moa}, which dynamically adjusts sparse attention configurations across different heads and layers to optimize accuracy-latency trade-offs. Further, \textit{Sparse Window Attention (SWA)}, as introduced in ALISA \cite{zhao2024alisa}, enhances inference memory usage and response speed by capturing only the most essential attention patterns.
Future research could further extend these sparsification strategies into distributed settings, enabling adaptive sparsification across multiple nodes for efficient federated and edge-based AI inference. Additionally, integrating structured pruning techniques \cite{sanh2020movement} for distributed settings, or for LLMS \cite{nguyen2024automatic, shi2019understanding} could further reduce computational complexity without significantly impacting accuracy.

\end{itemize}

\noindent $\diamondsuit$ \textbf{Edge Computing and Mobile Intelligence}
\begin{itemize}

\item \textbf{Energy-Efficient Algorithms.} Optimizing power consumption and reducing the carbon footprint is a crucial direction for future research, especially in edge computing, where energy efficiency during training is critical \cite{3li2024distflashattnLLM,4nabli2024accoLLM,11ren2024taskLLM}. As larger LLMs drive higher energy consumption, future efforts can explore techniques such as quantization, gradient compression, and checkpointing to reduce both carbon footprints and operational costs. For instance, model partitioning for fine-tuning LLMs in edge computing \cite{li2025energy} can be integrated with gradient compression to improve efficiency. Implementing scheduling-based adaptive power-capping strategies \cite{petoumenos2015power} and early stopping methods can further minimize energy consumption without compromising model performance.

\item \textbf{Green Edge AI}
Advancing the sustainability of LLM-based solutions on low-power devices necessitates optimizing model size, power consumption, and inference latency \cite{9qu2024mobileLLM,84fu2024serverlessllm}. A promising research direction involves leveraging energy-efficient edge computing techniques \cite{9qu2024mobileLLM}, integrating advanced model compression strategies tailored for LLMs \cite{shi2024efficient}, and employing adaptive backpropagation to mitigate the carbon footprint of training and inference processes \cite{huang2023towards}. Furthermore, Green Hardware-Software Co-Design \cite{huang2024edgellm} can be explored to enhance energy efficiency and computational throughput in DLLMs by optimizing architectural synergies. Furthermore, integrating renewable energy sources and carbon-aware scheduling mechanisms \cite{kim2023greenscale} into edge computing infrastructures presents a compelling avenue for reducing the environmental impact of large-scale LLM deployment. These directions collectively contribute to the development of sustainable, high-performance DLLMs, fostering broader accessibility and ecological responsibility in AI-driven applications.

\item \textbf{Multilingual DLLM} 
Because nodes in a distributed environment are likely to encompass a wide range of demographic and linguistic data, future research can explore novel collaborative multilingual learning approaches for text and speech to enhance the adaptability of LLMs across multiple linguistic contexts in distributed settings \cite{24ye2024fedllmLLM}. Investigating federated multilingual adaptation strategies, where models dynamically adjust language representations based on regional or device-specific training data, could improve global usability.  Additional research directions include: Federated Multilingual Personalization \cite{moskvoretskii2024low}, Training models that dynamically adapt per linguistic region while maintaining global coherence \cite{ahia2024magnet}, Code-Switching and Mixed-Language Adaptation which can swiftly switch the language in one conversation \cite{yoo2024code} for DLLM setting. 

\item \textbf{Personalized DLLM Strategies}:  
In distributed settings, local data is often non-IID, creating challenges for model convergence, fairness, and performance heterogeneity across devices \cite{8qin2023federatedLLM,17ling2024convergenceLLM}. Personalized Large Language Models (PLLMs) and their distributed counterparts are at the forefront of AI research, aiming to tailor interactions and content to individual user preferences. Specialized optimization techniques, such as zero-order optimization and adapter-based fine-tuning, can improve model personalization while ensuring convergence in distributed systems. A critical research direction is the development of adaptive personalization strategies tailored to resource-constrained edge-cloud environments. Lightweight personalized FL  methods, such as adaptive layer modulation and heterogeneous model architectures, can allow models to dynamically adjust complexity based on device constraints \cite{chen2022pfl, tan2022towards}. PEFT techniques, including LoRA and prefix-tuning, could further reduce the computational overhead of personalization while preserving model effectiveness \cite{hu2021lora,tan2022towards}.  Additionally, clustered FL techniques, where clients with similar data distributions share model components, can enhance local accuracy while maintaining a robust global model \cite{liu2023sparse} and could be adapted for DLLMs. Another promising direction involves meta-learning-based personalization for DLLMS, where clients leverage meta-gradients from previous tasks to rapidly adapt to new data distributions with minimal updates \cite{fallah2020personalized}. Expanding adaptive layer-wise PFL algorithms (e.g., Saadati et al. \cite{saadati2025pmixfed}) for DLLM. These techniques can significantly reduce computational costs and make LLM personalization more feasible for low-power devices.

\end{itemize}

\noindent $\diamondsuit$ \textbf{Communication Efficient Algorithms}

\begin{itemize}
    \item \textbf{Advanced Gradient Compression and Pruning} Novel compression strategies \cite{44liu2024resource,89raje2024communication}, Combining pruning, quantization, and novel checkpoint methods can minimize bandwidth requirements can reduce the volume of exchanged gradients, which is essential particularly for large-scale LLM training where network bandwidth is limited \cite{48markov2023quantized,92lin2024splitlora}. Considering LLMs as gradient priors and transforming them into compressed text for zero-shot learning reduces the communication overhead and could be applied to zero-shot distributed LLM setting \cite{wang2024language}. On the other hand, deploying Gradient-Based Language Model Pruner \cite{das2023beyond} and applying sparsification for fine-tuning LLMs \cite{yangsparse} could be the next future direction that could be applied for gradient compression in distributed setting. 
    
\item \textbf{Adaptive Aggregation Mechanisms} Adaptive aggregation mechanisms play a crucial role in improving the efficiency and scalability of DLLMs, particularly in resource-constrained environments such as FL. Integrating lightweight and scalable aggregation techniques, such as those proposed in \cite{choukroun2024adaptive} and \cite{shen2024adaptive}, can enhance both communication and computational efficiency in distributed edge-cloud LLM environments \cite{27zhang2024fedpitLLM}. Furthermore, fine-grained and coarse-grained updates \cite{19zhang2024fedrdmaLLM,68yang2024research} can be strategically combined to balance communication overhead with convergence rates. The use of adaptive aggregation \cite{wang2024dynamic} can further optimize the dynamic nature of client-server interactions \cite{26wu2024fedbiotLLM}. Additionally, extending adaptive attention-based aggregation methods for LLMs \cite{ji2019learning} represents a promising direction for future research.

\item \textbf{Hierarchical or Clustering Aggregation} Deploying hierarchical or clustering aggregation mechanisms, where nodes are grouped based on metrics —such as data similarity, resource availability, processing speed, and geographical location—can be highly beneficial for resource limited edge-cloud environments. Traditional aggregation schemes operate centrally, leading to increased communication overhead in edge-cloud settings where bandwidth is limited. Clustering-based aggregation can be performed among only a small group of trusted nodes that complement each other in terms of latency and computational constraints \cite{wang2021resource, chen2022federated}. Integrating adaptive aggregation clustering mechanisms, such as those proposed in \cite{jia2024adaptive, ali2025dy}, for LLMs could be a promising direction for DLLM future research.

\item \textbf{Semantic-based Communication} These approaches significantly reduce communication costs in distributed edge settings by transmitting only meaningful and essential information rather than raw model data \cite{99zhang2024distributed}. Introducing task-specific communication methods that focus on conveying critical task-specific model updates or insights—rather than transmitting raw parameters—holds great potential for further bandwidth reduction in both cloud and edge environments. Moreover, extending dynamic mechanisms for methods such as ACCO (Accumulate while you Communicate) \cite{4nabli2024accoLLM} can enhance communication and computational efficiency in large-scale LLM-based distributed environments. Additionally, integrating advanced semantic encoding techniques \cite{abrahamyan2021learned}, where model updates are distilled into more compact and contextually relevant representations, presents another promising direction for DLLM future research.

\end{itemize}

\noindent $\diamondsuit$ \textbf{Privacy-Preserving Distributed LLMs} 
\begin{itemize}
\item \textbf{Secure DLLMs}: 
Enhancing the privacy and security of Distributed LLM environments is a critical research direction. Integrating efficient and scalable extensions of homomorphic encryption, multi-party computation, secure aggregation, and differential privacy—techniques previously employed in decentralized learning—specifically tailored for LLMs, could be highly impactful. Exploring privacy-preserving inference techniques \cite{16xu2023fwdllmLLM, 10khoshsirat2024decentralizedLLM} for low-resource edge-cloud environments can further strengthen data confidentiality while maintaining computational efficiency. Leveraging Trusted Execution Environments (TEEs), which provide isolated and secure computation, in conjunction with specialized hardware accelerators such as GPUs and TPUs \cite{ren2023accshield}, can enhance processing speed, optimize energy consumption, and fortify the overall security posture of the system \cite{28huang2024frameworkLLM}. Furthermore, employing adaptive encryption mechanisms that dynamically adjust based on resource and network constraints \cite{28huang2024frameworkLLM} is essential. These techniques must be optimized to minimize performance overhead while ensuring robust data protection.

\item \textbf{Federated Learning}:  
Although FL provides a privacy-preserving framework by ensuring that raw data remains on local devices, advanced cryptographic techniques such as secure aggregation and differential privacy \cite{7borzunov2024distributedLLM,21ye2024openfedllmLLM} can further enhance data security and confidentiality in LLM-based FL training. Hybrid federated fine-tuning approaches, combined with communication-efficient techniques such as FedKSeed \cite{8qin2023federatedLLM}, which reduces communication overhead during training, offer promising directions for improving scalability. Additionally, exploring secure LLM-based FL algorithms designed for synchronous training and adapting them for asynchronous settings represents another critical research area . Another key challenge is the development of adaptive real-time malicious attack detection mechanisms in FL. Expanding these mechanisms to detect a wide range of attack types, including adversarial attacks, and applying them to diverse datasets could significantly enhance robustness in heterogeneous, large-scale, multimodal FL environments \cite{23ye2024safetyLLM}. Furthermore, securing Federated Transfer Learning (FTL) frameworks with differential privacy or secure aggregation remains an open research challenge that warrants further investigation.

\end{itemize}

\noindent $\diamondsuit$ \textbf{Addressing Heterogeneity in DLLMs}
    \begin{itemize}

\item \textbf{Data Heterogeneity}  
Data heterogeneity, particularly in non-IID distributed environments, poses a significant challenge in distributed systems, especially in LLM-based distributed edge-cloud settings. Future research directions to tackle data heterogeneity could be Expanding Ferret algorithm, which combines first-order and zero-order optimization methods \cite{20shu2024ferretLLM}, to better support heterogeneous data distributions. One potential enhancement is client-based adaptive update mechanisms, where gradient updates are weighted based on local data distributions to accelerate convergence rate and reduce global-local model divergence. Moreover, aggregating local LLM updates dynamically based on data heterogeneity metrics could be the next idea. By assigning adaptive importance weights to different client models, a more generalized and fair global model can be trained, improving representation across diverse local datasets. Introducing synthetic data augmentation techniques, such as instruction augmentation, adversarial training \cite{efthymiadis2024advanced}, or Generative Adversarial Networks (GANs), to mitigate non-IID effects and improve model accuracy could be the next future direction in addressing  \cite{25wang2024federatedLLM} \cite{behera2022fedsyn}. Synthetic data can enhance dataset diversity, improve generalization, and be particularly beneficial for zero-shot or few-shot learning. By addressing data heterogeneity at multiple levels, future research can develop fairer, more robust, and generalizable DLLMs that effectively function across diverse resource-constrained environments.

\item \textbf{Task and Model Heterogeneity}:  
Addressing diverse learning objectives and supporting heterogeneous data modalities in federated and distributed learning frameworks is a key future research direction. Extending federated multi-task learning frameworks to optimize multiple learning objectives concurrently or leveraging co-optimization techniques for training foundation models while simultaneously optimizing different objectives presents a promising avenue \cite{40yu2023federated, chen2023fedbone}.  Research into cross-domain alignment for multi-modal DLLMs could enhance predictive analytics by enabling seamless knowledge transfer across heterogeneous data sources. Such techniques could be particularly beneficial for applications that require fusing structured (e.g., tabular financial data) and unstructured data (e.g., text or images) to improve decision-making accuracy.  
Another important direction is distributed time-series forecasting for multi-modal adaptation, where time-sensitive modalities—such as financial transactions, IoT sensor streams, and real-time healthcare monitoring—demand specialized distributed forecasting techniques \cite{33liu2024timeFFMLLM}. Investigatin federated temporal learning mechanisms that efficiently process time-evolving and asynchronous data distributions could further enhance DLLM applications in dynamic environments.  Moreover, integrating ensemble learning strategies within federated and distributed settings could enhance overall model robustness by combining the strengths of diverse models. For instance, adaptive mixture-of-experts (MoE) architectures could dynamically allocate specialized submodels to different tasks, optimizing both efficiency and performance across varying objectives \cite{huang2025ensemble}.

\item \textbf{Hardware Heterogeneity}:  
Heterogeneous platform support is essential for scalable and efficient deployment of Distributed LLMs  across diverse hardware architectures. Seamlessly integrating GPUs, TPUs, edge accelerators, and domain-specific AI hardware \cite{6he2024distributedLLM,77xu2024hethub} will be crucial for next-generation large-scale deployments. To ensure robust model training in such environments, techniques such as partial gradient updates, asynchronous processing, and distributed checkpointing \cite{51liu2024asynchronous,70tang2024fusionllm} can help stabilize convergence under high network and hardware heterogeneity. However, these approaches require rigorous consistency analysis to prevent training instability.  Dynamically scheduling model updates in federated LLM training, or using Adaptive learning rate scheduling based on device-specific resource constraints and the trade-off with convergence speed can significantly improve training efficiency, particularly in highly heterogeneous environments \cite{20shu2024ferretLLM}.  Additionally, adaptive client participation strategies, where devices contribute to training based on their computational capacity, network bandwidth, and energy availability, could improve both personalization efficiency and system scalability \cite{li2024adafl}.     

\end{itemize}

\noindent $\diamondsuit$ \textbf{Vision-Language Models}
 \begin{itemize}
{\item \textbf{Mitigation of Hallucination:} Reducing hallucinations in VLMs is one of the critical concern to address for ensuring the reliability and appropriateness of the generated content. Hallucinations refer to an event when a model generates irrelevant or partially incorrect outcome. This issue is persistent in high-stakes applications, such as healthcare care and public safety, where it can lead to incorrect decision-making. Hallucination may occur due to the inherent complexity of information integration from different modalities (e.g., text, video, image) \cite{143vlmfavero2024multi}. To reduce hallucinations, several techniques can be applied. VLMs need to recognize intra-modality elements as well as the contextual relevance and interrelations among different modalities \cite{145vlmguan2024hallusionbench}. Besides, the VLMs need to be trained with a wide range of data along with effective methods to synthesize the knowledge and apply that across various scenarios \cite{144vlmwang2024haloquest}. In addition, preventing overfitting of the VLMs can reduce hallucination, which can enable VLMs to perform accurately not only on training data, but also prevent VLMs from hallucination when faced with divergent real-world data.

\item \textbf{Fine-grained Correlation Modeling of Visual and Textual Components:} Modeling the fine-grained correlation of local vision-language correspondence information is an emerging area of research \cite{139vlmyao2022detclip, 141vlmxu2022groupvit}. This strategy enables the large VLMs to identify pixels and patches within visual data more accurately, strengthening their reliability in crucial prediction tasks like image captioning, object detection, scene understanding, and semantic segmentation. Kuang et al. provided a comprehensive overview of Visual Question Answering (VQA) as a benchmark task in MLLMs \cite{kuang2024natural}. They further covered recent advancements in model development, vision-language pretraining, knowledge reasoning, dataset curation, and evaluation metrics \cite{kuang2024natural}. Although research in this area is currently limited, as evidenced by the few studies \cite{141vlmxu2022groupvit, 139vlmyao2022detclip, 140vlmdou2022coarse, 142vlmzhong2022regionclip}, there is a strong expectation for broader investigation into fine-grained VLM pre-training specifically for zero-shot dense prediction tasks. This anticipated expansion in research aims to enhance the capabilities of VLMs in accurately processing and predicting detailed visual tasks without prior explicit examples.

\item \textbf{Commonsense Reasoning:} Another future direction is to improve the commonsense reasoning of VLMs. This refers to the deeper understanding of intuitive logic and real-world knowledge that humans commonly use \cite{146vlmbitton2024visual}. Comprehensive datasets and algorithms must be integrated into VLMs to achieve better context interpretation capability and generate an appropriate outcome that meets realistic human behavior and expectations. Such capability can improve the effectiveness of the VLMs in wide-range of applications, from customer service to emotion understanding and autonomous vehicles, where scene and context understanding, and anticipations of future events or possibilities are crucial to prove the systems' effectiveness \cite{147vlmkwon2024toward, 148vlmxenos2024vllms}. For instance, using commonsense reasoning, a VLM enhanced with commonsense reasoning should understand that a child playing by the side of a road or near the street may suddenly run towards the driving road, prompting the vehicle to slow down. 
\item \textbf{Robust VLMs:} Enhancing the resilience of VLMs against sophisticated cyberattacks is a major challenge for the VLMs. The reliability of the VLMs will be at stake if the cyberattacks are not counteract, including backdoor attack that embed hidden malicious behavior \cite{152vlmzhu2024seer, 153vlmlyu2024trojvlm}, model poisoning attacks where the model's performance is degraded through manipulation of the training data \cite{155vlmliu2024survey}, or adversarial attacks where subtle changes of the model input may deceive the VLMs \cite{154vlmwang2024break}. To combat such sophisticated threats, comprehensive research on the development of an advanced defensive mechanism that can identify and neutralize such threats are required. Several promising efforts are already being conducted towards this goal such as, \cite{149vlmhossain2024sim, 150vlmschlarmann2024robust, 151vlmhossain2024securing}. In addition, future research directions could focus on improving the adaptability of VLMs to emerging cyber threats by adjusting security measures by analyzing the attack patterns, model exposure to certain scenarios, and learning from recent vulnerabilities. 
\item \textbf{Energy Efficiency:} Reducing the energy consumption of large VLMs by enhancing their model architectures to require less computational power for training and inference is a crucial future research directions. Such improvement of the VLMs can make them suitable to deploy on edge devices, thereby extending their utility in environments where computing and power and resources are limited \cite{156vlmlaurenccon2024matters, 157vlmxing2024survey, 158vlmlee2024vhelm}. For instance, deploying a VLM on a mobile device to assist patients  might need energy-efficient model that works without draining the mobile device battery quickly. Potential solutions could be to develop lightweight VLM model architectures and the integration of efficient pruning and quantization techniques. Pruning technique helps to remove non-essential part of the VLM model architectures, such as neurons and edges or tokens that contribute less to the final outcome, thereby optimizing model size and accelerating  model inference. Besides, quantization can contribute by further reducing the energy demand by lowering  the precision of the numbers used during computation, enabling VLMs to consume less power and run faster \cite{159vlmxie2024advancing, 160vlmhao2024quantized, 161vlm10903230}. Another solution could be to apply knowledge distillation approach, where a student model, which is smaller in size and more energy efficient is trained to imitate the behavior of a pre-trained teacher model, which is larger in size. Such approach can enable to develop a lightweight VLM that is well-suited for resource-constrained environment.

\item \textbf{Mathematical Reasoning:} Enhancing mathematical reasoning capabilities in VLMs is essential to improving their performance in technical and scientific applications. Unlike traditional language tasks, mathematical reasoning involves symbolic understanding, logical deductions, and numerical computations, which are challenging for current VLMs \cite{166vlmzoudynamath,167vlmlu2021inter}. Moreover, recent VLMs often lack the ability to perform multi-step reasoning,  which is essential for better mathematical reasoning \cite{168vlmpeng2024multimath}. Furthermore, current VLMs exhibit inconsistent performance when faced with different variants of the same problem, as they tend to rely on memorized patterns from pretraining rather than genuine problem-solving \cite{169vlmlu2023mathvista}.
Future research should focus on incorporating structured representations, such as formal logic and symbolic reasoning, to improve the numerical and analytical problem-solving ability of VLMs. Additionally, integrating Chain of Thought (CoT) reasoning could improve logical consistency in multi-step problem-solving, making models more reliable even when faced with multiple variations of the same problem \cite{168vlmpeng2024multimath}. Beyond CoT, reinforcement learning techniques could be employed to correct reasoning errors at each step rather than solely at the final answer, leading to more robust mathematical reasoning \cite{169vlmlu2023mathvista,166vlmzoudynamath}. Furthermore, integrating external computational engines within VLMs could significantly enhance their ability to handle complex arithmetic and algebraic tasks, bridging the gap between pattern-based reasoning and true mathematical understanding.

\item \textbf{Continual Learning in VLMs:} Continual Learning is a crucial capability for VLMs that enables them to adapt to new knowledge over time without catastrophic forgetting. Unlike traditional static training, where models are trained once on a fixed dataset, continual learning involves learning from a continuous stream of data while retaining previously acquired knowledge. This is particularly important for VLMs as they operate in dynamic environments where new concepts, objects, and tasks emerge over time. Some recent studies have explored various techniques to enhance the continual learning capabilities of VLMs. For instance, Zhou et al. \cite{171vlmzhou2025learning} proposed a projection fusion-based framework, which preserves past knowledge by freezing pre-trained VLM encoders and adding task-specific projection layers, thereby mitigating catastrophic forgetting. Similarly, Zheng et al. \cite{172vlmzheng2023preventing} introduced a zero-shot continual learning framework, which prevents zero-shot transfer degradation in CLIP-based VLMs by using feature-space distillation and parameter-space regularization, ensuring that knowledge acquired from prior tasks. Another approach, CLAP4CLIP \cite{170vlmjha2024clap4clip}, applies probabilistic fine-tuning to improve cross-modal alignment and uncertainty estimation in continual learning scenarios. To advance continual learning in VLMs, future research should focus on developing structured memory mechanisms, such as knowledge graphs and external memory banks, to dynamically store and retrieve information without disrupting learned representations. Further, incorporating self-supervised objectives could enable VLMs to learn continuously from unlabeled data streams, ensuring that they remain adaptive without excessive reliance on human annotations. Enhancing cross-modal consistency regularization through techniques like contrastive learning with historical embeddings can further mitigate representation drift and maintain knowledge alignment across tasks.

\item \textbf{Explainability and Transparency:} VLMs are becoming increasingly powerful, but their lack of explainability poses significant challenges in critical applications such as healthcare, finance, and autonomous systems. Unlike traditional machine learning models, VLMs process both visual and textual data through multimodal fusion, making their decision-making process difficult to interpret. While extensive research has focused on improving the explainability of LLMs \cite{175vlmyuan2024back, 173vlmzhao2024explainability}, similar efforts are now being extended to VLMs to enhance their transparency and trustworthiness. Kazmierczak et al. \cite{176vlmkazmierczak5106267explainability} conducted a comprehensive study on explainability in vision foundation models, identifying key trends and challenges in integrating interpretability into large-scale vision models. Their study highlights how the increasing complexity of these models exacerbates transparency issues, even as they are leveraged to develop more explainable AI systems. Similarly, in \cite{174vlmben2024lvlm}, the authors introduced LVLM-Interpret, an interactive tool designed to analyze the inner workings of LVLMs by visualizing attention maps, relevancy scores, and causal explanations, helping to uncover biases and reasoning failures in multimodal tasks. Future research should focus on developing inherently interpretable VLMs, inspired by recent LLM-based techniques, to enhance transparency by embedding explainability directly into model architectures rather than relying on post-hoc methods. Further exploration of causal interpretability methods could provide deeper insights into how different features contribute to model decisions, ensuring more reliable outputs. Another critical direction is multimodal consistency verification, as current VLMs often struggle with inconsistencies when processing vision and language inputs simultaneously. Addressing these challenges will improve the trustworthiness and applicability of VLMs in real-world scenarios.}
\end{itemize}

\noindent $\diamondsuit$ \textbf{Decentralized Multi-Modal LLMs}
    \begin{itemize}
{\item \textbf{Advanced Multi-Modal Learning}  
Future research could explore multi-modal learning in DLLMs, where edge-based systems must concurrently handle text, images, speech, sensor, and time-series data \cite{33liu2024timeFFMLLM,74du2024distributed,11ren2024taskLLM}. Expanding multi-modal capabilities in DLLMs requires efficient quantization and compression techniques tailored to different data types, ensuring optimized memory usage and computational efficiency in resource-constrained environments \cite{16xu2023fwdllmLLM}. Additionally, further research could focus on integrating \textit{Efficient Multi-Modal Representation Learning} techniques Developing cross-modal fusion \cite{lyu2024unibind}, grouping \cite{huang2024efficient}, attention pruning, and fine-tuning the Layer Normalization \cite{zhao2023tuning} to reduce computational overhead. Next promising future direction is 
applying modality-specific compression \cite{jiang2023multi} and entropy-aware quantization \cite{wang2024q} to optimize inference. Moreover, Implementing adaptive offloading strategies \cite{zhu2024energy} to DLLM for balancing energy efficiency, and  Developing novel multi-modal sparsification methods \cite{zhou2024multimax} could further enhance scalability and efficiency in distributed AI environments.}

\item \textbf{Distributed Model Optimization for MLLMs}
Expanding MLLMs into distributed settings is a critical research direction, particularly for resource-constrained edge environments \cite{74du2024distributed}. Given that different modalities (e.g., text, image, audio, video, and structured data) may originate from distinct sources, developing adaptive optimization mechanisms that dynamically adjust model depth, architecture, and hardware utilization based on both device constraints and data flow modality could significantly enhance scalability in heterogeneous distributed systems. This strategy can potentially enable efficient deployment across a spectrum of devices, ranging from low-power edge nodes to high-performance cloud servers. A promising approach is modality-aware distributed training, which can be integrated with adaptive scheduling mechanisms to allocate computational resources based on data modality. For instance, high-resolution video processing demands specialized hardware accelerators, while long-sequence text processing may benefit from distributed attention mechanisms for memory efficiency \cite{li2024advances}.  
Since communication efficiency remains a key challenge in distributed learning, hybrid MLLM approaches that incorporate communication-reduction techniques in distributed setting—such as gradient compression, quantization, and activation checkpointing—can further minimize communication overhead and latency in multi-modal learning scenarios. Future work should explore dynamic communication strategies, where model updates are dynamically optimized based on modality-specific data transmission constraints.  
Additionally, knowledge distillation remain under-explored in the context of distributed MLLMs. Techniques such as channel pruning \cite{molchanov2016pruning, he2017channel}, weight factorization \cite{denton2014exploiting}, and cross-modal distillation \cite{albanie2018emotion} could be leveraged to create specialized, lightweight sub-models from large models, making distributed deployment feasible for memory-constrained devices \cite{huang2024efficient}.

\end{itemize}

\noindent $\diamondsuit$ \textbf{Other Emerging Areas of Research} \\ In addition to the above-mentioned future research directions, we recognize emerging topics in the LLM domain that have the potential for future research.

    \begin{itemize}
    \item \textbf{Score Entropy Discrete Diffusion models} Lou et al \cite{lou2023discrete} introduces Score Entropy Discrete Diffusion (SEDD), a novel framework that extends diffusion models to discrete data domains such as natural language. SEDD leverages a new score entropy loss that generalizes score matching for discrete spaces, leading to significant improvements in text generation, inference efficiency, and controllability. The integration of discrete diffusion models such as SEDD \cite{lou2023discrete} into distributed LLM and MLLM research is a promising research opportunity to improve scalability, efficiency, and robustness in decentralized AI systems. By reducing reliance on sequential decoding, enabling parallelized inference, and optimizing memory efficiency, these models can revisit  the foundations of distributed learning for large-scale multi-modal AI. Future research can focus on adapting SEDD-based approaches to hierarchical, federated, and edge-cloud (M)LLM architectures, ensuring efficient and privacy-preserving deployment in various applications.

\item \textbf{Large Concept Models }  
 Large Concept Models (LCMs)\footnote{LCM \cite{themeta2024large} introduces a novel way to process language and other data by focusing on high-level ``concepts'' instead of individual words or tokens, which is how most current AI models work. A ``concept''  is similar to an idea or meaning often represented by a sentence. The model operates in a universal space where meaning is detached from specific languages or modalities (e.g., text, speech, or images). This approach enables the LCM to handle tasks like summarization and text expansion efficiently across 200 languages. Unlike traditional AI models, LCM mimics human-like reasoning by working hierarchically and  addresses ideas first and then adding details. This makes it more flexible and capable of generalizing without extensive fine-tuning. The LCM relies on a sentence embedding space (SONAR) to represent concepts and achieves impressive performance on multilingual tasks with fewer computational resources compared to token-based models \cite{themeta2024large}.
}\cite{themeta2024large}  can address major challenges of distributed MLLMs  (such as scalability, high computational burden, and inefficiencies in processing multimodal data types such as text, video, and audio) by introducing a hierarchical abstraction layer that reasons over high-level concepts rather than individual tokens. This can  significantly reduce computational complexity and enable shorter sequences that deal with the quadratic scaling limitations of traditional transformers. Their modular, language- and modality-agnostic design facilitates seamless integration of multiple modalities without competition,   simplifies the alignment of heterogeneous data in distributed systems, and improves scalability and resource efficiency. By enabling localized processing of abstract concepts in edge computing environments, LCMs reduce communication overhead between nodes while supporting  multimodal tasks. Future research can explore optimizing embedding spaces such as SONAR for low-resource settings and extending hierarchical reasoning to paragraph or section-level abstractions and further enhance the robustness and scalability of distributed MLLMs for real-world applications.

\item \textbf{DeepSeek-R1-Zero}   Guo et al \cite{guo2025deepseek} introduced a novel RL-based framework for advancing reasoning capabilities in LLMs without extensive reliance on supervised data. They  proposed DeepSeek-R1-Zero, which leverages pure RL to exhibit self-evolution in reasoning tasks and achieving significant improvements in benchmarks such as AIME and MATH-500. In order to address issues related to readability and language mixing in outputs,   DeepSeek-R1 uses a multi-stage training approach involving cold-start data, reasoning-focused RL, and rejection sampling. This hybrid pipeline allows the model to converge faster and produce  user-friendly high-performance reasoning outputs. Further, the reasoning abilities of DeepSeek-R1 were distilled into smaller dense models and represented promoting results on reasoning benchmarks such as GPQA and LiveCodeBench. Guo et al  also discussed the challenges of RL in small models and highlights the potential for iterative improvements through advanced techniques in RL and distillation. This approach establishes a scalable pathway for creating efficient  reasoning-optimized LLMs while open-sourcing tools for broader  adoption \cite{guo2025deepseek}. 
\end{itemize}

\section{Conclusion}
\label{conclusion}

This paper presented a  survey on distributed LLMs and MLLMs. We explored  advancements, challenges, and future directions of a wide range of existing studies. We described the implementation challenges while performing distributed training, inference, and deployment. Further, we  described emerging challenges such as  scalability and growing computational demands of these models. We highlighted how decentralized/distributed approaches offer essential solutions to enhance scalability and deployment at the edge.  Recent research has enabled distributed training and inference across diverse computational resources. We provided an overview of LLMs, VLMs, MLLMs, and SLMs, with a   particular emphasis on decentralizing LLMs and MLLMs.  

We further categorized selected studies based on six critical aspects of \textit{Distributed Training}, \textit{Distributed Inference and Optimization};  \textit{Distributed Computing Infrastructures},  \textit{ Federated Learning and Fine-tuning};  \textit{Edge Computing and Mobile Intelligence}, and  \textit{Communication Efficiency in Distributed Systems} to have a structured analysis of the progress and gaps in the field. We further summarized practical innovations such as edge deployment in low-cost devices and parameter-efficient techniques, while highlighting challenges such as  hallucination and data heterogeneity. We further provide an outline of future research directions to explore the need to develop novel strategies at the intersection of distributed learning and (Multimodal)LLMs. For a number of selected technical papers that we summarized, we provided the potential future directions that are aligned with this survey. This survey (which is evolving to include more studies in the next versions) serves as a valuable resource for researchers and practitioners who are interested in conducting research or using  advanced distributed (M)LLMs across different application  domains. 

\textbf{Limitations}: Although this survey focused on providing a review of recent advances in distributed/decentralized LLM and MLLM studies, due to the large number of publications on these topics, we may not have covered all related works. In order to mitigate this, we have developed a GitHub page for the paper to include other relevant studies. 

{\color{blue}The corresponding GitHub page is available at: \href{https://github.com/solidlabnetwork/awesome-distributed-LLM}{\textbf{Link}}.

We kindly ask the researchers and practitioners to share any related works/suggestions with us via email to \textbf{solidlabnetwork@gmail.com} and cc \textbf{hadi.amini@ieee.org}. Those suggestions might be reviewed for relevance to focus on our survey and might be included in the GitHub page accordingly.}

\bibliographystyle{unsrt}
\bibliography{LLm-cite}

~~~\section*{Appendix A: Details of Selected  Small Language Models}

\subsection{\href{https://huggingface.co/gpt2} {GPT-2 Small} }
\noindent \textbf{Parameters}: 124M \\
\textbf{Description}: The GPT-2 Small variant is the most lightweight model in the GPT-2 series, making it an ideal choice for resource-constrained devices. This transformer-based model strikes a balance between computational efficiency and performance by requiring significantly less memory and computational power. Despite its compact size, GPT-2 Small retains the coherence in text generation and the robust, versatile language modeling capabilities characteristic of the GPT-2 series. Furthermore, it is highly compatible with common Deep Neural Network (DNN) architectures, simplifying the fine-tuning process across heterogeneous devices.
In distributed environments such as FL, GPT-2 Small offers substantial advantages by significantly reducing communication costs. Only a small number of parameters need to be shared with the central server, making it an efficient solution for FL setups. This lightweight LLM is a practical choice for scenarios where computational and communication constraints are critical considerations.\\
\textbf{Link}: \href{https://huggingface.co/gpt2}{GPT-2 on Hugging Face} \\
 \begin{center}
\begin{tcolorbox}
\vspace{-0.05in}
\tiny
\noindent \textbf{Python Command:}

\begin{verbatim}
from transformers import GPT2LMHeadModel, GPT2Tokenizer

# Load the GPT-2 Small model and tokenizer
model_name = "gpt2" 
model = GPT2LMHeadModel.from_pretrained(model_name)
tokenizer = GPT2Tokenizer.from_pretrained(model_name)

\end{verbatim} 
\end{tcolorbox}
\end{center}

\subsection{\href{https://huggingface.co/distilgpt2}{DistilGPT-2}}
\noindent \textbf{Parameters}: 82M \\
\textbf{Description}: The lightweight version of the GPT-2 model leverages knowledge distillation, a technique where a smaller student model (DistilGPT-2) learns from a larger teacher model (GPT-2). This process involves the smaller, less computationally complex model estimating the output distribution of the larger model. As a result, DistilGPT-2 successfully reduces computational requirements by approximately 40\% while retaining 97\% of the original GPT-2’s performance. The reduced model size requires less memory and bandwidth for transferring model weights during the broadcasting and aggregation stages in distributed settings.
The communication and computational efficiency of DistilGPT-2 makes it an excellent choice for small IoT devices and smartphones. Additionally, this model demonstrates reduced vulnerability to overfitting in non-IID heterogeneous environments, which are common in decentralized settings. Notably, DistilGPT-2 is both faster and smaller than GPT-2 Small—for instance, its average forward pass is approximately twice as fast—with only minimal performance degradation. \\
\textbf{Link}: \href{https://huggingface.co/distilgpt2}{DistilGPT-2 on Hugging Face} \\ 
\begin{center}
\begin{tcolorbox}
\vspace{-0.05in}
\tiny 
\textbf{Python Command:}
\begin{verbatim}

from transformers import GPT2LMHeadModel, GPT2Tokenizer

# Load the DistilGPT-2 model and tokenizer
model_name = "distilgpt2"
model = GPT2LMHeadModel.from_pretrained(model_name)
tokenizer = GPT2Tokenizer.from_pretrained(model_name)

\end{verbatim} 
\end{tcolorbox}
\end{center}

\subsection{\href{https://huggingface.co/EleutherAI/gpt-neo-125M} {GPT-Neo 125M}}
\noindent \textbf{Parameters}: 125M \\
\textbf{Description}: Released by EleutherAI, GPT-Neo is an imitation of the GPT-3 architecture, trained on the Pile, an 800GB dataset. GPT-Neo has been widely adopted for tasks such as text generation and language understanding, owing to its robustness in language modeling. Although it is slightly heavier than DistilGPT-2, this fully open-source LLM provides customizable models tailored for different domains and applications for public use. These models can be easily adapted for heterogeneous devices with non-IID datasets.
Trained on the diverse web-based dataset "The Pile," GPT-Neo exhibits strong generalization capabilities, which are particularly advantageous in decentralized settings where varied datasets are distributed across devices. However, a notable drawback of GPT-Neo is its potential to generate socially unacceptable content due to the nature of the data it was trained on. As such, implementing content-filtering mechanisms and algorithms is highly recommended to mitigate this risk.  \\
\textbf{Link}: \href{https://huggingface.co/EleutherAI/gpt-neo-125M}{GPT-Neo 125M on Hugging Face} \\
\begin{center}
\begin{tcolorbox}
\vspace{-0.05in}
\tiny 
\textbf{Python Command:}
\begin{verbatim}
from transformers import GPTNeoForCausalLM, GPT2Tokenizer

# Load the GPT-Neo model and tokenizer
model_name = "EleutherAI/gpt-neo-125M"
model = GPTNeoForCausalLM.from_pretrained(model_name)
tokenizer = GPT2Tokenizer.from_pretrained(model_name)

\end{verbatim} 
\end{tcolorbox}
\end{center}

\subsection{\href{https://huggingface.co/meta-llama/Llama-2-7b-hf} {LLaMA-2 7B}}
\noindent \textbf{Parameters}: 7B \\
\textbf{Description}: This transformer-based LLM, part of the LLaMA 2 series introduced by Meta, is available in three size variants: 7 billion, 13 billion, and 70 billion parameters. While the smallest version (7B) is still very large compared to other lightweight LLMs, it has consistently demonstrated strong performance, particularly in complex tasks such as multilingual NLP, automated code generation, and debugging. LLaMA 2 has also been successfully deployed in cross-silo FL settings (e.g., APPFL) with heterogeneous cloud and high-performance resources \cite{li2024advances}.
However, this high performance comes at the cost of significant computational and communication overhead, making it less suitable for resource-constrained devices. As a result, LLaMA 2 is not the most practical choice for decentralized environments with limited hardware capabilities, where efficiency and low resource consumption are critical considerations. \\
\textbf{Link}: \href{https://huggingface.co/meta-llama/Llama-2-7b-hf}{LLAMA 7B on Hugging Face} \\
\begin{center}
\begin{tcolorbox}
\vspace{-0.05in}
\tiny 
\textbf{Python Command:}
\begin{verbatim}
from transformers import LlamaForCausalLM, LlamaTokenizer

# Load the Llama-2 model and tokenizer
model_name = "meta-llama/Llama-2-7b-hf"
model = LlamaForCausalLM.from_pretrained(model_name)
tokenizer = LlamaTokenizer.from_pretrained(model_name)
\end{verbatim}
\end{tcolorbox}
\end{center}

\subsection{\href{https://huggingface.co/albert-base-v2} {ALBERT-Base} }
\noindent \textbf{Parameters}: Base version ~11M \\
\textbf{Description}: ALBERT (A Lite BERT) is a transformer-based LLM designed by applying parameter reduction techniques to the BERT language model. ALBERT comes in various versions and sizes, with the most lightweight being ALBERT-Base, which contains only 10\% of the parameters of the BERT-Base model (a 110-million-parameter LLM). As the most lightweight LLMs introduced to date, ALBERT is an excellent candidate for edge devices and smartphones, where memory and computational resources are limited.
ALBERT is renowned for its powerful natural language understanding and sentence relation extraction capabilities, which are particularly valuable in decentralized settings. It employs a novel loss function for Sentence Order Prediction (SOP), which enhances its ability to generate coherent sentences. Additionally, ALBERT utilizes cross-layer parameter sharing, where parameters from one layer are reused across other layers, significantly reducing the model's parameter count without noticeable accuracy loss. 
Furthermore, ALBERT implements factorized embedding, splitting large matrices into smaller ones, drastically reducing memory and computational resource requirements. These features make ALBERT a suitable choice for distributed ML applications such as FL. The downside of this lightweight LLM is it's limited ability in different NLP task domains e.g., text generation and creative question answering. Hence, the parameter reduction, limits the ability of ALBERT for complex tasks requires contextual understanding and reasoning. \\
\textbf{Link}: \href{https://huggingface.co/albert-base-v2}{ALBERT on Hugging Face} \\
\begin{center}
\begin{tcolorbox}
\vspace{-0.05in}
\tiny 
\textbf{Python Command:}
\begin{verbatim}
from transformers import AlbertForMaskedLM, AlbertTokenizer

# Load the ALBERT-BASE model and tokenizer
model_name = "albert-base-v2"
model = AlbertForMaskedLM.from_pretrained(model_name)
tokenizer = AlbertTokenizer.from_pretrained(model_name)
\end{verbatim}
\end{tcolorbox}
\end{center}

\subsection{TinyBERT}
\noindent \textbf{Parameters}: ~14.5M \\
\textbf{Description}: TinyBERT is a compact version of BERT, developed using a novel transformer distillation method. This small LLM learns from a larger BERT-Base model, transferring its knowledge to a smaller student model with only a minor performance drop (a 4\% performance reduction on the GLUE benchmark dataset). TinyBERT requires less memory, offers ~10X faster inference, and demands lower computational power, making it well-suited for resource-constrained devices.
However, TinyBERT lacks the robustness required for complex NLP tasks such as text and code generation or open-ended tasks. Instead, similar to ALBERT-Base, TinyBERT is primarily designed for natural language understanding (NLU) tasks, including classification, question answering, and sentiment analysis. For tasks within the NLU domain, TinyBERT is among the best options for devices with limited computational resources   \\
\textbf{Link}: \href{https://huggingface.co/huawei-noah/TinyBERT_General_4L_312D}{TinyBERT on Hugging Face} \\
\begin{center}
\begin{tcolorbox}
\vspace{-0.05in}
\tiny 
\textbf{Python Command:}
\begin{verbatim}
from transformers import BertForSequenceClassification, BertTokenizer

model_name = "huawei-noah/TinyBERT_General_4L_312D"
model = BertForSequenceClassification.from_pretrained(model_name)
tokenizer = BertTokenizer.from_pretrained(model_name)
\end{verbatim}
\end{tcolorbox}
\end{center}

\subsection{DistilBERT}
\textbf{Parameters}: ~66M \\
\textbf{Description}: This BERT-based lightweight LLM utilizes knowledge distillation, similar to TinyBERT, but places greater emphasis on performance rather than size reduction. The model is 40\% smaller than the BERT-Base model (110 million parameters) and achieves a 60\% speedup in computation, with only a 3\% performance reduction.
While it offers better capabilities for complex tasks compared to TinyBERT, it still falls short of the performance of the BERT-Base model for text generation. Nonetheless, it represents a well-balanced choice for NLU tasks on resource-constrained devices, where a tradeoff between performance and size is critical.  \\
\textbf{Link}: \href{https://huggingface.co/distilbert-base-uncased}{DistilBERT on Hugging Face} \\
\begin{center}
\begin{tcolorbox}
\vspace{-0.05in}
\tiny 
\textbf{Python Command:}
\begin{verbatim}
from transformers import DistilBertForSequenceClassification,
DistilBertTokenizer

model_name = "distilbert-base-uncased"
model = DistilBertForSequenceClassification.from_pretrained(model_name)
tokenizer = DistilBertTokenizer.from_pretrained(model_name)
\end{verbatim}
\end{tcolorbox}
\end{center}

\subsection{MobileBert}
\textbf{Parameters}: 25.3 M \\
\textbf{Description}: MobileBERT is a compact extension of the BERT-Base LLM, specifically designed for resource-limited mobile devices and edge environments. Unlike BERT-Base, MobileBERT features a deeper architecture with 24 transformer layers (compared to BERT-Base’s 12 layers) but significantly reduces the number of parameters per layer. This design achieves an overall size reduction of approximately 4.3 times compared to BERT-Base, making it highly efficient in memory usage.
A key innovation in MobileBERT is its use of a bottleneck transformer structure, inspired by MobileNet. This architecture improves memory efficiency and computational speed, resulting in a 5.5x speedup over BERT-Base. Despite its compact size, MobileBERT delivers performance comparable to BERT-Base on benchmarks such as the GLUE dataset and even outperforms it on specific tasks like SQuAD v1.1 and v2.0 for question answering.
However, the bottleneck structure introduces additional complexity during the fine-tuning process, requiring careful selection of hyperparameters to achieve optimal performance. While this may pose challenges, MobileBERT remains a robust choice for resource-constrained devices, particularly in applications like IoT and distributed settings such as FL. Its efficient design and strong task performance make it ideal for mobile and edge-based AI applications where computational resources are limited. \\
\textbf{Link}: \href{https://huggingface.co/docs/transformers/en/model_doc/mobilebert}{MobileBert on Hugging Face} \\
\begin{center}
\begin{tcolorbox}
\vspace{-0.05in}
\tiny 
\textbf{Python Command:}
\begin{verbatim}
rom transformers import MobileBertTokenizer, MobileBertForSequenceClassification
import torch

tokenizer = MobileBertTokenizer.from_pretrained("google/mobilebert-uncased")
model = MobileBertForSequenceClassification.from_pretrained
("google/mobilebert-uncased")
\end{verbatim}
\end{tcolorbox}
\end{center}

\subsection{T5 Small}
\textbf{Parameters}: 60M \\
\textbf{Description}: T5-Small is a lightweight extension of the T5 (Text-to-Text Transformer) model introduced by Google. Trained on T5-Base (220 million parameters) and T5-Large (770 million parameters), T5-Small is designed to be efficient for resource-constrained devices. It has demonstrated successful results in tasks such as machine translation, document summarization, and other NLU applications.
Although T5-Small is slightly larger than TinyBERT and ALBERT, it is capable of handling more complex task domains. Additionally, its unified sequence-to-sequence architecture makes it an excellent choice for multi-task FL settings, where diverse task requirements can benefit from its versatility and efficiency.\\
\textbf{Link}: \href{https://huggingface.co/t5-small}{T5 Small on Hugging Face} \\
\begin{center}
\begin{tcolorbox}
\vspace{-0.05in}
\tiny 
\textbf{Python Command:}
\begin{verbatim}
from transformers import T5ForConditionalGeneration, T5Tokenizer

model_name = "t5-small"
model = T5ForConditionalGeneration.from_pretrained(model_name)
tokenizer = T5Tokenizer.from_pretrained(model_name)
\end{verbatim}
\end{tcolorbox}
\end{center}

\subsection{Phi-2 }
\textbf{Parameters}: 2.7B \\
\textbf{Description}: Phi-2 is an LLM developed by Microsoft, known for its exceptional efficiency and ability to outperform models up to 25 times larger. This efficiency is achieved through several architectural innovations, including: (1) sparse connections or computations, which reduce computational overhead without sacrificing accuracy, (2) advanced attention mechanisms for optimized information flow, and (3) cross-layer parameter sharing, similar to ALBERT, which decreases the model’s overall parameter count while maintaining high performance.
Phi-2 has demonstrated impressive performance across a wide range of datasets and tasks, including mathematical reasoning, coding, debugging, and logical reasoning. Its ability to handle diverse tasks is enhanced by pre-training on multiple tasks simultaneously, making it highly adaptable to new and unseen datasets. Unlike models trained on broad datasets, Phi-2 uses a targeted training approach on carefully curated datasets tailored to specific objectives. This strategy ensures robust and efficient performance across various applications, especially in resource-constrained environments.
Phi-2 also has a more lightweight counterpart, Phi-1.5, which consists of 1.3 billion parameters compared to Phi-2’s 2.7 billion parameters. Designed specifically for resource-constrained devices, Phi-1.5 maintains several of the architectural advantages of Phi-2, including sparse connections and attention optimizations, but it trades off some performance for reduced size and computational requirements. While Phi-1.5 is suitable for on-device applications and environments with limited hardware, it does not match Phi-2’s capabilities in tasks such as conversational AI, advanced reasoning, and complex logical problem-solving.
Notably, Phi-1.5 is well-suited for tasks requiring efficiency and adaptability, such as natural language understanding (NLU) and lightweight code analysis, making it an excellent choice for mobile devices and edge computing. However, users must weigh the trade-offs between size, efficiency, and task performance when selecting between Phi-1.5 and Phi-2. \\
\textbf{Link}: \href{https://www.microsoft.com/en-us/research/blog/phi-2-the-surprising-power-of-small-language-models/}{Phi-2 Model Details} \href{https://huggingface.co/microsoft/phi-1.5}{Phi-1.5 on Hugging Face} \\
\begin{center}
\begin{tcolorbox}
\vspace{-0.05in}
\tiny 
\textbf{Python Command:}
\begin{verbatim}
from transformers import AutoTokenizer, AutoModelForCausalLM

# Load the Phi-2 tokenizer and model
tokenizer = AutoTokenizer.from_pretrained("microsoft/phi-2")
model = AutoModelForCausalLM.from_pretrained("microsoft/phi-2")

# Load the Phi-1.5 tokenizer and model
model_name = "microsoft/phi-1.5"
model = AutoModelForCausalLM.from_pretrained(model_name)
tokenizer = AutoTokenizer.from_pretrained(model_name)
\end{verbatim}
\end{tcolorbox}
\end{center}

\subsection{Gemini Nano}
\textbf{Parameters}: 1.8 B \\
\textbf{Description}: Gemini Nano, introduced by Google, is a specialized LLM designed for smartphones and mobile devices. This language model is available in two versions: (1) Nano-1, featuring 1.8 billion parameters for low-memory devices, and (2) Nano-2, with 3.25 billion parameters for devices with higher memory capacity. Gemini Nano has been deployed on Android and iOS platforms for mobile-specific tasks such as document summarization, image and video editing, and speech processing.
This model is designed for on-device training, allowing it to operate on private user data without reliance on the cloud or a central server. As such, Gemini Nano is an excellent choice for privacy-preserving distributed learning platforms like FL. Its ability to enable personalized on-device training without requiring an internet connection makes it one of the most efficient LLMs for smartphones, providing a tailored solution for resource-constrained mobile environments.  \\
\textbf{Link}: \href{https://deepmind.google/technologies/gemini/nano/?utm_source=chatgpt.com}{Gemini Nano Details} \\
\begin{center}
\begin{tcolorbox}
\vspace{-0.05in}
\tiny 
\textbf{Python Command: The code for Gemini-Nano is not publicly available}
\begin{verbatim}

\end{verbatim}
\end{tcolorbox}
\end{center}

\subsection{Claude Instant}
\textbf{Parameters}: $\leq$ 1 B (not disclosed) \\
\textbf{Description}: Claude Instant, developed by Anthropic (founded by former OpenAI researchers), is a compact version of Claude 2. It has been released as a faster, cheaper, and smaller LLM, making it an excellent candidate for smartphones and resource-limited devices. Despite its reduced size, Claude Instant has demonstrated a high-quality training process, excelling in complex tasks such as casual dialogue, document comprehension, and text generation compared to the larger models such as Claude 2.
This model is particularly well-suited for real-time applications, such as chatbots and virtual assistants, as it outperforms larger models while requiring significantly less memory. Additionally, aligned with Anthropic’s goals, Claude Instant emphasizes safety, with thorough studies ensuring it mitigates biased, harmful, or unethical outputs. Furthermore, this LLM can be tuned and updated based on user feedback and fine-tuned in accordance with ethical guidelines, ensuring adaptability and responsible deployment. \\
\textbf{Link}: \href{https://aibusiness.com/nlp/anthropic-s-claude-instant-a-smaller-faster-and-cheaper-language-model}{Claude Instant Overview} \\
\begin{center}
\begin{tcolorbox}
\vspace{-0.05in}
\tiny 
\textbf{Python Command: The code for Claude Instant is only accessible through Anthropic API}
\begin{verbatim}

\end{verbatim}
\end{tcolorbox}
\end{center}

\subsection{ByT5 Small}
\textbf{Parameters}: 300M \\
\textbf{Description}: Similar to T5-Small, ByT5 is a larger version of the T5 model introduced by Google. This tokenization-free LLM processes text at the byte level rather than relying on tokenized inputs, making it particularly effective for diverse languages that are sensitive to spelling and pronunciation variations. ByT5-Small is resilient and robust against noise, making it well-suited for handling short text sequences and unstructured data. For example, it has demonstrated superior performance on the TweetQA dataset, which contains informal and varied text, compared to T5-Small.
While the tokenization-free characteristic of ByT5-Small is advantageous for resource-constrained edge devices—eliminating the need for intensive text pre-processing—it results in longer text sequences, which require more computational resources during training and inference. Despite this trade-off, ByT5-Small’s robustness makes it a compelling choice for applications involving noisy or informal text data.    \\
\textbf{Link}: \href{https://huggingface.co/google/byt5-small}{ByT5 Small on Hugging Face} \\
\begin{center}
\begin{tcolorbox}
\vspace{-0.05in}
\tiny 
\textbf{Python Command:}
\begin{verbatim}
from transformers import T5ForConditionalGeneration, T5Tokenizer

model_name = "google/byt5-small"
model = T5ForConditionalGeneration.from_pretrained(model_name)
tokenizer = T5Tokenizer.from_pretrained(model_name)
\end{verbatim}
\end{tcolorbox}
\end{center}

\subsection{Flan-T5 Small}
\textbf{Parameters}: 80M \\
\textbf{Description}: Flan-T5 Small is another lightweight variant of Google’s T5 model, trained specifically on human-like prompts. It has been fine-tuned on a wide range of multi-modal datasets simultaneously, enhancing its adaptability to new tasks and unseen data. Additionally, Flan-T5 Small has been instruction-tuned, meaning it is highly responsive to human language commands and user-based instructions.
This model demonstrates superior generalization performance, making it a strong choice for decentralized settings and local devices. However, to fully leverage its capabilities, Flan-T5 Small requires pre-training on multi-modal and versatile datasets, which may not always be accessible in resource-constrained environments. Despite this limitation, its ability to handle diverse tasks and its responsiveness to instructions make it a valuable tool for various applications. \\
\textbf{Link}: \href{https://huggingface.co/google/flan-t5-small}{Flan-T5 Small on Hugging Face} \\
\begin{center}
\begin{tcolorbox}
\vspace{-0.05in}
\tiny 
\textbf{Python Command:}
\begin{verbatim}
from transformers import T5ForConditionalGeneration, T5Tokenizer

model_name = "google/flan-t5-small"
model = T5ForConditionalGeneration.from_pretrained(model_name)
tokenizer = T5Tokenizer.from_pretrained(model_name)
\end{verbatim}
\end{tcolorbox}
\end{center}

\subsection{BLOOMZ 560M}
\textbf{Parameters}: 560M \\
\textbf{Description}: BLOOMZ, a smaller version of BLOOM introduced by BigScience, is optimized for zero-shot tasks. This model has been fine-tuned across a wide range of tasks and multiple languages, making it a powerful LLM for applications such as machine translation and question answering. BLOOMZ is trained on the xP3 dataset, an extension of the P3 (Public Pool of Prompts) dataset, which incorporates multilingual prompts and instructions. Similar to Flan-T5 Small, this training provides BLOOMZ with strong instruction-following capabilities.
Additionally, BLOOMZ can transfer knowledge from high-resource languages (with abundant samples) to low-resource languages (with fewer samples), making it particularly beneficial in decentralized settings and across diverse mobile and IoT devices that handle multiple languages. Like other compact models, BLOOMZ is primarily used for natural language understanding (NLU) tasks and translation, offering efficiency and adaptability in multilingual environments.   \\
\textbf{Link}: \href{https://huggingface.co/bigscience/bloomz-560m}{BLOOMZ 560M on Hugging Face} \\
\begin{center}
\begin{tcolorbox}
\vspace{-0.05in}
\tiny 
\textbf{Python Command:}
\begin{verbatim}
from transformers import BloomForCausalLM, BloomTokenizerFast

model_name = "bigscience/bloomz-560m"
model = BloomForCausalLM.from_pretrained(model_name)
tokenizer = BloomTokenizerFast.from_pretrained(model_name)
\end{verbatim}
\end{tcolorbox}
\end{center}

\subsection{OPT 350M}
\textbf{Parameters}: 350M \\
\textbf{Description}: OPT-350M is a model within the Open Pre-trained Transformer (OPT) series of LLMs developed by Meta AI. The OPT series, designed primarily for research purposes, includes models ranging from 125 million to 175 billion parameters. OPT-350M strikes a solid balance between performance and efficiency, making it suitable for on-device training, particularly due to its transformer-based architecture.
Like GPT-3, OPT-350M is trained using causal language modeling, which makes it well-suited for tasks such as text generation and dialogue. Additionally, this model is open-source and compatible with GPT-3 pre-trained models, allowing for flexibility and integration into a variety of applications. Its performance-efficiency trade-off and open accessibility make it a valuable tool for resource-constrained environments and research-driven projects.   \\
\textbf{Link}: \href{https://huggingface.co/facebook/opt-350m}{OPT 350M on Hugging Face} \\
\textbf{Python Command}:
\begin{center}
\begin{tcolorbox}
\vspace{-0.05in}
\tiny 
\textbf{Python Command:}
\begin{verbatim}
from transformers import OPTForCausalLM, AutoTokenizer

model_name = "facebook/opt-350m"
model = OPTForCausalLM.from_pretrained(model_name)
tokenizer = AutoTokenizer.from_pretrained(model_name)
\end{verbatim}
\end{tcolorbox}
\end{center}

\subsection{Galactica 125M}
\textbf{Parameters}: 125M \\
\textbf{Description}: Galactica-125M is the smallest model in Meta AI’s Galactica series, which includes language models ranging from 125 million to 120 billion parameters. The Galactica series is specifically designed for scientific use, with models trained exclusively on scientific corpora. These models are tailored for tasks such as scientific question answering, mathematical reasoning and equations, molecular property prediction, and generating text with accurate citations.
Despite its relatively small size, Galactica-125M excels in task-specific scientific domains compared to other small LLMs, making it an excellent candidate for resource-constrained mobile devices in scientific applications. However, it lacks the generalization capabilities required for more complex NLP tasks, such as open-ended text generation or human dialogue, limiting its use outside the scientific domain. \\
\textbf{Link}: \href{https://huggingface.co/meta/galactica-125m}{Galactica 125M on Hugging Face} \\
\textbf{Python Command}:
\begin{center}
\begin{tcolorbox}
\vspace{-0.05in}
\tiny 
\textbf{Python Command:}
\begin{verbatim}
from transformers import AutoModelForCausalLM, AutoTokenizer

model_name = "meta/galactica-125m"
model = AutoModelForCausalLM.from_pretrained(model_name)
tokenizer = AutoTokenizer.from_pretrained(model_name)
\end{verbatim}
\end{tcolorbox}
\end{center}

\subsection{Pythia 160M}
\textbf{Parameters}: 160M \\
\textbf{Description}: Pythia-160M is a compact model in the Pythia suite, which ranges from 70 million to 12 billion parameters. Released by EleutherAI, Pythia models are trained on The Pile, an 800GB corpus of diverse text data gathered from sources such as webpages, books, and magazines, similar to GPT-Neo. Pythia is open-source and free of licensing requirements, offering model checkpoints that enable researchers to observe and study model behavior, making it valuable for research purposes.
Although Pythia-160M is not the smallest model in the suite, it strikes a balance between generalization and efficiency, making it well-suited for resource-limited devices and smartphones in distributed settings. Like the GPT series, Pythia models use a decoder-based transformer architecture and are proficient in natural language understanding (NLU) tasks as well as text generation. This combination of accessibility, adaptability, and performance makes Pythia-160M a compelling choice for lightweight applications. \\
\textbf{Link}: \href{https://huggingface.co/EleutherAI/pythia-160m}{Pythia 160M on Hugging Face} \\
\textbf{Python Command}:
\begin{center}
\begin{tcolorbox}
\vspace{-0.05in}
\tiny 
\textbf{Python Command:}
\begin{verbatim}
from transformers import GPTNeoXForCausalLM, AutoTokenizer

model_name = "EleutherAI/pythia-160m"
model = GPTNeoXForCausalLM.from_pretrained(model_name)
tokenizer = AutoTokenizer.from_pretrained(model_name)
\end{verbatim}
\end{tcolorbox}
\end{center}

\subsection{Cerebras-GPT 256M}
\textbf{Parameters}: 256M \\
\textbf{Description}: Cerebras-GPT is a compact model in the Cerebras-GPT family, introduced by Cerebras Systems. This series ranges from 111 million to 13 billion parameters and is trained on The Pile corpus, similar to Pythia and GPT-Neo. It utilizes a decoder-only transformer architecture, aligning with the GPT LLM series. As a result, Cerebras-GPT models are well-suited for both natural language understanding (NLU) and text generation tasks, as well as research-oriented deployments due to their scalability and open-source availability.
A notable highlight of the Cerebras-GPT series, compared to models like Pythia and GPT-Neo, is its adherence to the Chinchilla Scaling Law rather than traditional scaling laws such as Kaplan’s Laws. Chinchilla Scaling optimizes the model’s density by training smaller models on larger datasets, leading to a better size-to-data ratio and more efficient use of computational and memory resources. This optimization makes Cerebras-GPT particularly advantageous for mobile and resource-constrained devices, where efficiency per model size is critical.  \\
\textbf{Link}: \href{https://huggingface.co/cerebras/Cerebras-GPT-256M}{Cerebras-GPT 256M on Hugging Face} \\
\textbf{Python Command}:
\begin{center}
\begin{tcolorbox}
\vspace{-0.05in}
\tiny 
\textbf{Python Command:}
\begin{verbatim}
from transformers import GPT2LMHeadModel, GPT2Tokenizer

model_name = "cerebras/Cerebras-GPT-256M"
model = GPT2LMHeadModel.from_pretrained(model_name)
tokenizer = GPT2Tokenizer.from_pretrained(model_name)
\end{verbatim}
\end{tcolorbox}
\end{center}

\subsection{TinyLlama 1.1B}
\textbf{Parameters}: 1.1B \\
\textbf{Description}: TinyLLaMA is a compact version of the LLaMA-2 series, optimized for resource-constrained environments. This open-source LLM outperforms other models of similar size in relatively complex tasks such as text generation and natural language understanding (NLU). Its superior performance is attributed to its training on a massive dataset of 3 trillion tokens over three rounds of training.
TinyLLaMA reduces computational and memory overhead by incorporating advanced techniques like FlashAttention and LitGPT. FlashAttention eliminates the need to store intermediate computation matrices, enhancing memory efficiency, while LitGPT provides a lightweight, modular design of the GPT architecture, specifically tailored for deployment on resource-constrained devices and distributed settings like FL.
While TinyLLaMA demonstrates strong generalization across a wide range of tasks, pretraining it for domain-specific applications requires significant computational resources (e.g., approximately 90 days on 16 A100-40G GPUs), making it impractical for resource-limited devices. However, once pretrained, its moderate size and efficiency allow it to excel in on-device training, outperforming many comparable LLMs in similar environments.  \\
\textbf{Link}: \href{https://huggingface.co/meta/tiny-llama-1.1b}{TinyLlama 1.1B on Hugging Face} \\
\textbf{Python Command}:
\begin{center}
\begin{tcolorbox}
\vspace{-0.05in}
\tiny 
\textbf{Python Command:}
\begin{verbatim}
from transformers import LlamaForCausalLM, LlamaTokenizer

model_name = "meta/tiny-llama-1.1b"
model = LlamaForCausalLM.from_pretrained(model_name)
tokenizer = LlamaTokenizer.from_pretrained(model_name)
\end{verbatim}
\end{tcolorbox}
\end{center}

\subsection{MiniCPM-1.2B }
\textbf{Parameters}: 1.2 B \\
\textbf{Description}: MiniCPM-1.2B is a moderately sized version of the MiniCPM model by OpenBMB, with the series ranging from 1.2 billion to 13 billion parameters. It is designed to provide efficient performance on resource-constrained edge devices. Trained on a large text corpus, MiniCPM has demonstrated results comparable to much larger models like LLaMA 2 (7B+), making it a competitive option in its size category.
This open-source LLM supports multilingual tasks, including Chinese and English, and excels in applications such as text summarization and image and video understanding. Notably, MiniCPM has been recognized as the first real-time video understanding VLM for edge devices, highlighting its innovative capabilities.
MiniCPM-1.2B incorporates advanced techniques such as the Warmup-Stable-Decay (WSD) learning rate scheduler and "Wind Tunnel" experiments to enhance its performance. These methods improve multi-task adaptability and processing speed, making it a versatile and efficient choice for edge device deployments.  \\
\textbf{Link}: \href{https://huggingface.co/openbmb/MiniCPM-V-2_6}{
MiniCPM-2B on Hugging Face} \\
\textbf{Python Command}:
\begin{center}
\begin{tcolorbox}
\vspace{-0.05in}
\tiny 
\textbf{Python Command:}
\begin{verbatim}
from transformers import AutoModelForCausalLM, AutoTokenizer
import torch
torch.manual_seed(0)

path = 'openbmb/MiniCPM-2B-sft-bf16'
tokenizer = AutoTokenizer.from_pretrained(path)
model = AutoModelForCausalLM.from_pretrained(path,
torch_dtype=torch.bfloat16, device_map='cuda', trust_remote_code=True)
\end{verbatim}
\end{tcolorbox}
\end{center}

\subsection{Gemma (2B)}
\textbf{Parameters}: 2B \\
\textbf{Description}: Gemma-2B is a Google-based model from the Gemma family, designed as a lightweight LLM for laptops, PCs, and personal devices. This open-source, moderately sized model incorporates advanced features such as the GeGLU activation function, which employs a gating mechanism to enhance and control information flow and propagation through the network, resulting in improved generalization. Additionally, it utilizes Grouped Query Attention (GQA) to further optimize performance. More over, Gamma-2B consists of 26-layer and is exploiting GeGLU (Gated Linear Unit with GeLU) with 8 attention heads. As a result, this large extensive network with 18,432 feed-forward dimension is prominent in handling complex tasks with longer sequences (e.g., longer conversation reasoning, long document summarization and remains as one of the medium-size compact LLMs which still could be used in laptop devices and smartphones. 
Gemma-2B has been trained on a massive multilingual dataset comprising 2 trillion tokens, making it capable of handling diverse language tasks. For instruction-based tasks, the fine-tuned variant Gemma-2B-IT has been trained on human dialects and instruction-based data, making it particularly suitable for conversational AI applications. This combination of efficiency, multilingual support, and adaptability makes Gemma-2B a strong choice for personal devices and real-time conversational systems.  \\
\textbf{Link}: \href{https://huggingface.co/google/gemma-2b}{Gemma 2B on Hugging Face} \\
\textbf{Python Command}:
\begin{center}
\begin{tcolorbox}
\vspace{-0.05in}
\tiny 
\textbf{Python Command:}
\begin{verbatim}
from transformers import AutoModelForCausalLM, AutoTokenizer
import torch
torch.manual_seed(0)

path = 'openbmb/MiniCPM-2B-sft-bf16'
tokenizer = AutoTokenizer.from_pretrained(path)
model = AutoModelForCausalLM.from_pretrained(path,
torch_dtype=torch.bfloat16, device_map='cuda', trust_remote_code=True)
\end{verbatim}
\end{tcolorbox}
\end{center}

\subsection{Reformer}
\textbf{Parameters}: 110 M \\
\textbf{Description}: Reformer is a memory-efficient LLM introduced by Google that incorporates several innovative techniques to reduce computational complexity and memory usage. One of its key features is the use of Locality-Sensitive Hashing (LSH) attention, which reduces the computational complexity of traditional transformers from $O(n^2) \rightarrow O(nlogn)$, enabling efficient processing of longer sequences. Additionally, Reformer utilizes Reversible Residual Layers, which recreate intermediate activations during back-propagation instead of storing them, significantly reducing memory requirements. Furthermore, its large layers are partitioned into smaller chunks for computation, further optimizing memory usage.
These three key innovations—LSH attention, Reversible Residual Layers, and layer partitioning—substantially decrease the memory footprint and computational demands of the model, making it highly suitable for resource-constrained environments. In terms of architecture, the Reformer model is similar in size to BERT (12 layers) but is capable of handling much more complex tasks and extracting dependencies over sequences approximately 10 times longer, comparable to models like Gemma-2B.
However, while the Reformer excels at handling long sequences (e.g., long document summarization), its performance on shorter sequences does not necessarily surpass other models. In summary, the Reformer’s memory efficiency and reduced computational complexity make it an excellent choice for applications involving long sequences on resource-constrained devices, such as smartphones and other edge devices.  \\
\textbf{Link}: \href{https://huggingface.co/docs/transformers/en/model_doc/reformer}{Reformer model on Hugging Face} \\
\begin{center}
\begin{tcolorbox}
\vspace{-0.05in}
\tiny 
\textbf{Python Command:}
\begin{verbatim}
from transformers import ReformerTokenizer, ReformerModelWithLMHead

tokenizer = ReformerTokenizer.from_pretrained
("google/reformer-crime-and-punishment")
model = ReformerModelWithLMHead.from_pretrained
("google/reformer-crime-and-punishment")
\end{verbatim}
\end{tcolorbox}
\end{center}

\end{document}